\documentclass[letterpaper, 10pt, journal,twoside]{IEEEtran}

\IEEEoverridecommandlockouts                              
\usepackage{graphicx}
\usepackage[utf8]{inputenc}
\usepackage[T1]{fontenc}
\usepackage{amsfonts}
\newcommand{\fig}[1]{Fig.~\ref{#1}}
\newcommand{\tab}[1]{Table~\ref{#1}}

\usepackage{amsmath} 
\usepackage{amssymb}  
\usepackage{subcaption}
\usepackage{color}
\usepackage{verbatim}
\usepackage{wasysym}
\usepackage[table,xcdraw]{xcolor}
\usepackage{gensymb}
\usepackage{latexsym}
\usepackage{multirow}
\usepackage{siunitx}
\usepackage{booktabs}
\usepackage[noadjust]{cite}
\usepackage[flushleft]{threeparttable}
\usepackage[normalem]{ulem} 

\newcommand{\rotateaxisleft}{\rotatebox{90}{\raisebox{0.15ex}{\sout{\,\textbf{\resizebox{1.4ex}{1.5ex}{\leftturn}}\,}}}}

\newcommand{\RNum}[1]{\uppercase\expandafter{\romannumeral #1\relax}}
\makeatletter
\newlength\tmp@\newlength\t@mp
\newcommand{\comp}[3]
  {\mathop{ \settowidth\tmp@{$\displaystyle\mathop{#1}^{#3}_{#2}$}
  \hbox to \tmp@{\hss \settowidth\t@mp{$\displaystyle #1$}\setlength\t@mp{.45\t@mp}
  $\displaystyle\mathop{#1}^{\hspace\t@mp #3}_{\hspace{-\t@mp}#2}$
  \hss} }}
\newcommand{\red}[1]{\textcolor{black}{#1}}
\newcommand{\redrev}[1]{\textcolor{black}{#1}}

\makeatother
\title{\LARGE \bf
SCALER: Versatile Multi-Limbed Robot for Free-Climbing in Extreme Terrains}

\author{Yusuke Tanaka$^{1}$, Yuki Shirai$^{1}$, Alexander Schperberg$^{1}$, Xuan Lin$^{1}$, and Dennis Hong$^{1}$
}

\begin{document}
\twocolumn[{%
\renewcommand\twocolumn[1][]{#1}%
    \maketitle
    \begin{center}
        \centering
        \includegraphics[width=0.99\textwidth,trim={0cm 0cm 0cm 0cm},clip]{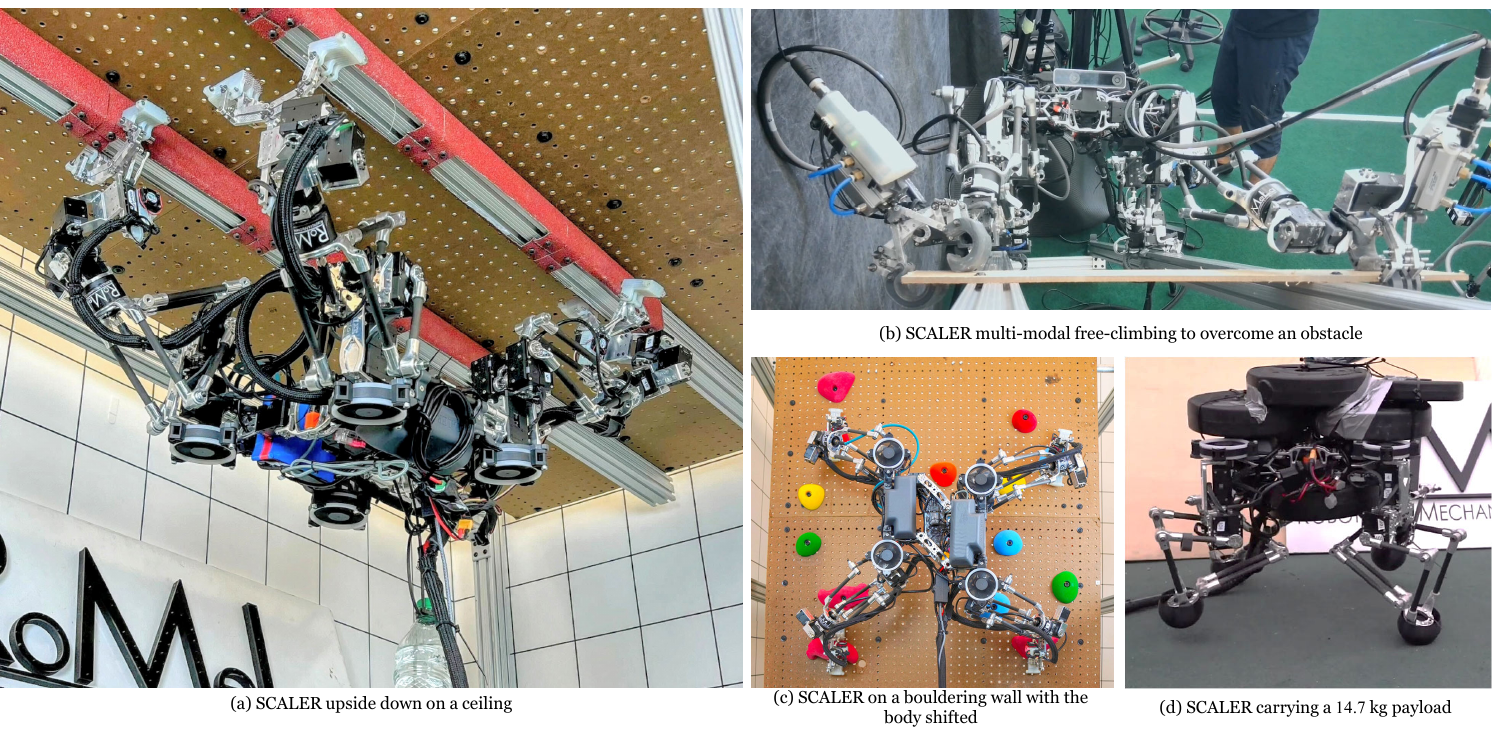} %
    \captionof{figure}{SCALER, a versatile multi-limbed robot for free-climbing in extreme terrains.}
    \label{fig:fig1}
    \end{center}%
    }]
    \footnotetext[1]{All authors are with the Department of Mechanical and Aerospace Engineering, University of California, Los Angeles, CA, USA 90095 {\tt\footnotesize \{yusuketanaka, yukishirai4869, aschperberg28, maynight, dennishong\}@g.ucla.edu.}}%
\thispagestyle{empty}
\pagestyle{empty}

\begin{abstract}
This paper presents SCALER, a versatile free-climbing multi-limbed robot that is designed to achieve tightly coupled simultaneous locomotion and dexterous grasping. 
While existing quadrupedal-limbed robots have demonstrated impressive dexterous capabilities, achieving a balance between power-demanding locomotion and precise grasping remains a critical challenge.
We design a torso mechanism and a parallel-serial limb to meet the conflicting requirements that pose unique challenges in hardware design. 
SCALER employs underactuated two-fingered GOAT grippers that can mechanically adapt and offer seven modes of grasping, enabling SCALER to traverse extreme terrains with multi-modal grasping strategies. We study the whole-body approach, where SCALER utilizes its body and limbs to generate additional forces for stable grasping in various environments, thereby further enhancing its versatility. 
Furthermore, we improve the GOAT gripper actuation speed to realize more dynamic climbing in a closed-loop control fashion.
With these proposed technologies, SCALER can traverse vertical, overhanging, upside-down, slippery terrains and bouldering walls with non-convex-shaped climbing holds under the Earth's gravity.
\end{abstract}


\section{INTRODUCTION}
\bstctlcite{IEEEexample:BSTcontrol}

\begin{figure}[t!]
 \centering
    \includegraphics[width=0.49\textwidth, trim={0cm 0cm 0cm 0cm},clip]{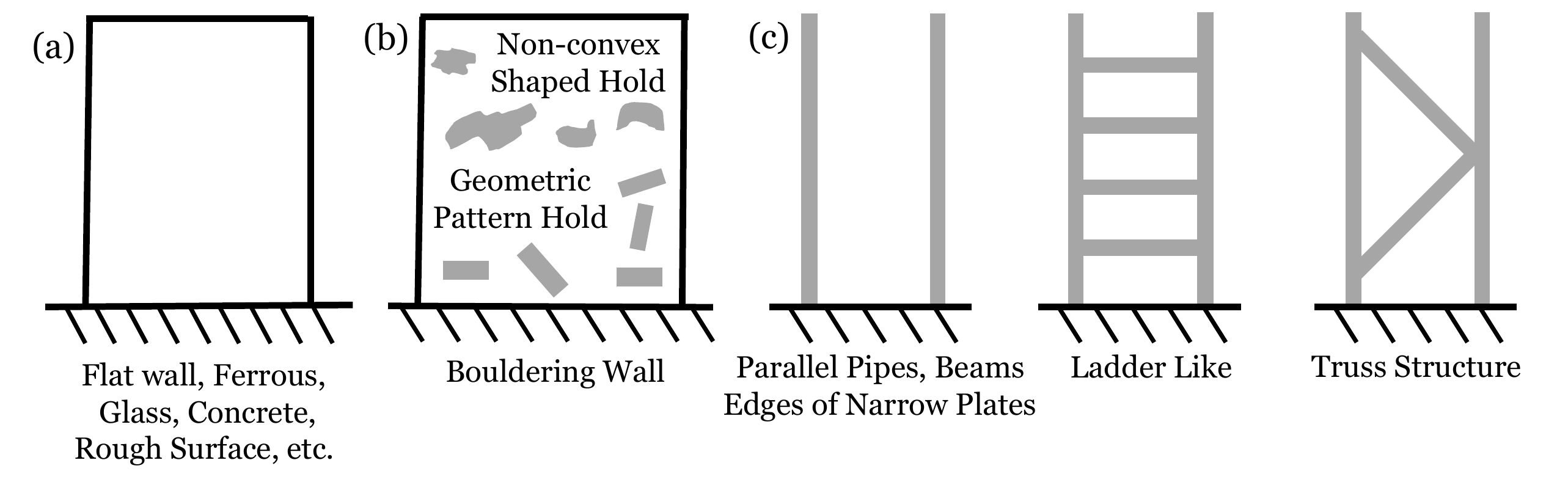}
  \caption{The type of climbing environments in terms of their continuity. (a) a continuous environment. (b) a discrete environment. (c) directionally continuous environments.}
  \label{fig:environment}
\end{figure}

The field of legged robotics has set the stage for applications such as search and rescue, delivery, and even extraterrestrial exploration \cite{hubrobo}. They have shown versatility and maneuverability in locomotion \cite{anymal} and have demonstrated contact-rich tasks such as object manipulation with their own bodies \cite{sombolestan2022hierarchical}. Limbed robots have further showcased grasping, using collective legs \cite{shi2021circus}, utilizing a dedicated arm \cite{fu2023deep}, or employing grasping end-effectors \cite{6094406}. The simultaneous interaction between locomotion and grasping, called loco-grasping, allows them to execute tasks such as carrying objects, using tools, or traversing challenging terrains \cite{gong2023legged}. 
However, current limbed robots are not yet sufficient to accomplish one of the extreme cases of loco-grasping, which is free-climbing \cite{free_climb}, and it is essential for multi-limbed robots to facilitate even more versatile traversability over buildings, construction sites, caves, etc. This includes not only traversing the continuous terrains defined in \fig{fig:environment} but also discrete environments in \fig{fig:environment}b and directionally continuous in \fig{fig:environment}c.

\begin{figure}[t!]
    \centering
    \includegraphics[width=0.7\linewidth, trim={0cm 1cm 0cm 1.25cm},clip]{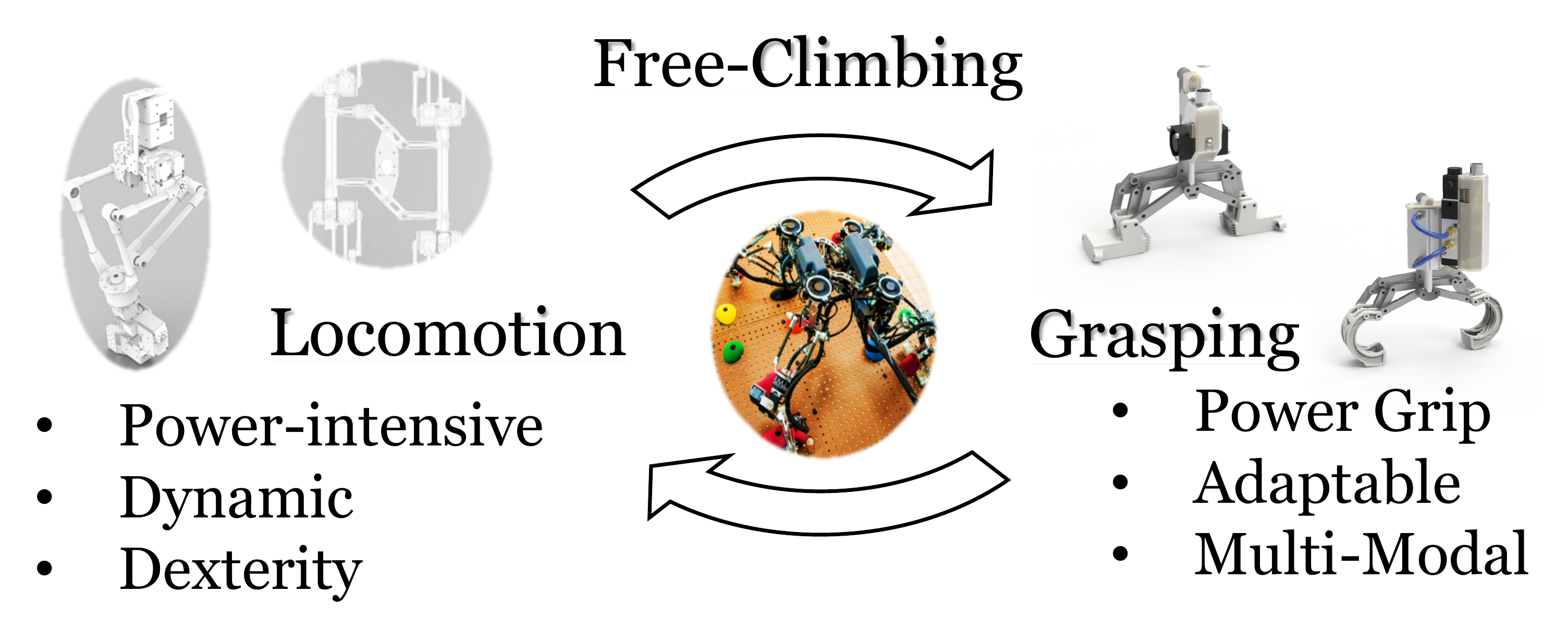}
     \caption{The circular dependency diagram of the strict loco-grasping, free-climbing problem. SCALER requires a balance of both locomotion and grasping capabilities to achieve free-climbing.
     \label{fig:cycle}}
\end{figure}

However, achieving free-climbing raises substantial challenges, which mandate various climbing techniques and grasping adaptability as demonstrated in \fig{fig:fig1}. 
Climbers must consider foot placements and body stability \cite{ladder_climb}, particularly in a discrete terrain environment such as in \fig{fig:environment}b. 
This imposes \textit{strict} loco-grasping requirements: failure in either locomotion or grasping can result in a fall and mission abort. Hence, free-climbing robots must consider closely coupled grasping and locomotion as illustrated in \fig{fig:cycle}. 

For tackling complex and cluttered terrains, performing multi-modal grasping becomes essential to improve climbing feasibility and stability.
Multi-modal grasping plays a fundamental role in human climbing to utilize various types of grips based on climbing \red{scenarios} \cite{beal2011bouldering}. 
Different grasping modes such as pinch, envelope, crimp, and pocket are strategically deployed to conquer terrains that would be otherwise difficult or impossible to travel \cite{beal2011bouldering}, \cite{balaguer2005climbing}. 
Whole-body approaches, such as the \textit{sidepull} grasping technique \cite{beal2011bouldering}, add a new set of potential abilities in climbing robots as they embrace the entire body and limbs to enhance grasping feasibility \red{when} holds are infeasible to grasp stably with one gripper.
Therefore, free-climbing tasks must involve successful approaches and stable grips by switching among various grasping modes, depending on the environmental context and climbing requirements. 

While existing platforms have paved the way for loco-grasping, there is still a gap in the strict loco-grasping and multi-modal ability in free-climbing. 
To address these challenges, we introduce SCALER: Spine-enhanced Climbing Autonomous Limbed Exploration Robot, a versatile quadruped-limbed research platform shown in \fig{fig:fig1}.
SCALER is designed to traverse a range of extreme surfaces. SCALER can cross obstacles using multi-modal \red{grasping} and whole-body approaches with our \red{proposed} unique C-shaped finger designs.

Our contributions are summarized as follows:
\begin{enumerate}
    \item SCALER's Mechanisms: We propose the SCALER torso and limb mechanisms for strictly coupled loco-grasping free-climbing. SCALER realizes versatile capabilities of traversing on the ground, vertical walls, overhangs, and ceilings with payload \textit{under the Earth's gravity}.
    \item Underactuated GOAT Grippers: We employ an improved mechanically adaptive GOAT gripper with spine tips.
    \item Hardware Validation: We extensively validate the SCALER free-climber platform with the spine GOAT grippers in hardware experiments.
\end{enumerate}

This paper extends our previous work \cite{yusuke_scaler_2022} by introducing additional contributions related to dynamic locomotion and versatile climbing strategies, thereby advancing state-of-the-art robotic free-climbing.
\begin{enumerate}
    \item Dynamic Climbing and Gait: We assess SCALER's capabilities during dynamic climbing with a closed-loop control fashion and untethered operation. We introduce a modified trot gait to \red{utilize limb stiffness and its model}. 
    \item Multi-Modal Grasping: We propose C-shaped finger GOAT grippers, C-GOATs that can realize up to $7$ modes of grasping with SCALER's dexterity. This capability enables SCALER to apply versatile strategies, such as a whole-body approach, to overcome scenarios that would otherwise be infeasible to climb.
    \item \red{Analysis and Multi-Modal Climbing with C-GOAT: We analyze the C-GOAT characteristics and grasping capabilities in climbing tasks.}
    \item Hardware Validation: We validate SCALER dynamic gait and multi-modality in various hardware experiments.
\end{enumerate}

To the best of our knowledge, SCALER is the first robot to demonstrate a versatile suite of free-climbing abilities under \redrev{the} Earth's gravity, using more than \redrev{three} modes of grasping, conducting a whole-body approach in free-climbing \redrev{with our C-GOAT gripper}.
SCALER achieves this with SCALER's mechanical designs and the GOAT grippers. Our approach makes SCALER a pioneering reference for future developments in robotic loco-grasping problems and free-climbing.

\section{Related Work}\label{sec:related_work}
In this section, we review the current state-of-the-art approaches \redrev{in legged robotics, climbing robotics, and climbing end-effector design.}

\subsection{Quadruped Architecture}
\subsubsection{Torso Mechanisms}
Torso mechanisms based on animal spines have been analyzed and validated for ground and climbing robots. Cheetah spine motions are implemented to improve locomotion efficiency by adding compliance in the body \cite{cheetah}. \red{One} degree of freedom (DoF) in the torso has demonstrated effectiveness in steering motions \cite{spine_horizontal}. Inchworm gait extends and bends the body \red{for} a small soft robot to climb on a pole \cite{inchworm}.
Slalom \cite{slalom} replicated gecko in-plane bending movement, which reduced energy consumption in climbing by half that of the rigid body on a \SI{30}{\degree} slope.
We introduce a novel torso mechanism that is advantageous in restricted, contact-rich environments and offers superior scalability.

\subsubsection{Leg Mechanisms}
The latest quadruped robot technology utilizes quasi-direct drive torque actuators, such as those found in the MIT Cheetah \cite{cheetah}, or series elastic actuators used in ANYmal \cite{anymal}. While near-direct drive technology has proven successful on flat surfaces, it is not the most \red{effective} option for climbing, requiring continuous and intense power throughout the operation. One key difference between climbing and locomotion is that the robot must fight against gravity. Thus, the leg design sustains normal and shear forces under various gravity directions to support the robot \cite{marvel}.

LEMUR $3$ is specialized for climbing under reduced gravity environments, which sacrifices locomotion speed and dynamic motion capability in exchange for rich continuous and holding torque provided by $1$:$1200$ gear ratio motors \cite{lemur3}. HubRobo has demonstrated bouldering free-climbing under $0.38$ G using high-gear ratio motors in serial \cite{hubrobo, hubrobo_gripper}. \redrev{The} parallel five-bar linkage mechanism legs used in Minitaur \cite{five_bar_leg} and Doggo \cite{stanford_doggo} are competent in dynamic motion and have high output force because the two actuators' output power is linked. Bobcat \cite{bobcat} optimized the Minitaur link design for dynamic climbing, but parallel mechanisms increase complexity.
\redrev{Balancing power density, speed, mechanical efficiency, and workspace simultaneously is crucial when a single robot operates on the ground, runs, climbs vertically, and grasps objects dexterously.}

\subsection{Climbing Robotics}
Both wheeled \cite{wheel_climbing} and linkage mechanism-based legged climbing robots \cite{rise_bd} have successfully demonstrated climbing capabilities with spine-enhanced contacts for rock and concrete or with dry adhesive for clean flat surfaces, such as windows \cite{gekko_robot}. 
Soft robotics inspired by inchworm  \cite{inchworm} or octopus \cite{octopus} have replicated unique locomotion and grasping on a small scale. 
\red{Inchworm style form factor legged robots have shown climbing over different climbing planes thanks to their high torso traversability \cite{magnet_inchworm}.}
A cable climber is intended for suspension bridge visual inspection  \cite{suspension_bridge}. 
While these robots have shown promising vertical climbing abilities, they employ fixed locomotion methods, making them unsuitable for versatile applications such as traversing complex and cluttered environments requiring various grasping strategies. 
In contrast, SCALER is designed to embrace its loco-grasping capability.

LEMUR $3$ is a high DoF quadruped-limbed robot for space exploration missions \cite{lemur3}. HubRobo \cite{hubrobo, hubrobo_gripper} has reduced the hold grasping problem by employing a spine gripper that can passively grip a hold. 
However, neither can demonstrate free-climbing under the Earth's gravity. 

Dynamic climbing necessitates robots with balanced holding torque and speed and end-effectors capable of rapid and reliable contact transitions \cite{marvel}. 
RiSE \cite{rise_bd} and Bobcat \cite{bobcat} can run on a vertical wall with spine claw on rough concrete and a mesh surface, respectively. Marvel \cite{marvel} has an electromagnetic foot that can \red{dynamically} climb on ferrous surfaces. 
These implicit and adhesive end effectors cannot consider dexterous contacts essential in the loco-grasping domain. SCALER focuses on loco-grasping in climbing, which is necessary to traverse discrete environments. 
SCALER uses pneumatically actuated grippers for dynamic climbing.

\subsection{Climbing Contact Mechanisms}
Grippers for climbing can be categorized based on the type of grasping mechanisms they employ, such as implicit adhesive methods or explicit grasping with fingers.

\subsubsection{Adhesive Contact}
For smooth surfaces, magnetic \cite{marvel}, \cite{magneto}, \cite{magnet_inchworm}, \cite{magnetic_gripper} or suction-based \cite{suction_gripper, moclora} end effectors, dry adhesive toes such as a gecko gripper \cite{gekko_robot}, and Ethylene Propylene Diene Monomer Rubber \cite{slalom}, are viable options for climbing robots. 
LEMUR $2$B \cite{lemur3} and Capuchin \cite{Capuchin} have demonstrated bouldering wall climbing with high-friction rubber-wrapped end-effector hooks, which let them hang onto the holds \red{under the Earth's gravity}. 
The pure frictional force is sufficient when a robot climbs between two walls \cite{multi-surface}. 
For concrete, non-magnetic rough surfaces or loose cloth, spine-enhanced feet are desirable, such as in Spinybot II \cite{spinybot}, CLASH \cite{clash}, or the SiLVIA two-wall climber \cite{risk_aware}. Spines or needles can get inserted into microcavities of rocky surfaces \cite{hubrobo, hubrobo_gripper}. With an array of spines, the gripper can grasp concave and convex shapes while collecting the environment geometries \cite{spine_array_uno}.
These implicit and adhesive types of contacts can reduce the loco-grasping problem in climbing tasks to just locomotion as long as they can stick to an environment.

\subsubsection{Finger Grasping Contact}\label{sec:finger_grasping_contact}
Two-fingered grippers have been utilized in climbing robotics, such as ROMA I \cite{balaguer2005climbing}, which traverses truss structures, and Climbot \cite{6094406}, designed for vertical pole climbing and manipulation tasks.
Humanoid robots have shown ladder climbing with grippers that hook or encompass the ladder step \cite{ladder_climb, HRP2_ladder}, \redrev{\cite{yoshiike2017development, yoshiike2019experimental}}. 
LEMUR $3$ and HubRobo consist of radially aligned micro-spines, supporting the \redrev{robot's} weights under reduced gravity.
A hand-shaped climbing gripper, SpinyHand \cite{spiny_hand}, has four underactuated tendon-driven fingers that can switch between \redrev{two crimp and pinch grasping modes and has promising results for a human-scale climbing robot, but integration into a robot is yet to be demonstrated.}
As free-climbing robotics communities study various grasping systems, selecting graspable locations correspondingly starts playing more important roles \cite{hauser2018efficient, xu2020grappling, uno2020non, contact_rich}. 

While these works show impressive grasping results in climbing, these systems still lack versatility and multi-modality in Section \ref{sec:multi-modal-contact} that \redrev{particularly our C-shape fingered GOAT gripper provides in SCALER with seven distinct grasping modes}. SCALER employs GOAT grippers \cite{GOAT}, whippletree mechanism-based, passive underactuated two-fingered grippers that can mechanically adapt. Two variants of fingertips, spine, and dry adhesive are used in the GOAT grippers to \red{address} different scenarios.

\subsubsection{Multi-Modal Contact\label{sec:multi-modal-contact}}
Multi-modal grasping, where more than one type of grasp can be realized based on the geometries of objects or tasks \cite{human_multi_modal}, is an effective method to improve the range of graspable objects. Multi-modal capable grippers consist of multiple DoFs and underactuation, such as the Robotiq 3-finger adaptive gripper, which has five grasp types \cite{robotiq}. Although these approaches are common in gripper communities, it is rare for climbing robots to use more than two grasping modes. SpinyHand \cite{spiny_hand} attempts to utilize multi-modality in climbing, but integration into the quadruped climbing robot is yet to be achieved. 

Furthermore, a human climber can utilize their whole body to make difficult-to-grasp holds workable. This whole-body approach \cite{beal2011bouldering} can apply additional force on a hand to stabilize the grasp by using the whole-body to pull it. The whole-body approach can also increase frictional force, resulting in climbing on otherwise impossible terrains.

Our \red{proposed C-}GOAT gripper enables \redrev{seven modes of grasping} using only a single actuator. This adaptability sets SCALER apart from existing climbing robots and lets SCALER free-climb over obstacles while grasping them.


 \begin{figure}[t!]
    \centering
    \includegraphics[width=0.42\textwidth, trim={0cm 1cm 1cm 12cm},clip]{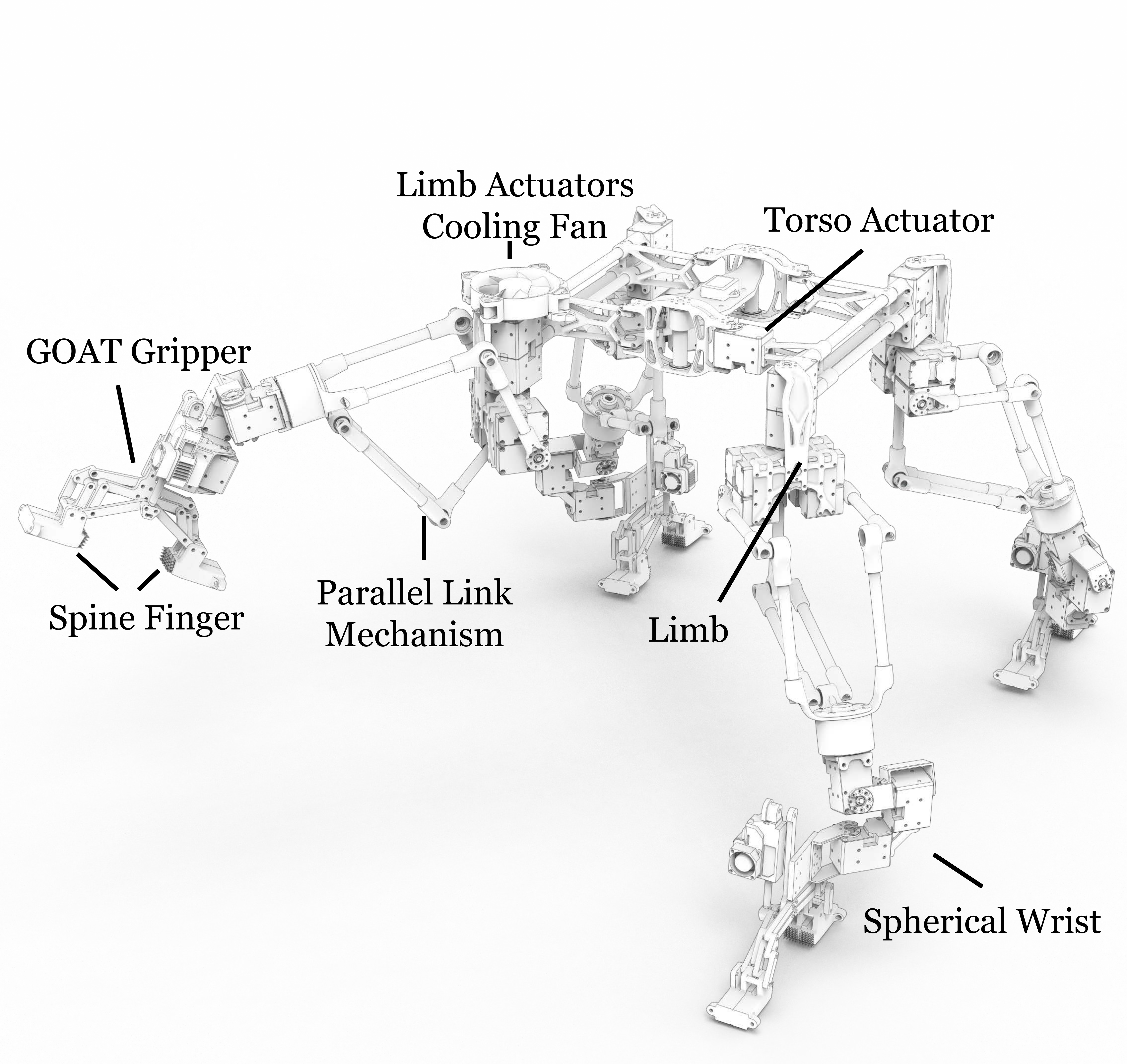}
     \caption{
     SCALER isometric rendering view. SCALER is a four-limbed climbing robot with six-DoF legs, each featuring an underactuated GOAT gripper with one actuator. The torso actuator adds 1-DoF translation motion in the body.
     \label{fig:SCALER}}
\end{figure}

\section{Mechanical Design\label{sec:hardware}}
SCALER, shown in \fig{fig:fig1} and \fig{fig:SCALER}, is designed to achieve tightly coupled loco-grasping capabilities and to meet the demands of power-intensive yet dexterous climbing tasks, while maintaining dynamic mobility. These capabilities are realized through \red{four key subsystems: (1) the torso mechanism (Section~\ref{sec:body_mechanisms}), (2) the parallel-serial link limb (Section~\ref{sec:limb_design}), (3) the gait designs (Section~\ref{sec:gait_design}), and (4) the mechanically adaptable GOAT gripper (Section~\ref{sec:goat}).}

\begin{figure}[t!]
    \centering
    \includegraphics[width=0.4\textwidth, trim={0cm 0cm 3cm 0cm}, ,clip]{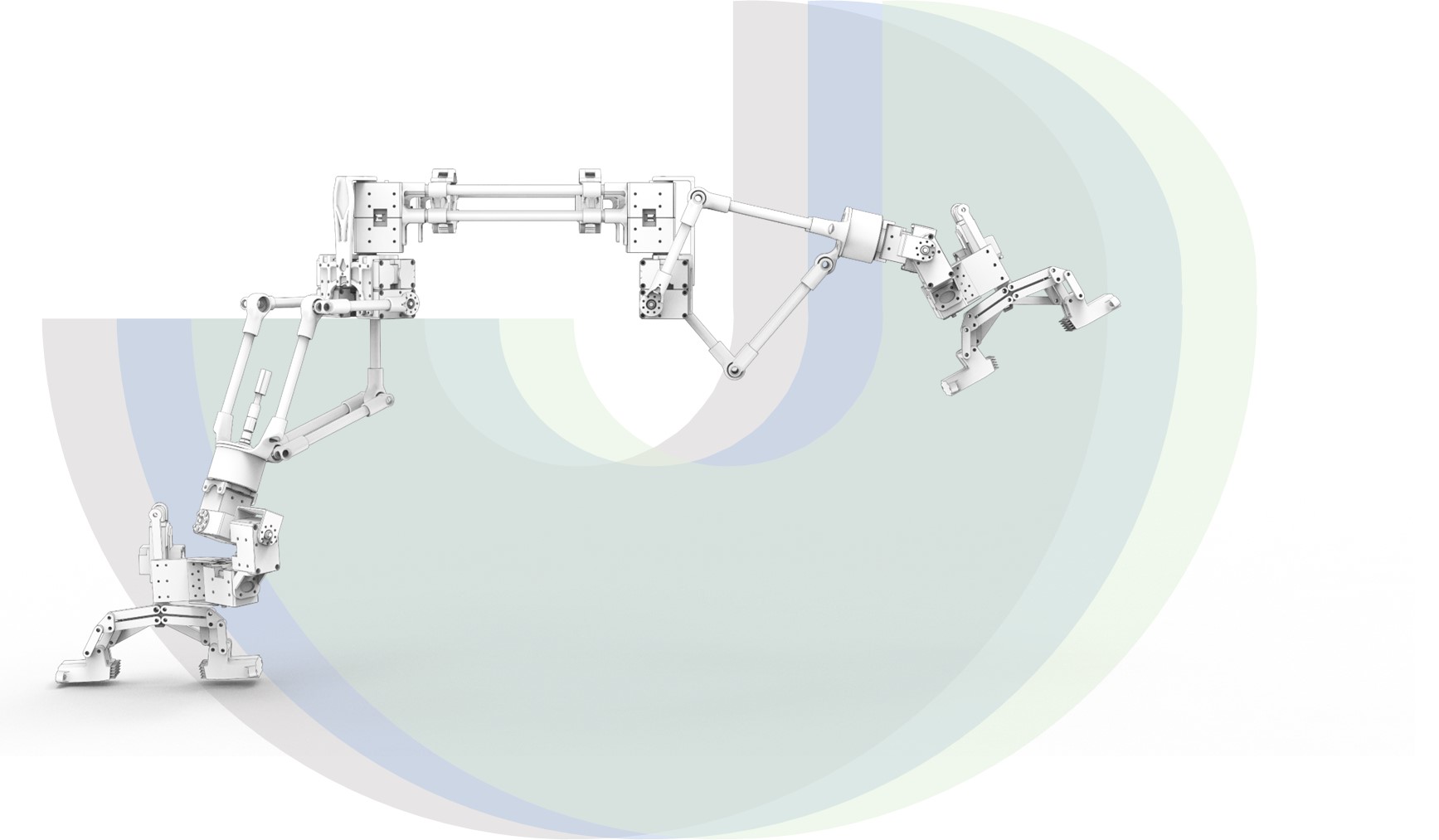}
     \caption{
     \red{SCALER 6-DoF limb workspace cross-section. The blue shadow area is the limb workspace with the nominal torso configuration, as shown in \ref{fig:one_leg_body_shift_gait}. The green and grey areas are when the torso is actuated $q_t=-45^\circ$ and $q_t=45^\circ$, respectively, as shown in \ref{fig:one_leg_body_shift_gait}. The workspace excludes unreachable areas due to parallel link collisions, such as elbows with the body.}
     \label{fig:leg_workspace}}
\end{figure}

\subsection{Torso Mechanism\label{sec:body_mechanisms}}
\begin{figure}[t!]
  \begin{subfigure}[t]{0.15\textwidth}
    \centering
    \includegraphics[width=\textwidth, height=3cm,trim={0cm 0cm 0cm 0cm},clip]{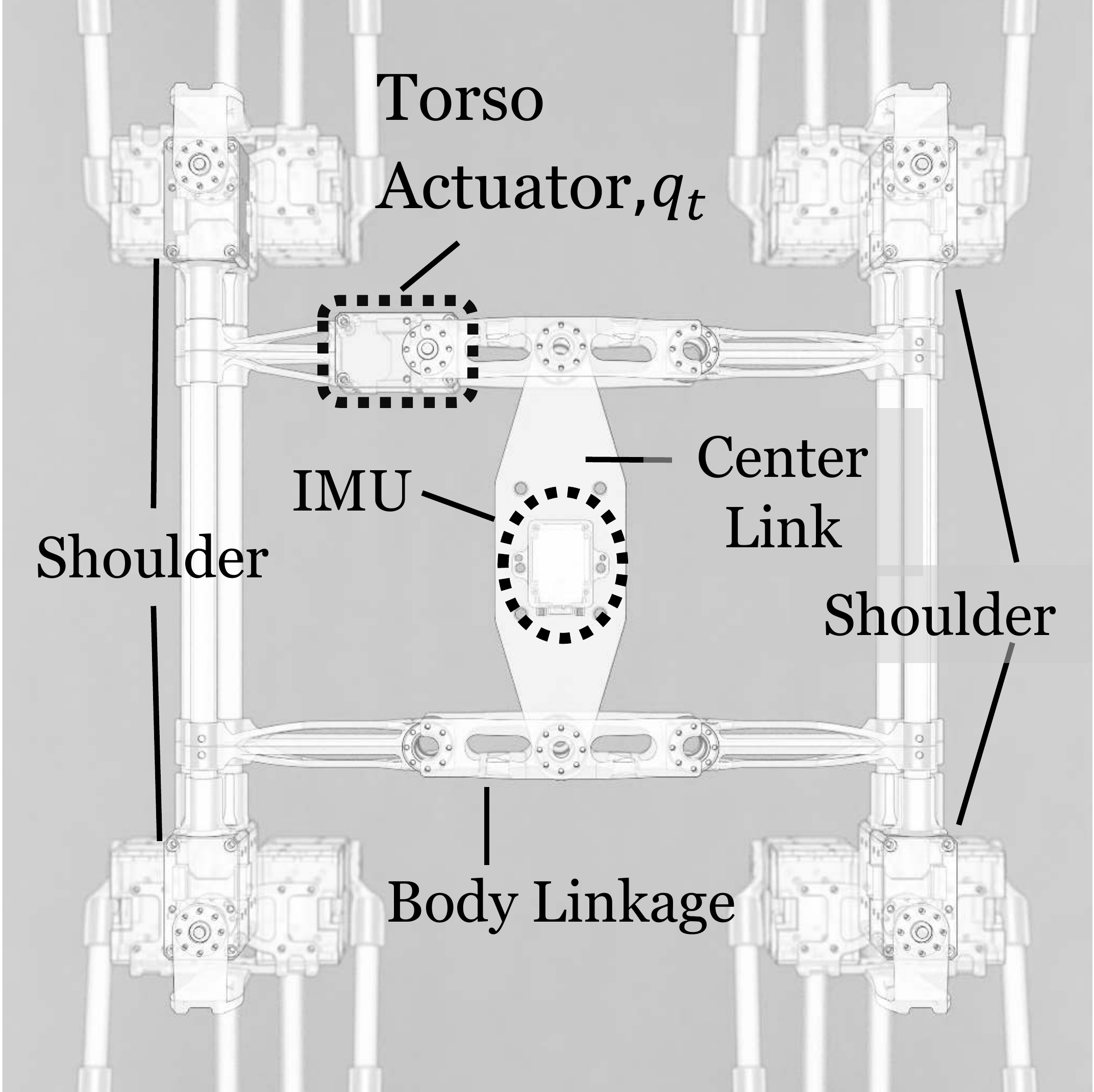}
    \caption{Torso actuator and the components at $q_t = \SI{0}{\degree}$.}
    \label{fig:body_nominal}
  \end{subfigure}
  \begin{subfigure}[t]{0.14\textwidth}
    \centering
    \includegraphics[width=\textwidth,height=3cm, trim={0cm 0cm 0cm 0cm},clip]{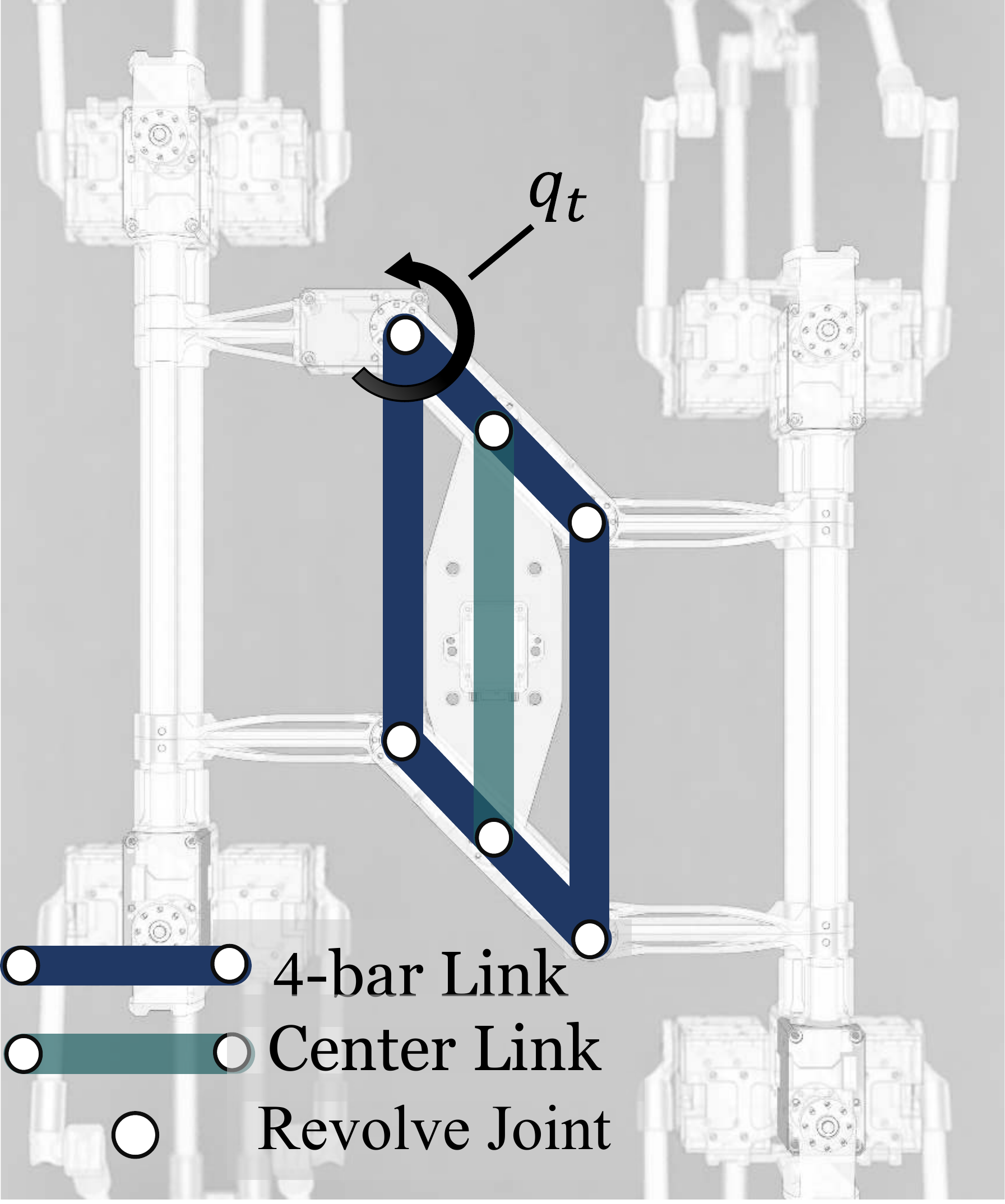}
    \caption{A torso state at $q_t = \SI{-45}{\degree}$. }
    \label{fig:body-b}
  \end{subfigure}
  \begin{subfigure}[t]{0.17\textwidth}
    \centering
    \includegraphics[width=\textwidth,height=3cm,trim={2.5cm 0.5cm 2.5cm 0cm},clip]{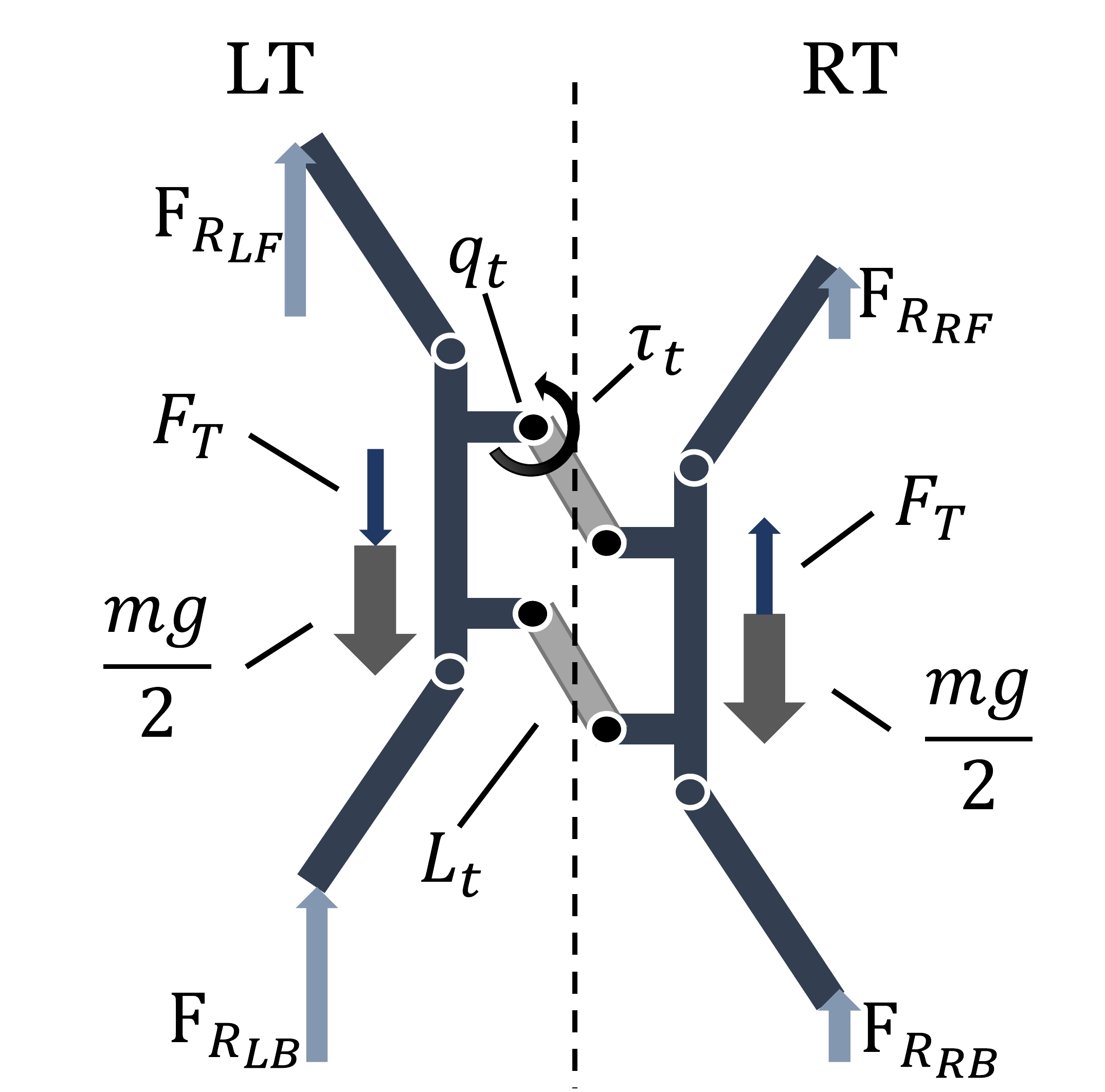}
     \caption{A free body diagram in 2D in the top.
     \label{fig:fbd_torso}}
  \end{subfigure}
  \caption{The renderings and schematic of SCALER body and torso mechanisms from the top view.}
  \label{fig:one_leg_body_shift_gait}
\end{figure}

The SCALER body employs a four-bar linkage mechanism driven by one actuator, as illustrated in \fig{fig:one_leg_body_shift_gait}, which provides advantages in terms of workspace and forces.
Conventional four-legged robots involving a rigid one-body and having a torso DoF are less commonly adopted due to the limited locomotion benefits compared to the complexity introduced. 
However, \red{climbing robots} can benefit from the additional DoF on the body, making the added complexity worthwhile. SCALER's torso includes a four-bar parallelogram linkage mechanism in \fig{fig:one_leg_body_shift_gait}, which enables the robot to achieve more climbing strategies. \fig{fig:body_nominal} \red{shows} the torso's nominal condition and components. 
\fig{fig:body-b} illustrates the 4-bar linkage torso configurations when the torso actuator is at $q_t = \SI{-45}{\degree}$. The center link shown in \fig{fig:body_nominal} houses an IMU, battery compartments, and a computer, which allow the center of mass to be closer to constant regardless of the torso state. 
This mechanism particularly benefits SCALER in two ways: 1) Workspace advantages, 2) Force advantages.

In terms of workspace advantages, this mechanism can grant the potential to shift the robot workspace on demand, as a human can stretch their arm, such as when trying to reach high objects. The limb workspace shift is visualized in \fig{fig:leg_workspace}. This is demonstrated in the bouldering climb experiment in Section~\ref{sec:bouldering}.
Furthermore, the translational torso motion can scale better to increase stride length than extending leg lengths, which would increase the torque requirements for the leg joints, and the limb workspace is spherical. 

Regarding force advantages, the torso actuator generates a thrust force and reduces the load to lift the torso by half when climbing. 
The 2D free body diagram is in \fig{fig:fbd_torso}, and the thrust force is calculated as $F_T = \tau_t / L_t$, where $F_T$ is the thrust force, $\tau_t$ is the torque due to the $q_t$ torso actuator, and the body linkage length, $L_t$. 
\red{We compared the case where the torso is actuated and free-moving. One side of the body is rigidly fixed. When the torso is free to move, a \SI{44}{\newton} force is required to pull up the half-body. With the torso actuated, this reduced to \SI{15}{\newton}, indicating \SI{29}{\newton} thrusting in hardware.}

\subsection{Limb\label{sec:limb_design}}
In this section, we discuss SCALER's  6-DoF limb design and the design principles behind it as follows:
\begin{enumerate}
    \item A redundantly actuated parallel linkage leg mechanism
    \item The leg compliance modeling
    \item Spherical wrist designs
\end{enumerate}
Each limb of SCALER acts as both a supporting leg and a grasping arm when climbing. Thus, the leg design must account for locomotion and grasping capabilities. 

\subsubsection{Redundantly Actuated Parallel-Serial Link Limb\label{sec:limb_design_parallel}}
\begin{figure}[t!]
    \centering
    \includegraphics[width=0.45\textwidth ,trim={0.5cm 0cm 0cm 0cm},clip]{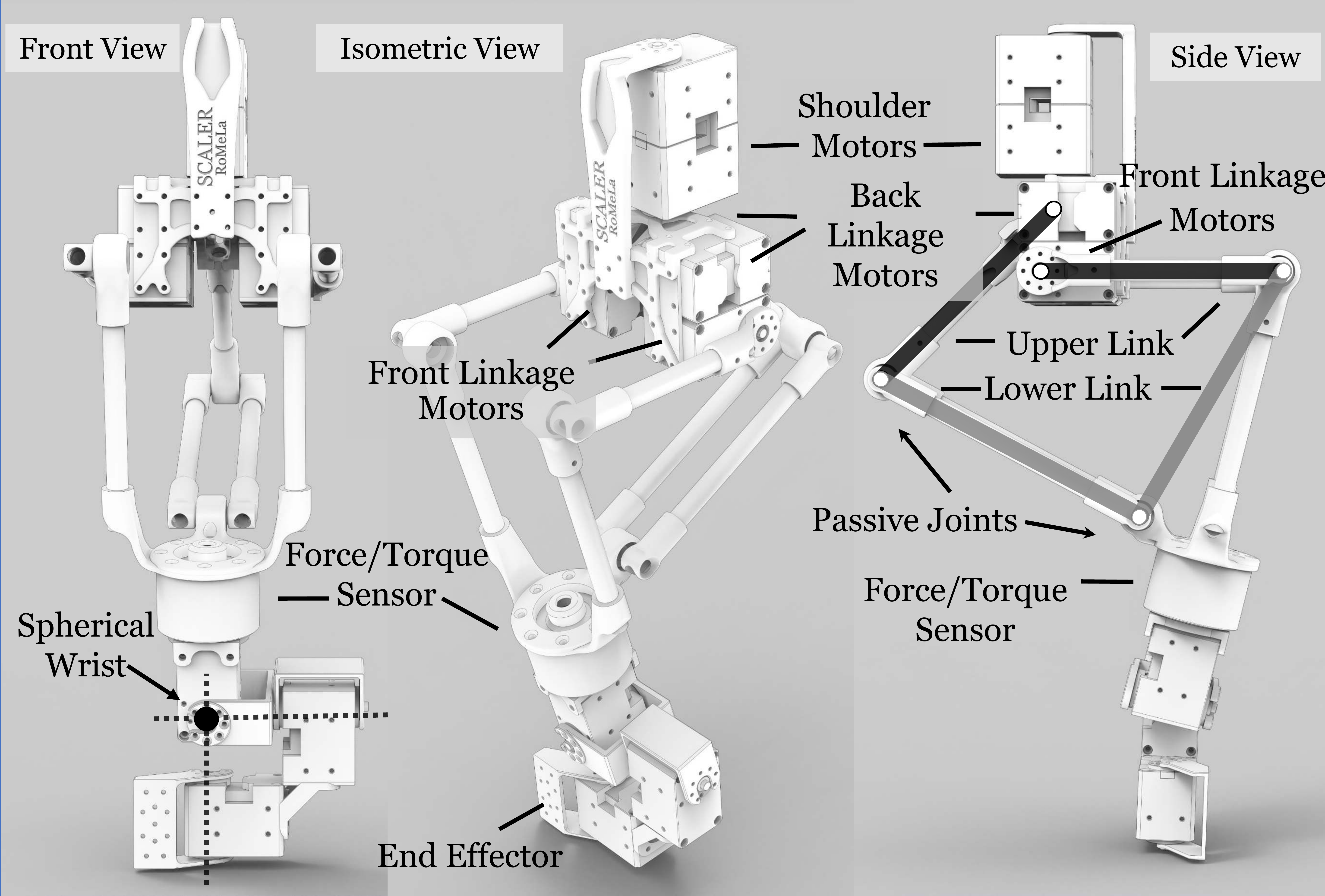}
     \caption{SCALER 6-DoF limb. The side view shows the five-bar linkage kinematic design. The upper and lower links are in black and grey, respectively. The shoulder and five-bar linkages are driven \redrev{by two motors each to increase continuous torque and reduce thermal rise as described in Section~\ref{sec:limb_design}}. The six-axis Force/Torque (F/T) sensor is installed at the end of the five-bar link, and the spherical wrist is on the sensor measurement face. 
     \label{fig:leg_design}}
\end{figure}

Climbing is a power-intensive form of locomotion, as robots must add potential energy and move against gravity. Hence, motor torque density and heat dissipation are critical. S
CALER has to support and lift itself and withstand moments due to gravity that would otherwise pull SCALER off the wall. However, a very high-gear ratio, as seen in LEMUR $3$ \cite{lemur3}, sacrifices motion speed significantly, whereas quasi-direct drive motors are less \red{suitable} due to less continuous torque density and overheating. 
SCALER limb consists of a five-bar linkage, which is serially combined with a shoulder joint and a spherical wrist to attain 6-DoF per leg as illustrated in \fig{fig:leg_design}. \red{SCALER limb workspace is visualized in \fig{fig:leg_workspace}.}
SCALER employs medium to high-gear ratio DC-servomotors, Dynamixel XM430-350 for all actuated joints.

The SCALER's five-bar parallel link mechanism utilizes two redundantly actuated joints to realize 2-DoF motions aligned in the climbing direction. Having two motors per actuated joint doubles the torque while increasing heat dissipation.
The symmetric five-bar design is known to be mechanically superior in terms of proprioceptive sensitivity, force production, and thermal cost of force among serial and two different five-bar linkage leg designs in \cite{five_bar_leg}. Although this choice of leg mechanism is ideal for power-intensive climbing operations, the parallel mechanism suffers from singularities in the middle of the workspace, such as where both front and back passive joints shown in \fig{fig:leg_design} are at the same axis. In SCALER, this condition only happens outside the regular operation (e.g., when the end effector is inside the body) since the back linkage in \fig{fig:leg_design} is marginally shorter. The passive elbow joints can collide with the environment or obstacles. However, it is uncommon for SCALER's climbing task since SCALER needs \redrev{to raise the gripper higher, such as in \fig{fig:leg_workspace}.}

In addition to the reduced thermal cost of force with the parallel mechanism, the coupled two motors at each joint distribute heat sources, keeping the motor temperature optimal and avoiding thermal throttling or damage, although this redundant motor configuration increases wiring and is more expensive than employing one larger actuator.
Such thermal considerations are vital since the power demands of free-climbing are continuous. 
The detachable fan above the leg cools the shoulder-leg actuators in \fig{fig:SCALER}. 

We compared the thermal performance of the SCALER redundant motor configuration (two Dynamixel XM430-350) and a single larger motor (Dynamixel XM540-150). We applied the same load and power \redrev{using a dedicated test rig with a mass attached at the end of the link. The test rig was thermally isolated from the ground and passively cooled.} 
XM430-350 was \SI{0.6}{\ampere} per motor, and XM540-150 was \SI{1.2}{\ampere} at \SI{13.8}{\volt}. The output torque difference was \SI{5}{\percent}. The SCALER's motor configuration case saturated to \SI{48.5}{\degreeCelsius} with a time constant \SI{9.8}{\minute}. Over \SI{1}{\hour} of testing, the SCALER motor configuration did not overheat. 
On the other hand, the single larger motor temperature quickly rose to \SI{50}{\degreeCelsius} after \SI{16}{\minute}. 
It was predicted to saturate at \SI{57}{\degreeCelsius}, but it would have damaged the motor \cite{dynamixel_overheat}. 
This indicates that having redundant motors per joint, as in SCALER, is thermally advantageous.
 
The weight of SCALER's limb is distributed on both ends: the body and the wrist. The intermediate links are structured with carbon fiber tubes.
When the wrist and grippers are not installed, the leg consists of relatively low inertia, improving the swing leg dynamics.

\begin{figure}[t!]
    \centering
    \includegraphics[width=0.45\textwidth ,clip]{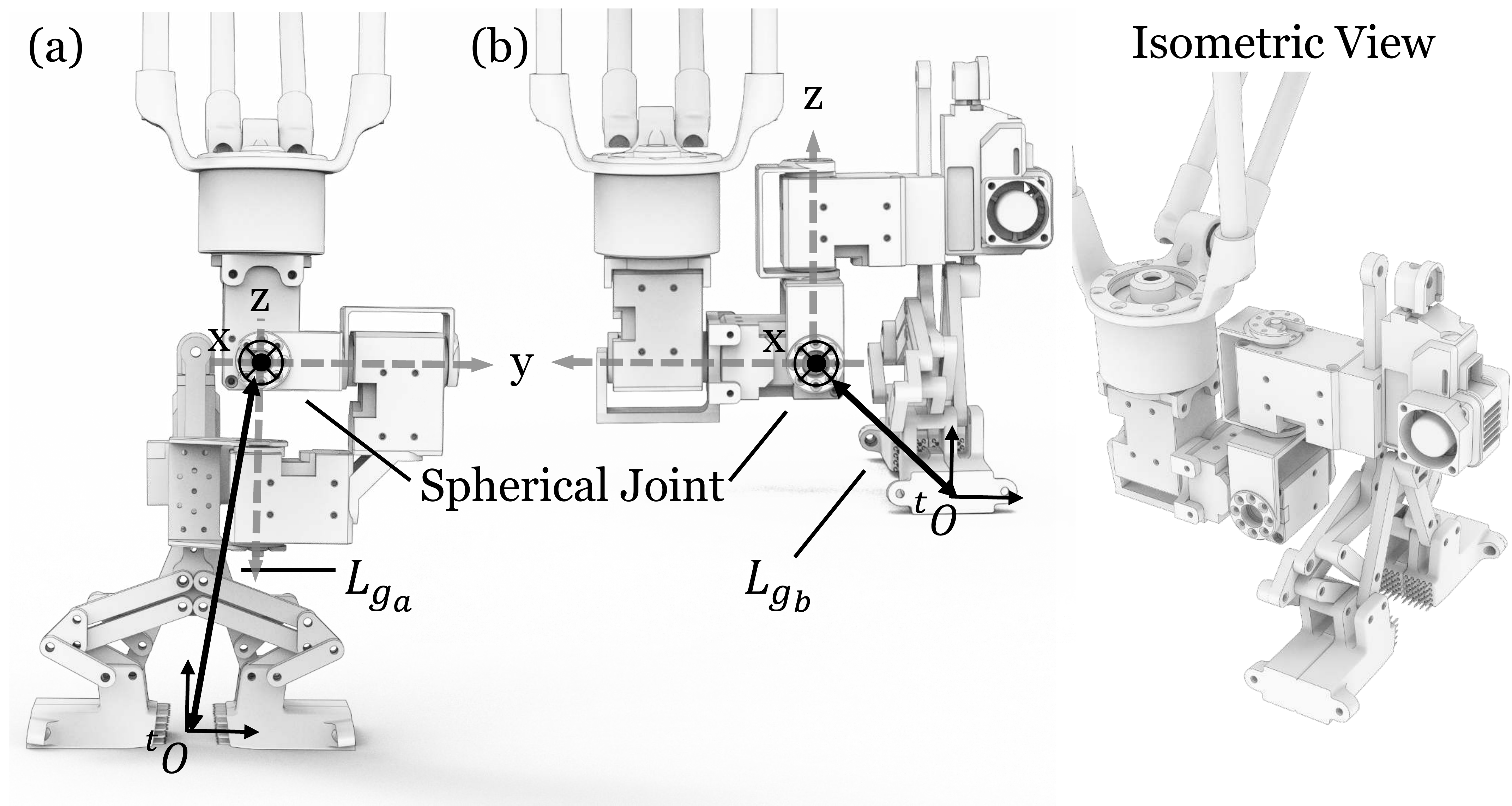}
     \caption{SCALER 6-DoF spherical wrist variations. (a) Longer offset lengths can avoid gripper collision and have greater joint ranges. (b) Shorter and compact wrist, but smaller effective range of motions. Their joint ranges and leg-to-wrist length ratios are in Table~\ref{tb:wrist}.
     \label{fig:wrist}}
\end{figure}

\begin{table}[t!]
\begin{threeparttable}
\caption{Wrist configurations and parameters. 
\label{tb:wrist}
}
\begin{tabular}{ccccc}
\hline
\begin{tabular}[c]{@{}c@{}}Wrist Design\end{tabular} & $\zeta$ & \rotateaxisleft $x$ & \rotateaxisleft $y$ & \rotateaxisleft $z$ \\ \hline
\rowcolor[HTML]{EFEFEF} 
\fig{fig:wrist}a & $30.0$ \% & $[-25, 90]^\circ$ & $[-110, 110]^\circ$ & $[-180, 90]^\circ$\\ \hline
\fig{fig:wrist}b & $15.2$ \% & $[-10, 90]^\circ$ & $[-120, 120]^\circ$ & $[-90, 90]^\circ$
\end{tabular}
\begin{tablenotes}
      \footnotesize
      \item
      The ratio is defined as $\zeta = \frac{L_{g_j}}{L_l}  \times 100$, where $L_{g_j}, j \in \{a, b\}$ and $L_l$ are the wrist and the leg Euclidean length from the spherical joint and from the leg back linkage motor joint in \fig{fig:leg_design} to the fingertip frame, $^tO$ in \fig{fig:wrist}.
    \end{tablenotes}
\end{threeparttable}
\end{table}


\begin{table*}[t!]
 \centering
 \caption{SCALER configuration list.
\label{tb:configuration}
}
\begin{tabular}{cccccccc}
\hline
Configuration & \begin{tabular}[c]{@{}c@{}}Limb\\ DoF\end{tabular} & Wrist & \begin{tabular}[c]{@{}c@{}}GOAT Actuator\\ (Section \ref{sec:gripper_module})\end{tabular} & \begin{tabular}[c]{@{}c@{}}GOAT Finger\\ (Section \ref{sec:goat_finger})\end{tabular} & Applications & Fig. & Experiments Section\\ \hline
Walking & 3 & N/A & N/A & N/A & Ground Locomotion & \ref{fig:fig1}b \ref{fig:scaler-m}a & \ref{sec:ground_trot}, \ref{ground_payload}  \\
\rowcolor[HTML]{EFEFEF} 
Spine Climbing & 6 & \fig{fig:wrist}a & DC Linear & Spine Cell & Climbing on Rough Surfaces & \ref{fig:fig1}a,c, \ref{fig:SCALER} & \ref{sec:inverted}, \ref{sec:vertical_payload}, \ref{sec:bouldering} \\
\begin{tabular}[c]{@{}c@{}}Dynamic/Multi-Modal\\ Climbing\end{tabular} & 6 & \fig{fig:wrist}b & Pneumatic & \begin{tabular}[c]{@{}c@{}}C-Shaped\\ Dry Adhesive\end{tabular} & \begin{tabular}[c]{@{}c@{}}Dynamic  Multi-Modal Climbing\\ Slippery Terrain\end{tabular} & \ref{fig:fig1}d & \ref{sec:dynamic_trot}, \ref{sec:multi_modal_climb}, \ref{sec:whole_body} \\
\rowcolor[HTML]{EFEFEF} 
Bi-Manipulator & 6 & \fig{fig:wrist}a & N/A & N/A & Fixed-Base Manipulation & \ref{fig:scaler-m}b & VI-C in \cite{sim2real} \\ \hline
\end{tabular}
\end{table*}

\subsubsection{Limb Compliance Modeling\label{sec:compliance_model}}

The Virtual Joint Method (VJM) modeling \cite{gosselin1990stiffness} is used to approximate limb stiffness. This model is verified through experiments in Section \ref{sec:compliance_model_test}. 
The model is based on lumped modeling and assumes joint stiffness as linear springs for small angles with a rigid link \cite{gosselin1990stiffness}.
The VJM stiffness model \red{ is necessary for stiffness force control and estimation} in climbing tasks as demonstrated in \cite{multi-surface}. 
\red{The} model considers the shoulder and the parallel mechanisms, and the compliance in the metal wrist and gripper linkages is negligible compared to these parts. 

The stiffness matrix for SCALER's parallel-serial leg is formulated as follows:
\begin{equation}
K = \left(\frac{1}{k_0}+\frac{1}{k_1+k_2}\right)^{-1}J^\top J \label{eq:stiffness}
\end{equation}
Where $K$ is a stiffness matrix in the Cartesian coordinate, $k_0, k_1, k_2$ are joint stiffness at the shoulder and two parallel linkage drive joints shown in \fig{fig:leg_design}, and $J$ is a $\mathbb{R}^{3\times 3}$ Jacobian matrix for the leg excluding spherical joints. 
The condition number, $\kappa$ indicates relative stiffness at a specific joint configuration, and it is obtained from \eqref{eq:condition_number}:
\begin{equation}
    \frac{1}{\kappa}=\sqrt{\frac{\lambda_{\min }}{\lambda_{\max }}} \label{eq:condition_number}
\end{equation}
$\lambda_{\min }$ and $\lambda_{\max}$ are the minimum and maximum eigenvalues of the stiffness matrix, $K$. 
The stiffness map obtained from the hardware is in Section \ref{sec:compliance_model_test}. 

\subsubsection{Spherical Wrist}
The spherical wrist design contributes to the robot's workspace since free-climbing requires \red{dexterity}.
SCALER employs spherical wrists as shown in \fig{fig:leg_design} and \fig{fig:wrist}, which have uniform end-effector offsets for all rotations and simpler inverse kinematics.
Two SCALER wrist designs are shown in \fig{fig:wrist} and detailed in Table \ref{tb:wrist}. While the added wrist length helps prevent wrist collisions with the environment, it also limits reachability and raises torque needs at leg joints.
The wrist in \fig{fig:wrist}a used in \cite{yusuke_scaler_2022} occupies \SI{30}{\percent} of the limb length, whereas the wrist \fig{fig:wrist}b is $15$ \%. 

\begin{figure}[t]
    \centering
    \includegraphics[width=0.45\textwidth ,clip]{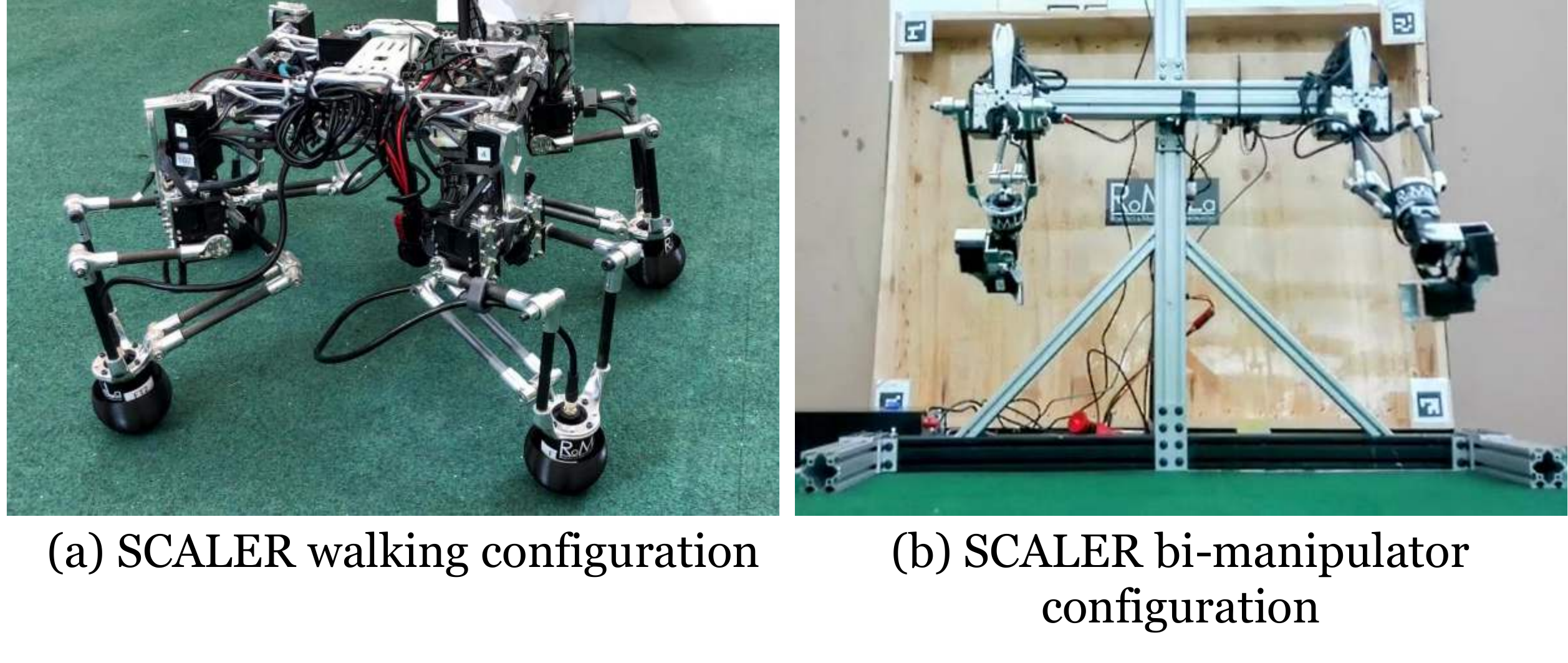}
     \caption{\red{Different configurations of SCALER.}
     \label{fig:scaler-m}}
\end{figure}

\subsection{Modularity}
SCALER design incorporates modularity, enhancing its applications for different objectives and research. This section explores various SCALER configurations and modules.
\subsubsection{SCALER Configurations}
Modularity provides both engineering benefits and additional capabilities beyond climbing, as shown in \fig{fig:scaler-m}.
Primary SCALER configurations are listed in Table \ref{tb:configuration}.
SCALER's climbing configurations include four 6-DoF limbs and a 1-DoF torso, resulting in 25-DoF. 
Each limb includes a two-fingered underactuated 2-DoF gripper driven by one actuator, as detailed in Section~\ref{sec:goat}. 

The walking configuration in \fig{fig:scaler-m}a has 3-DoF per leg by replacing the spherical wrist with a flexible semi-spherical foot cover to protect the force/torque (F/T) sensor.
The walking format benefits from the minimal inertia design of SCALER's parallel linkage leg.
A bimanual manipulator configuration uses two SCALER limbs attached to a fixed base as shown in \fig{fig:scaler-m}b and is experimented with in \cite{sim2real}.

\subsubsection{Untethered Operation Modules}
For untethered operations, SCALER is equipped with additional computing and power modules. The PC and battery can be latched on the top and bottom of the center link in \fig{fig:one_leg_body_shift_gait}, respectively. 
A Jetson Orin unit weighing \SI{0.97}{\kilogram} is employed for onboard vision and multi-depth camera processing, whereas a NUC6i5SYK unit weighing \SI{0.39}{\kilogram} is used for non-GPU intensive applications. 
Power modules comprise a MaxAmp 4S \SI{14.8}{\volt}, \SI{8}{} or \SI{12}{\ampere\hour} Li-Po battery, paired with DC-DC converters with a wireless emergency stop. 
The gripper is connected via M12 cables, and the pneumatic system described in Section \ref{sec:gripper_module} requires additional CO$_2$ gas tanks with a regulator on the body.

 \begin{figure}[t!]
 \centering
\includegraphics[width=0.85\linewidth,trim={0cm 0cm 0cm 0cm}, clip]{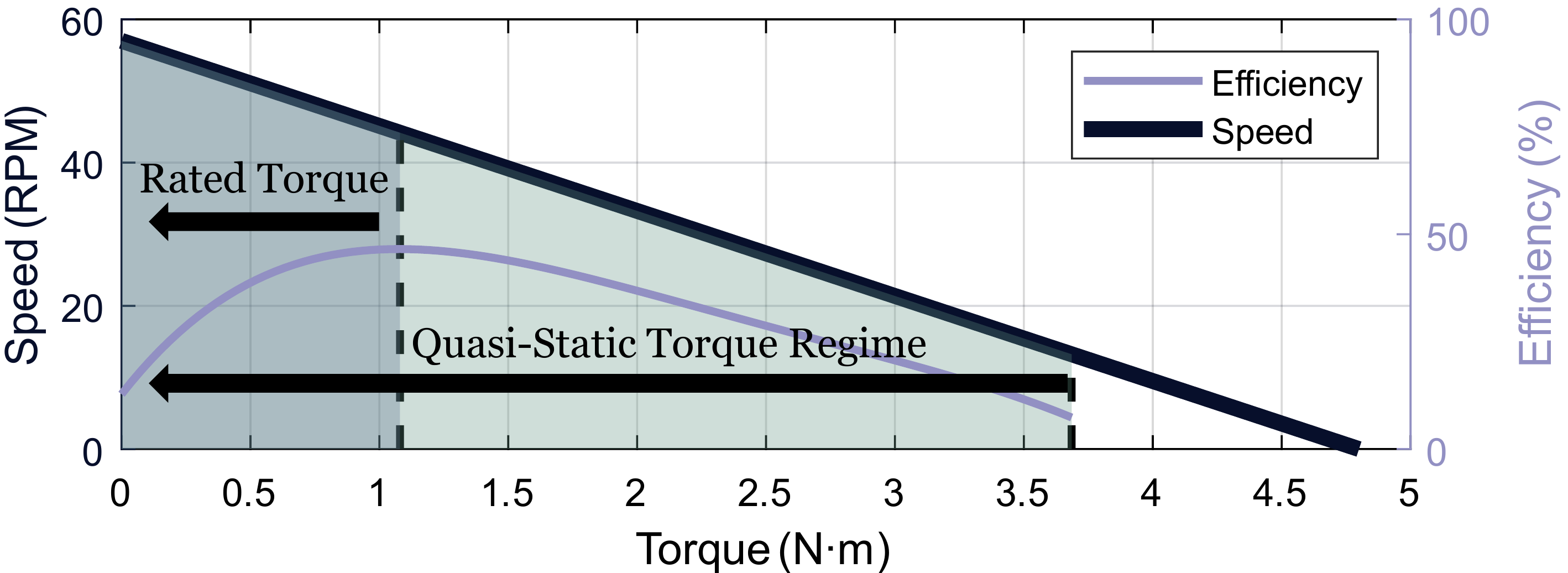}
     \caption{\red{Torque-speed curve for the servo motor used in SCALER at \SI{14.8}{\volt}.}
     }\label{fig:motor_curve}
\end{figure} 

\subsection{SCALER Gait\label{sec:gait_design}}
In this section, we introduce two new climbing gaits utilizing SCALER's torso mechanism and limb \red{stiffness}: \redrev{(1) SKATE Gait and (2) Modified Trot Gait.}
These gaits benefit SCALER with a mechanical advantage. \red{The SKATE gait ensures quasi-static climbing by having two limbs always stationary, whereas the modified trot gait is meant for tasks where force control is critical, such as dynamic climbing.}

\subsubsection{SKATE Gait\label{skate_gait}}
\begin{figure}[t!]
 \centering
             \begin{subfigure}[t]{0.26\textwidth}
\includegraphics[height=1.5cm,width=\textwidth, trim={0cm 0cm 0cm 0cm}, clip]{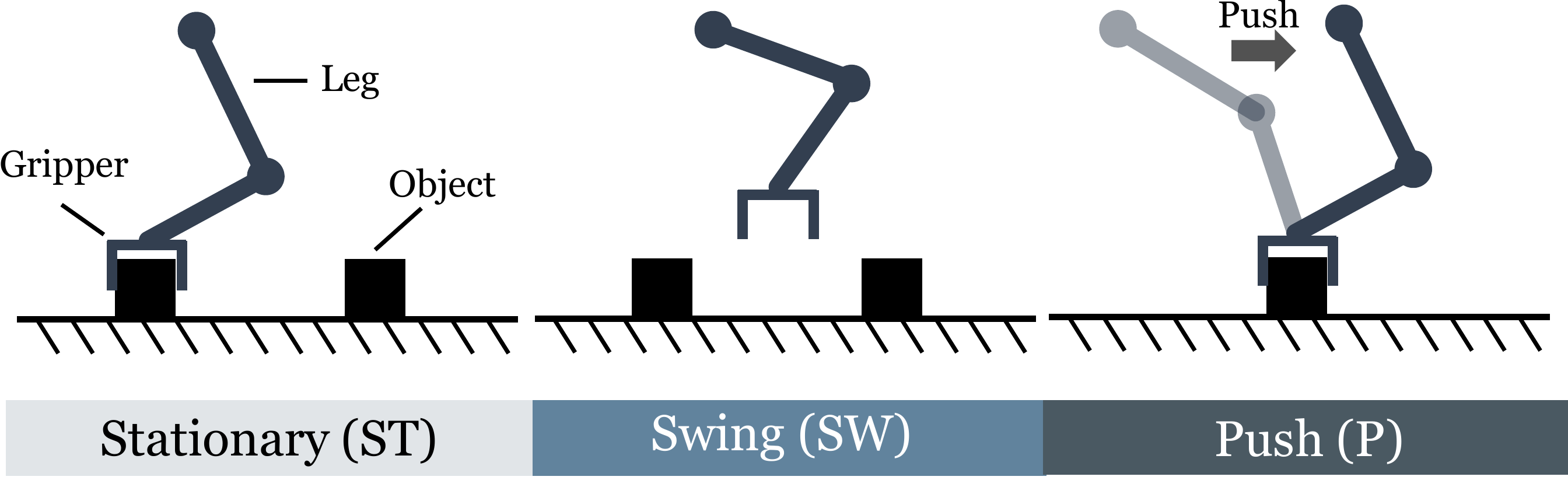}
\caption{Leg motion legend. When stationary, the leg has no movement. In the swing, the leg swings, and the gripper is open. In the push, the leg conducts a push motion, e.g., pushing the body forward.\label{fig:gait_leg_legend}}
    \end{subfigure}
\hfill
             \begin{subfigure}[t]{0.22\textwidth}
\includegraphics[height=1.5cm,width=\textwidth, trim={0cm 0cm 0cm 0cm}, clip]{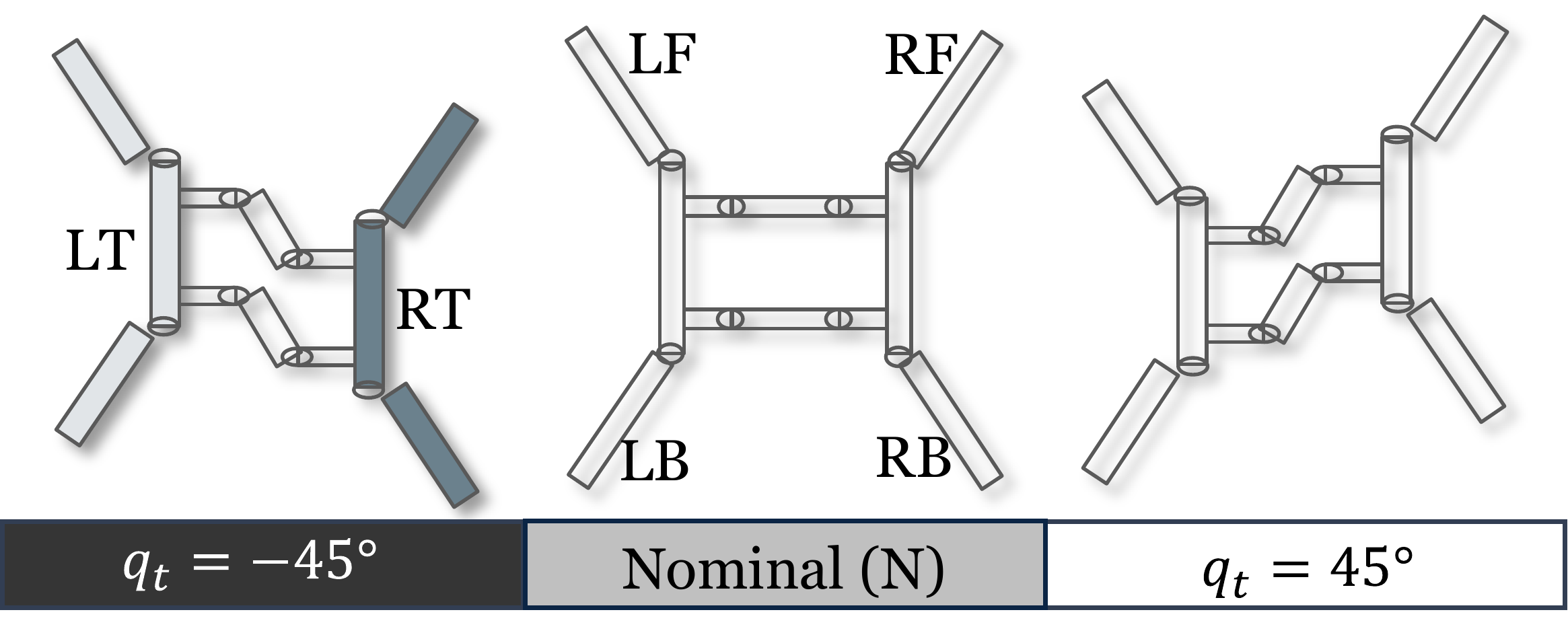}
\caption{Torso state legend. SCALER sketch from the top view. At Nominal, the torso actuator angle, $q_t$, is \SI{0}{\degree}. Gradation represents the state \red{and $q_t$} transition from \SIrange{-45}{45}{\degree}, or vice versa.
\label{fig:gait_body_legend}}
    \end{subfigure}
     \caption{Leg and torso state legends for gait schedules in \fig{fig:skate_compare} and \fig{fig:modified_trot_compare}.  
     }\label{fig:gait_legend}
\end{figure}

\begin{figure}[t!]
 \centering
          \begin{subfigure}[t]{0.26\textwidth}
    \centering
\includegraphics[width=\textwidth,trim={0cm 0cm 0cm 0cm}, clip]{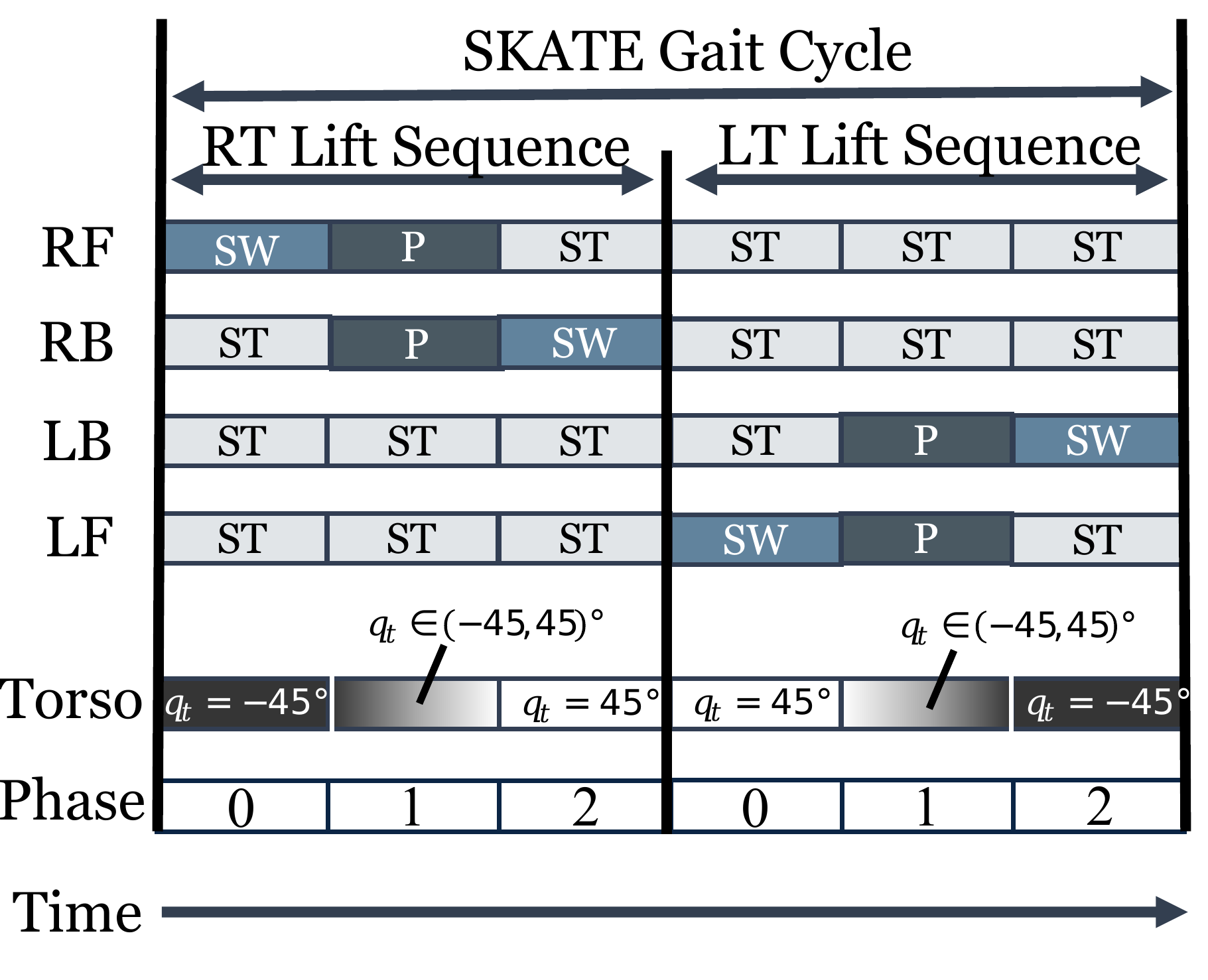}
\caption{SKATE gait.\label{fig:gait_skate}}
    \end{subfigure}
    \hfill
     \begin{subfigure}[t]{0.22\textwidth}
         \centering
    \includegraphics[width=\textwidth,trim={0cm 0cm 0cm 0cm}, clip]{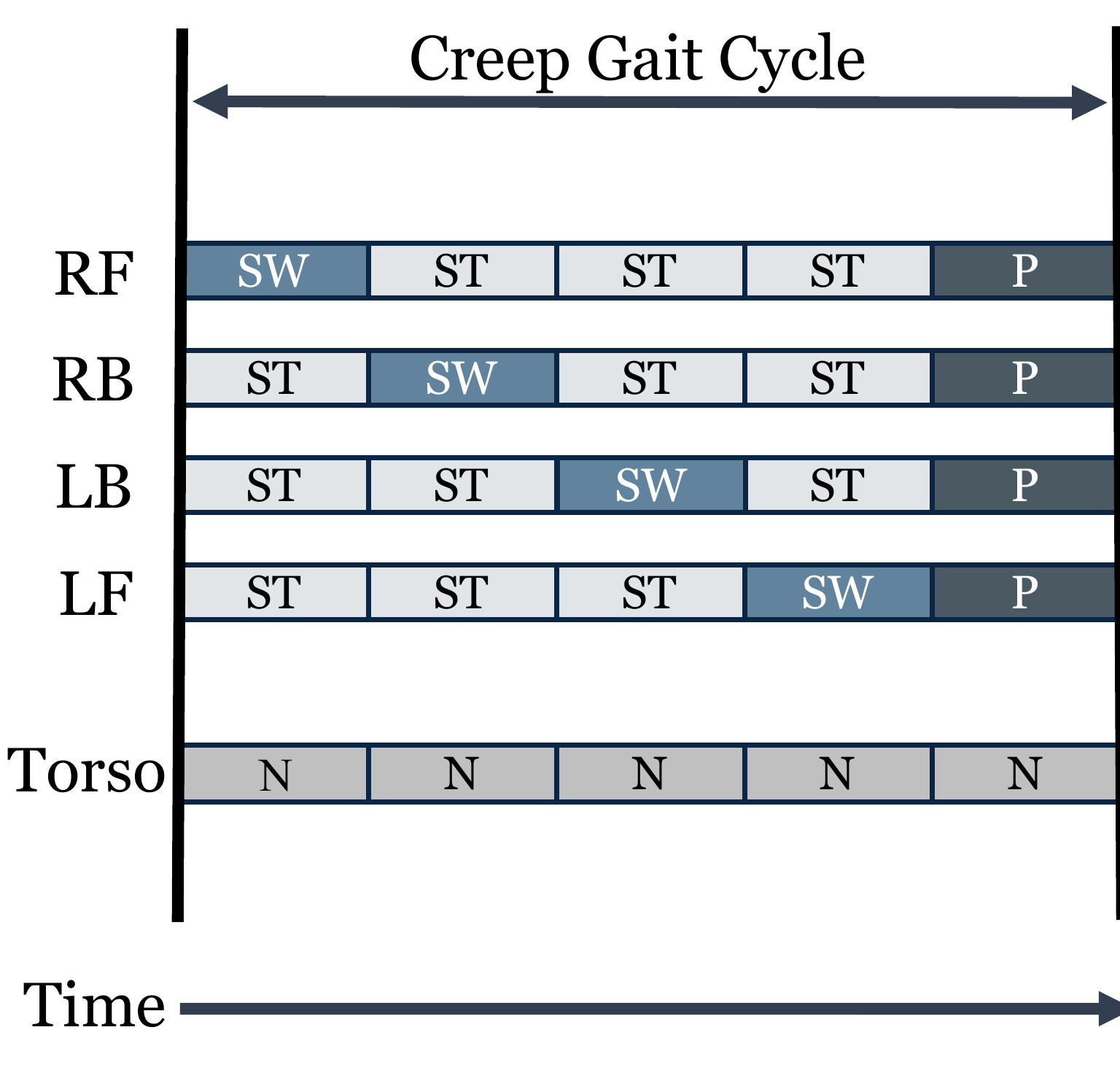}
    \caption{Creep gait \cite{mcghee1968stability}. \label{fig:gait_on_leg}}
     \end{subfigure}
     \caption{SKATE gait and creep gait schedules comparisons. The SKATE gait phases are defined in Section \ref{skate_gait}. Leg and torso state legends are in \fig{fig:gait_legend}.}\label{fig:skate_compare}
\end{figure}

SCALER's torso mechanism allows a unique gait called the Shifting Kinematics Adaptive Torso Extension (SKATE) gait. 
Our SKATE gait uses the torso actuation to maximize the advantage of the thrust force as described in Section \ref{sec:body_mechanisms}. \red{The main benefit of this SKATE gait is that half of the body is always stationary, which utilizes the quasi-static regime of the motor torque as visualized in \fig{fig:motor_curve}. This is a common approach in high gear ratio actuators \cite{eels_iros} at the cost of efficiency as in \fig{fig:motor_curve}.}
In the SKATE gait, only half of the body moves forward for one motion sequence, such as the Right Torso (RT), and then the Left Torso (LT) moves forward. The \red{gait legend and} SKATE gait schedule \red{are} shown in \red{\fig{fig:gait_legend} and} \fig{fig:gait_skate}.
The RT lift sequence uses the Right Front (RF) and the Right Back (RB) legs as follows:
\begin{enumerate}
    \item \textbf{Phase 0}: Swing the RF leg.
    \item \textbf{Phase 1}: Lift RT with RF, RB, and the torso actuator.
    \item \textbf{Phase 2}: Swing the RB leg.
\end{enumerate}
While RT is in this lift sequence, the LT, the other side of the body, the Left Front (LF), and the Left Back (LB) legs are stationary, meaning they are not in motion. During \textbf{Phase 1}, the torso actuator rotates from $q_t = \SI{-45}{\degree}$ to $q_t = \SI{45}{\degree}$. 
After \textbf{Phase 2}, LT enters the lift sequence, which repeats the same pattern. RT is stationary instead.
\begin{enumerate}
    \item \textbf{Phase 0}: Swing the LF leg.
    \item \textbf{Phase 1}: Lift LT with LF, LB, and the torso actuator.
    \item \textbf{Phase 2}: Swing the LB leg.
\end{enumerate}

Thus, SCALER alternately lifts RT and LT over the cycle of the SKATE gait.
The leg actuators can tolerate higher holding torque than in continuous motion \red{ as in \fig{fig:motor_curve}}. Furthermore, the \red{thrust force from the torso actuator assists the swing legs} as illustrated in \fig{fig:fbd_torso}. 
\redrev{Compared to the static creep gait in \fig{fig:gait_on_leg} with duty factor $>0.75$, the SKATE gait has two half-body lift phases, and at least two legs are always stationary, utilizing the quasi-static torque regime in \fig{fig:motor_curve}}.  


\subsubsection{Modified Trot Gait Sequence\label{sec:modified_trot}}
\begin{figure}[t!]
 \centering
          \begin{subfigure}{0.26\textwidth}
\includegraphics[width=\textwidth,trim={0cm 0cm 0cm 0cm}, clip]{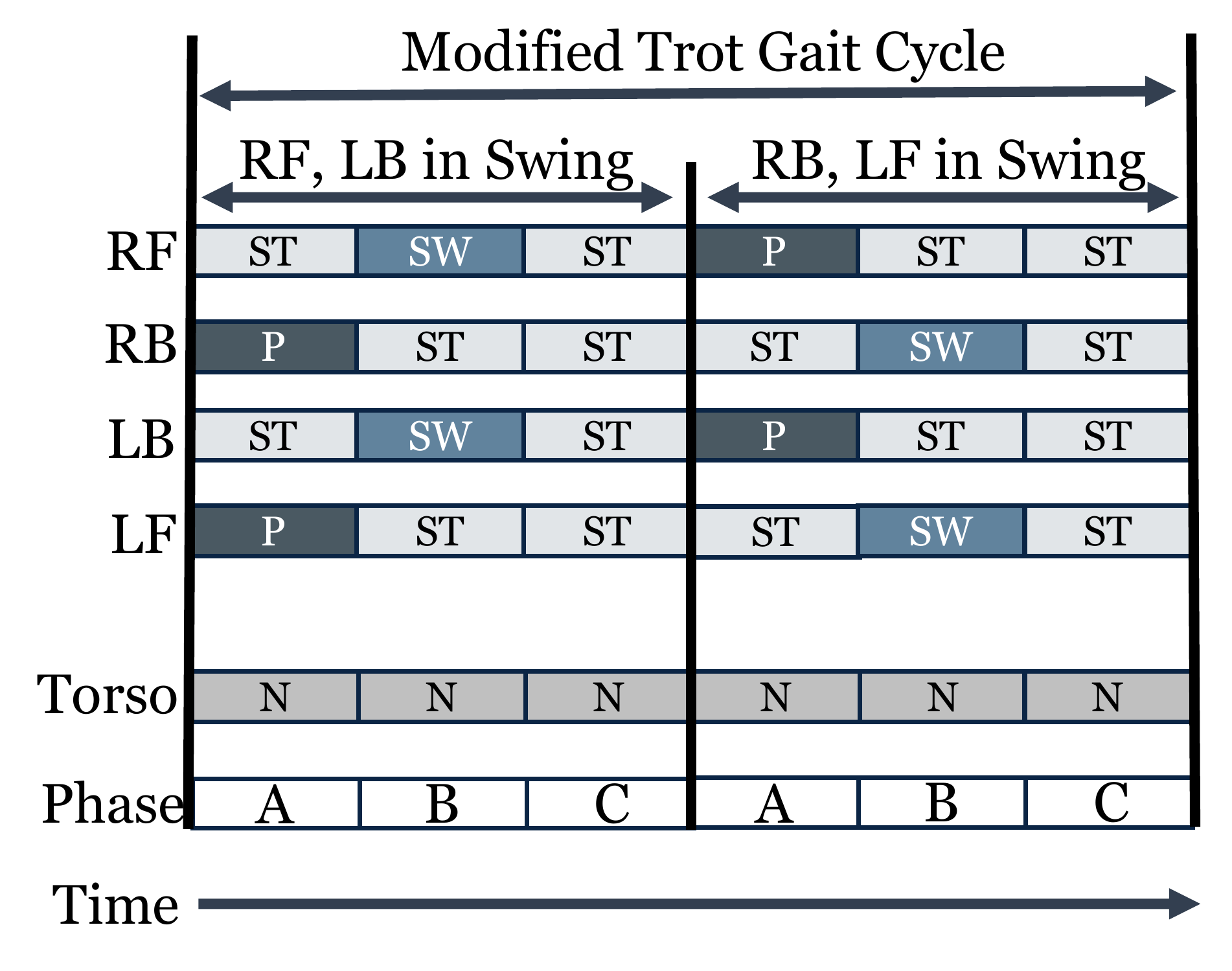}
\caption{Modified trot gait.\label{fig:gait_modified_trot}}
    \end{subfigure}
    \hfill
\begin{subfigure}{0.2\textwidth}
\includegraphics[width=\textwidth,trim={0cm 0cm 0cm 0cm}, clip]{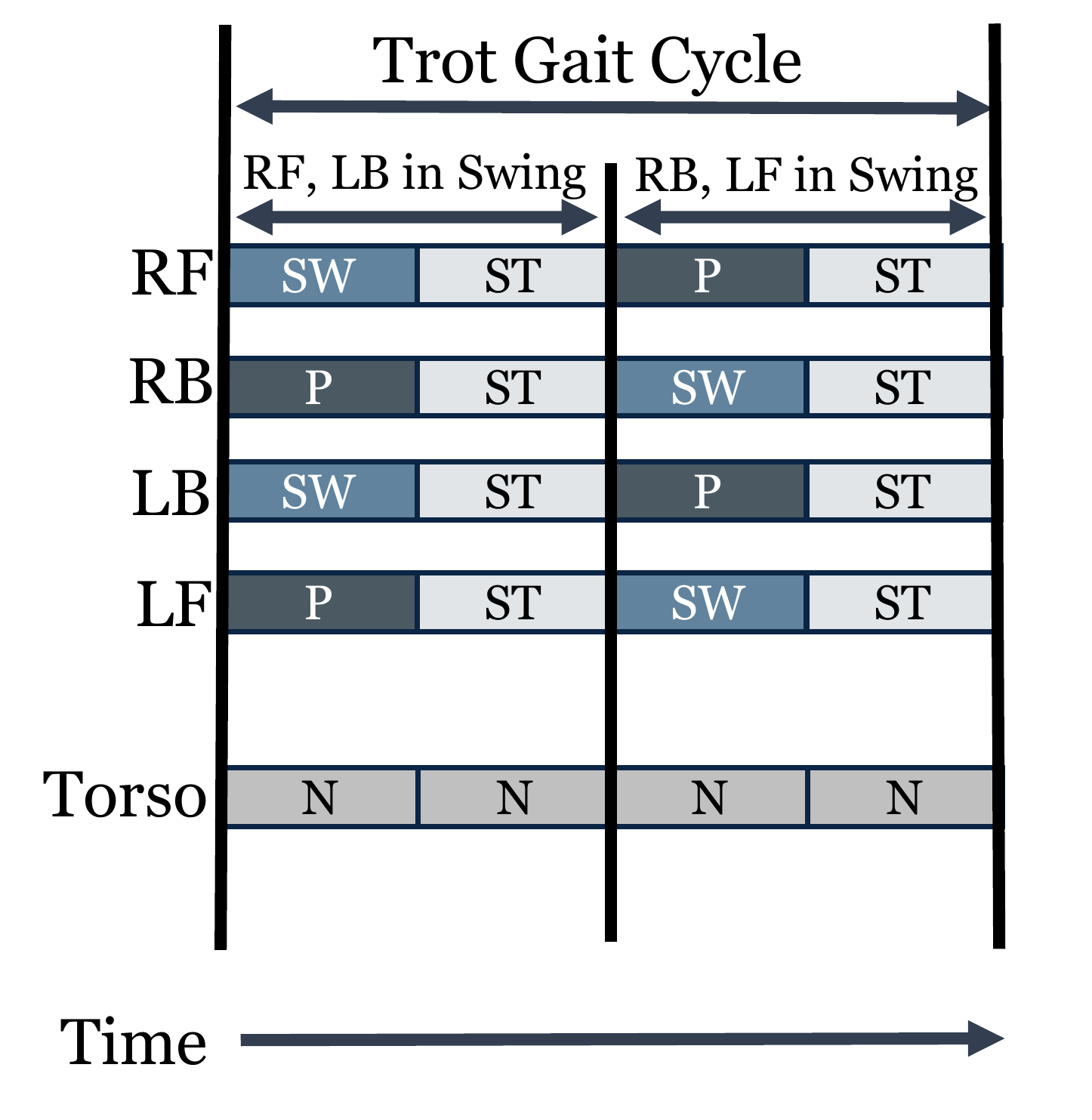}
\caption{Trot gait with a 4-leg stance \cite{509168}.\label{fig:gait_trot}}
    \end{subfigure}    
     \caption{Modified trot gait and trot gait with a 4-leg stance schedules comparisons. The modified trot gait phases are defined in Section \ref{sec:modified_trot}. Leg and torso state legends are in \fig{fig:gait_legend}.}\label{fig:modified_trot_compare}
\end{figure}

Dynamic climbing presents unique challenges, notably the sag-down effect. This effect arises from a combination of gravity, angular slips around the gripping points, and inherent mechanical compliance.
To address this problem, we propose a modified trot gait sequence with a dedicated stiffness force control pushing phase, and the gait schedule is shown in \fig{fig:gait_modified_trot}.

The sequence unfolds over three phases:
\begin{enumerate}
\item \textbf{Phase A:} A pushing action switches the support force from the RF/LB to the RB/LF leg \red{pair}. After this phase, the RB/LF \red{leg pair primarily supports weight}.
\item \textbf{Phase B:} The RF/LB grippers open and swing forward. 
\item \textbf{Phase C:} The RF/LB grippers re-engage, resulting in a four-leg stance.
\end{enumerate}


Compared to a \red{conventional} trot gait with a 4-leg stance presented in \fig{fig:gait_trot}, \red{the benefits of using this modified trot gait in climbing are by adding a dedicated push phase, \textbf{Phase A} in \fig{fig:gait_modified_trot}. It minimizes the change in supporting forces when the diagonal leg pair switches at the beginning of \textbf{Phase B}. This approach ensures a smooth and controlled transfer of supporting forces between trot leg pairs and reduces abrupt changes in support force distribution.}
A four-leg stance enhances gait stability, as argued in \cite{509168}. 
We demonstrate this modified trot gait in Section \ref{sec:dynamic_trot} and evaluate its performance in Section \ref{sec:modified_trot_test}, along with the associated mechanical stiffness in Section \ref{sec:compliance_model_test}.

\section{GOAT Gripper\label{sec:goat}}
\begin{figure}[t!]
  \begin{subfigure}{0.24\textwidth}
    \centering
\includegraphics[width=\textwidth ,clip]{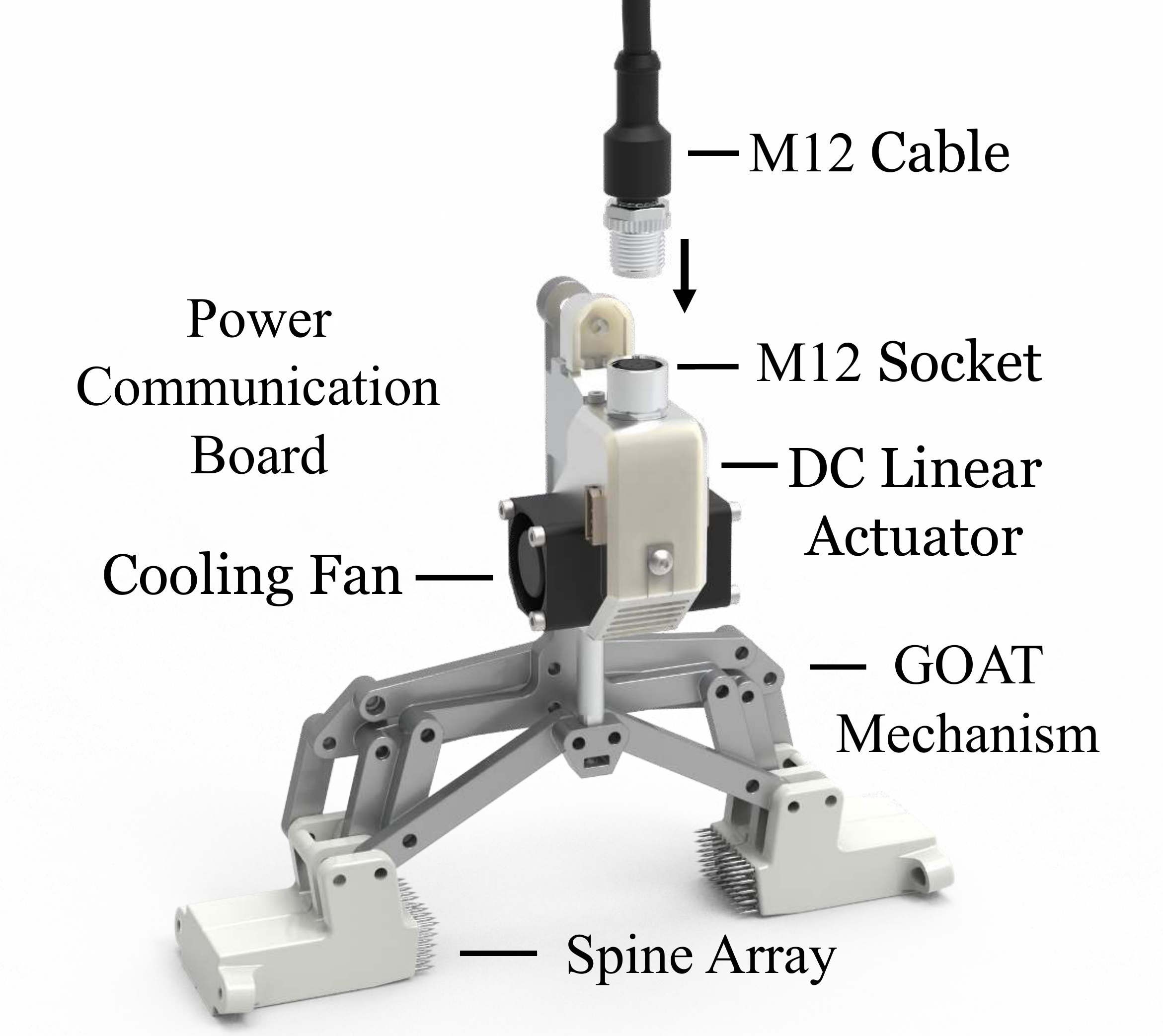}
     \caption{The GOAT gripper with a DC linear actuator and spine arrays. 
     \label{fig:hayato-goat}}
  \end{subfigure}
  \hfill
  \begin{subfigure}{0.24\textwidth}
    \centering   
    \includegraphics[width=\textwidth , clip]{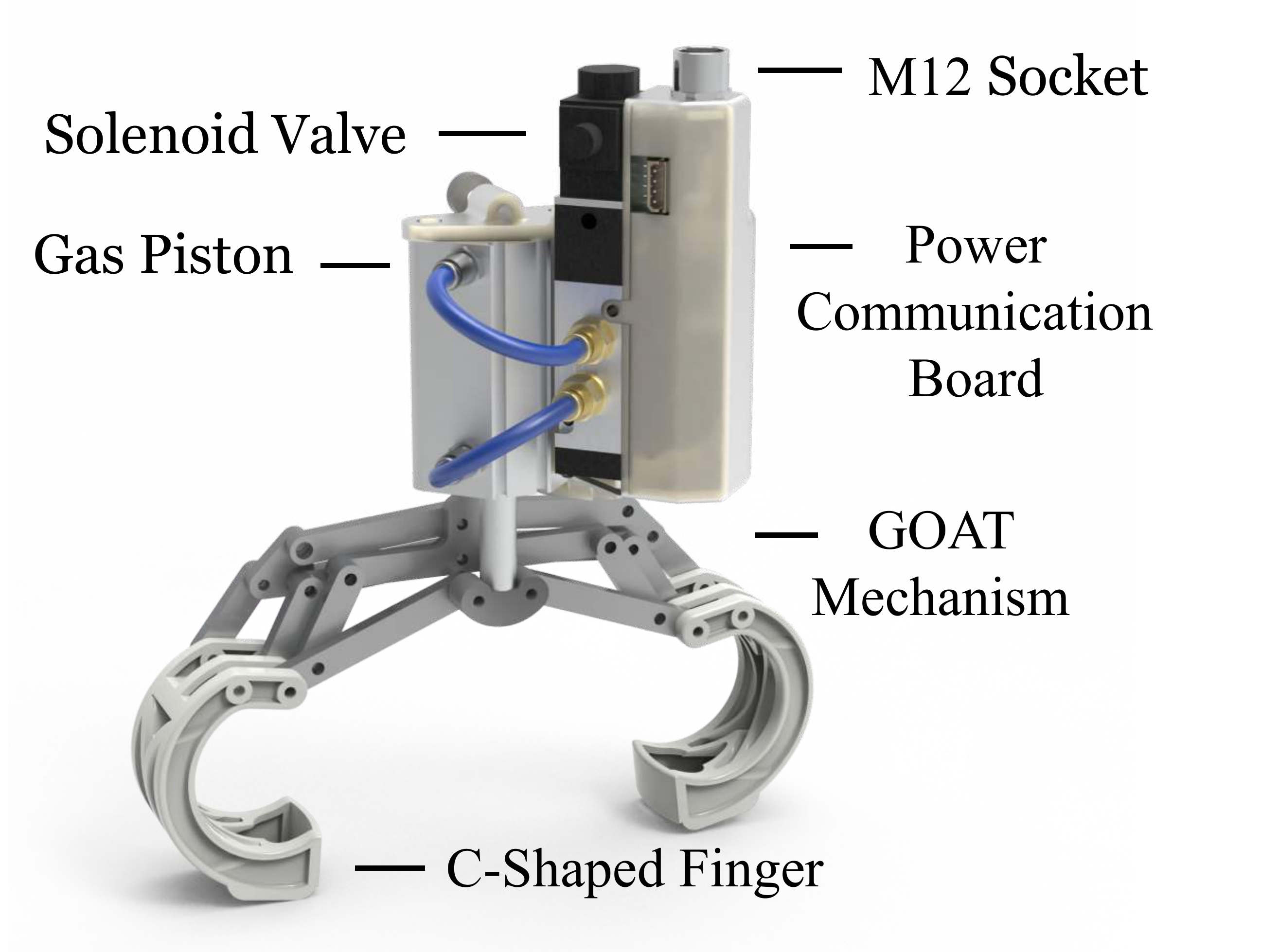}
     \caption{The \red{C-GOAT rendered image}. The fingers are covered with dry adhesive. 
     \label{fig:gas-goat}}
  \end{subfigure}
  \caption{Two variants of GOAT grippers.}
  \label{fig:both_goat}
\end{figure}

In the following section, we present the GOAT gripper \cite{GOAT}, a mechanically adaptable underactuated gripper. 
First, we outline the principles of the GOAT mechanism and its inherent capability in climbing environments.
Then, we describe \red{two} distinct modules of the gripper as follows:

\begin{itemize}
    \item DC linear actuator spine GOAT grippers in \fig{fig:hayato-goat}.
    \item Compressed gas pneumatic GOAT grippers with C-shaped fingers\red{: C-GOAT} in \fig{fig:gas-goat}.
\end{itemize}

The spine-enhanced fingertips are adequate on rough surfaces discussed in Section~\ref{sec:spinetip}, and the C-shaped fingertips can realize multi-modal grasping as detailed in Section~\ref{sec:goat_multi}, thereby illustrating versatility in diverse climbing challenges.

\subsection{The GOAT Mechanism\label{sec:gaot_mechanism}}
\begin{figure}[t!]
\centering
  \begin{subfigure}[t]{0.23\textwidth}
    \centering
    \includegraphics[height=4cm,width=\textwidth, trim={0cm 0cm 0cm 0cm},clip]{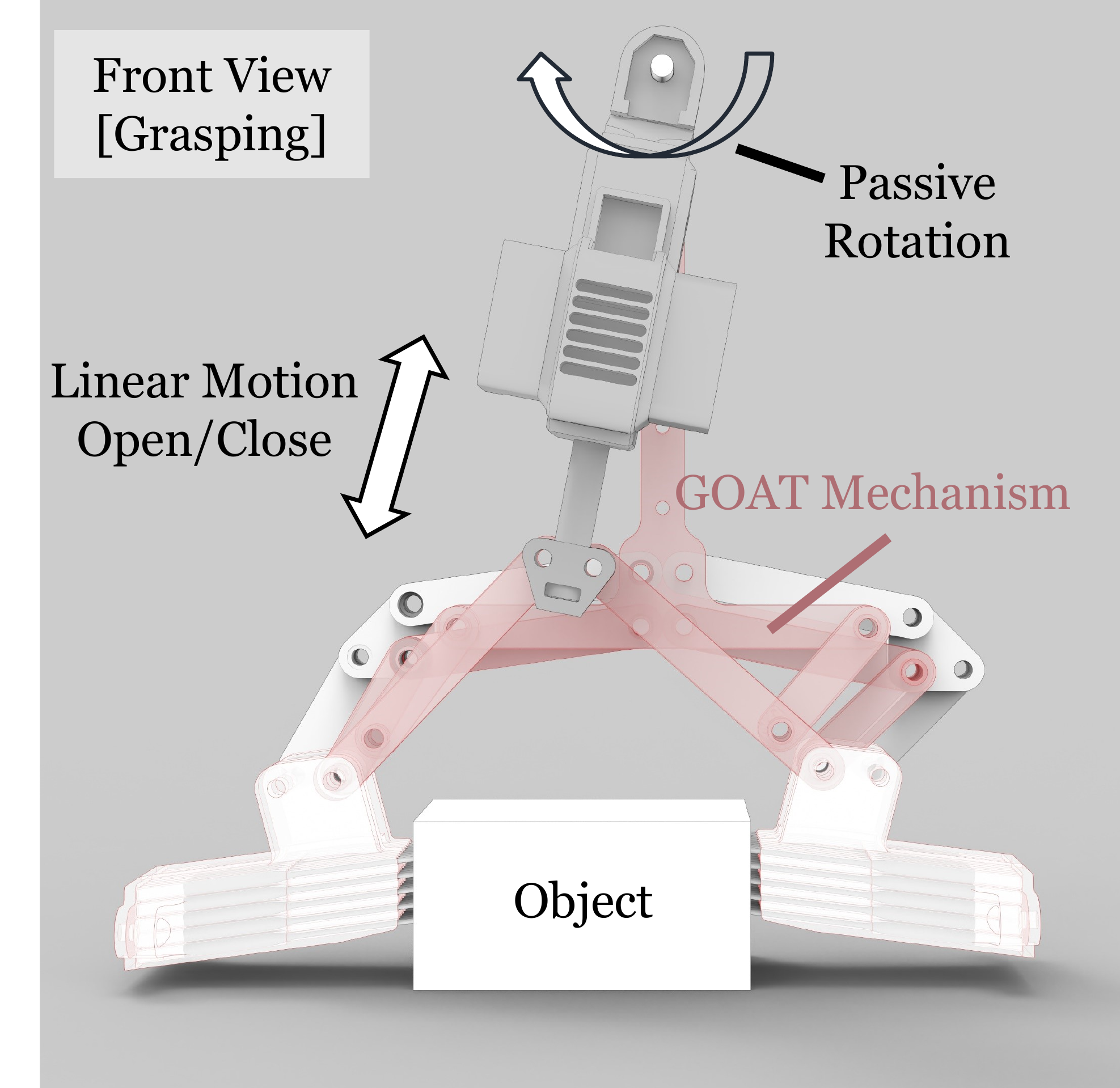}
     \caption{The GOAT gripper, when grasping at the off-axis. The GOAT mechanism \cite{GOAT} is visualized in red.\label{fig:GOAT}}
  \end{subfigure}
  \begin{subfigure}[t]{0.23\textwidth}
    \centering
    \includegraphics[height=4cm, width=\textwidth, trim={0cm 0cm 0cm 0cm},clip]{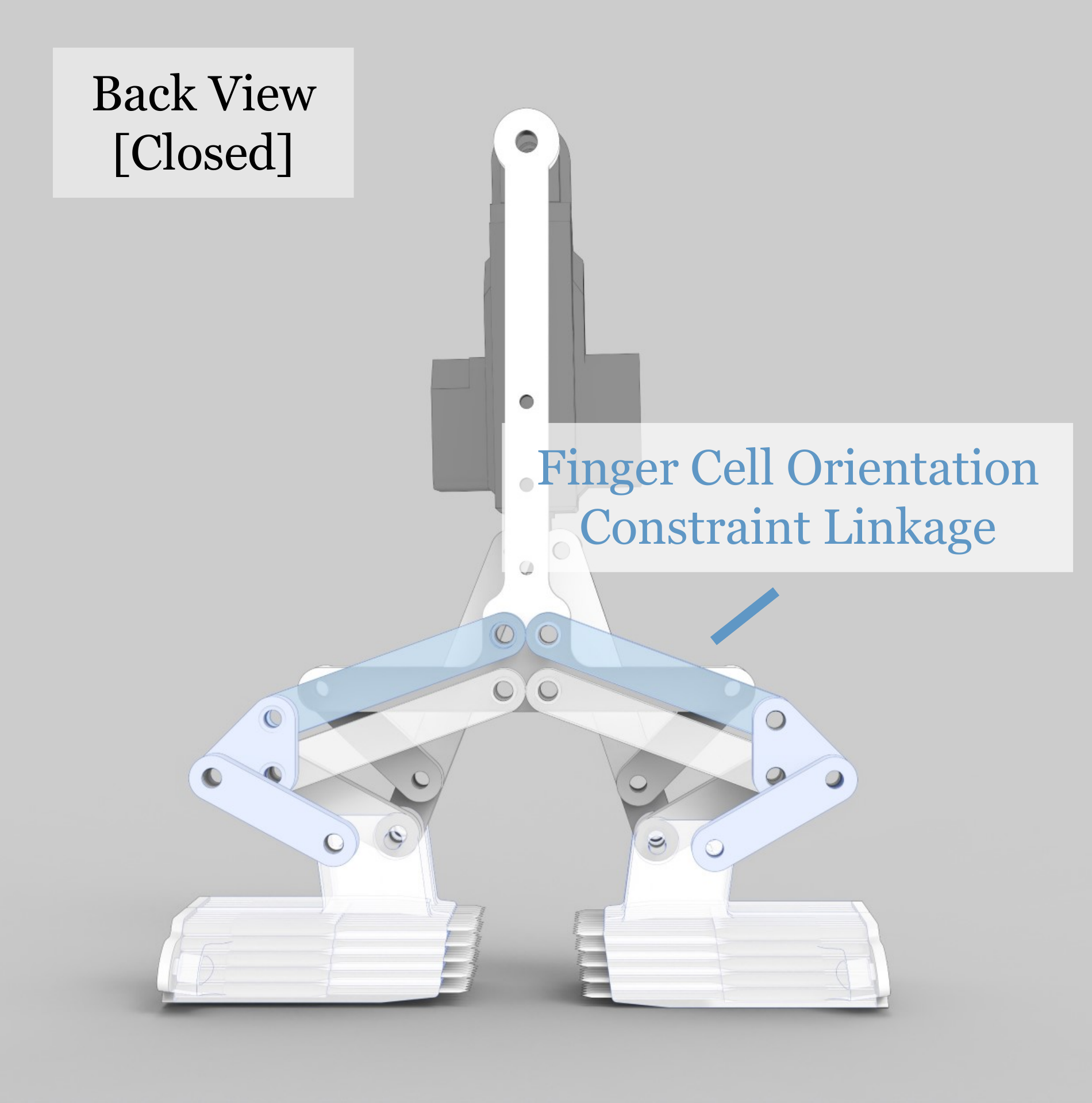}
     \caption{GOAT grippers when closed. The fingertip \red{angle} is constrained by the double parallelogram 4-bar linkage in blue.\label{fig:GOAT_const}}
  \end{subfigure}
  \caption{The GOAT mechanism and linkages.}
  \label{fig:goat_mechanisms}
\end{figure}

Here, we introduce the GOAT mechanism \cite{GOAT} visualized in \fig{fig:GOAT}, which is used in SCALER's GOAT grippers. 
The GOAT mechanism is a mechanically adaptable whippletree-based underactuated rigid linkage system.
Successful free-climbing in discrete environments requires grasping climbing holds despite various uncertainties. The GOAT gripper can help mitigate these challenges by compensating for end-effector configuration errors with one passive DoF.
Traditional parallel jaw grippers have only 1-DoF, either open or closed, and grasp objects at their central axis.
The GOAT's extra underactuation allows the GOAT gripper to adapt passively and grasp off-center objects as is, as illustrated in \fig{fig:GOAT}. This adaptability is further extended to realize multiple modes of grasping, as detailed in Section \ref{sec:goat_multi}. 

During the design process of the GOAT gripper, the linkage lengths, workspace, and forces are parametrically optimized, as discussed in \cite{GOAT}.  
The design process considered bouldering hold sizes and the force requirements necessary to support SCALER's weight. 
This formed a multi-objective nonlinear optimization problem with two objectives: two fingertips with active and passive DoF motion ranges and fingertip normal forces based on static force equilibrium. These two objectives are in a trade-off relationship, and the optimized GOAT mechanism is evaluated with bouldering holds in \cite{GOAT}.
\red{Since the optimization uses a 2D minimum bounding box, objects that meet this assumption and are in the size range are already considered, such as aluminum structures shown in \fig{fig:fig1}.}

\begin{figure}[t!]
    \centering
    \includegraphics[width=0.99\linewidth,trim={2.75cm 1.5cm 2cm 0cm} , clip]{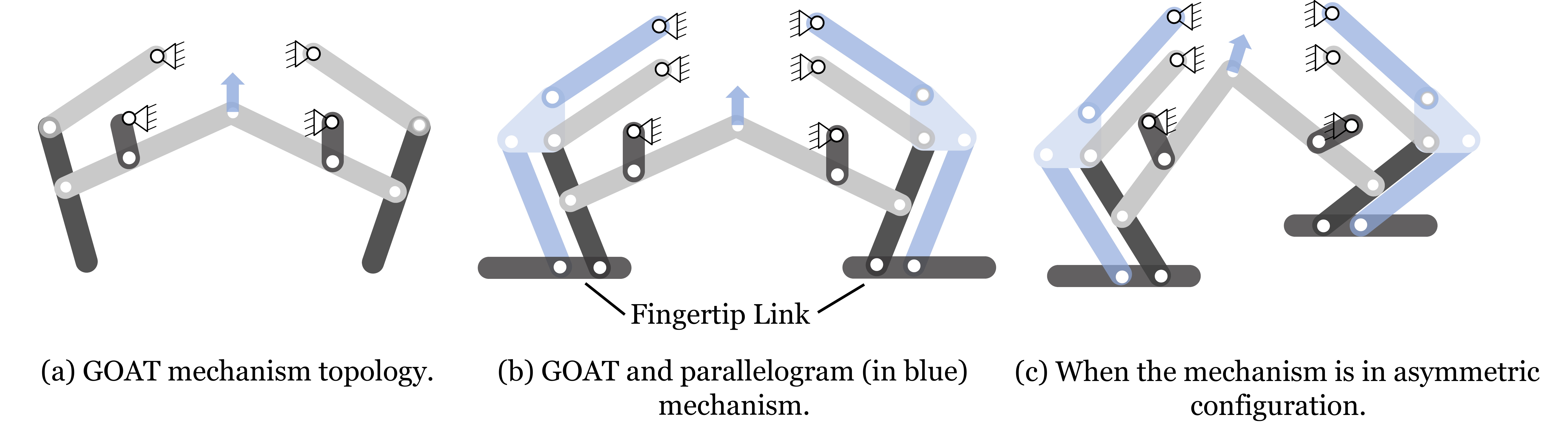}
     \caption{\red{The topology of the GOAT mechanism and the additional parallelogram mechanism constraining the fingertip orientations constant. The linkage lengths and configurations are non-optimal for visualization purposes.}
     \label{fig:goat_parallel}}
\end{figure}

In contrast to the design in \cite{GOAT}, the external four-bar linkages constrain the orientations of the fingertips to remain parallel to each other, as shown in \fig{fig:goat_parallel}, \ref{fig:GOAT_const}. This constraint ensures that the fingertip contact surface remains constant regardless of the GOAT gripper's kinematic configurations, as shown in \fig{fig:goat_parallel}b, \ref{fig:goat_parallel}c. The parallelogram mechanism linkages do not affect the gripper kinematics or optimization results. The parallelogram linkage lengths are trivial, which should be equal to those of the GOAT mechanism in parallel.

\subsection{Gripper Actuation\label{sec:gripper_module}}
\red{Here, }we discuss two actuators used in our GOAT gripper:
\begin{itemize}
    \item A DC linear actuator (\fig{fig:hayato-goat})
    \item A CO$_2$ high-pressure pneumatic actuator (\fig{fig:gas-goat})
\end{itemize}
While lightweight and capable of closed-loop controls, the DC linear actuators rendered in \fig{fig:hayato-goat} operate at a slower speed, which motivated us to employ a faster actuation. With the DC linear actuator, the GOAT grippers open and close fully at a frequency of \SI{0.12}{\hertz} with no load. In contrast, the high-pressure pneumatic actuator shown in \fig{fig:gas-goat} can run at \SI{5}{\hertz}. 
Additional gas tanks and regulators are installed on the SCALER's body.
The DC linear actuator is rated at \SI{100}{\newton} continuous output and a maximum of \SI{200}{\newton}. 
The pneumatic actuator can output a maximum of \SI{310}{\newton} force at \SI{1}{\mega\pascal}, but the nominal force is \SI{200}{\newton} at \SI{0.65}{\mega\pascal}. The higher force is necessary for C-GOAT since the friction coefficient is lower than that of the spine.
Consequently, the pneumatic-driven GOAT gripper enables SCALER to climb faster, albeit at the expense of payload due to the additional components and the absence of sensors compared to the DC linear actuator.

\begin{figure*}[t!]
    \centering
    \includegraphics[width=0.85\textwidth,trim={0cm 0cm 0cm 1cm},clip]{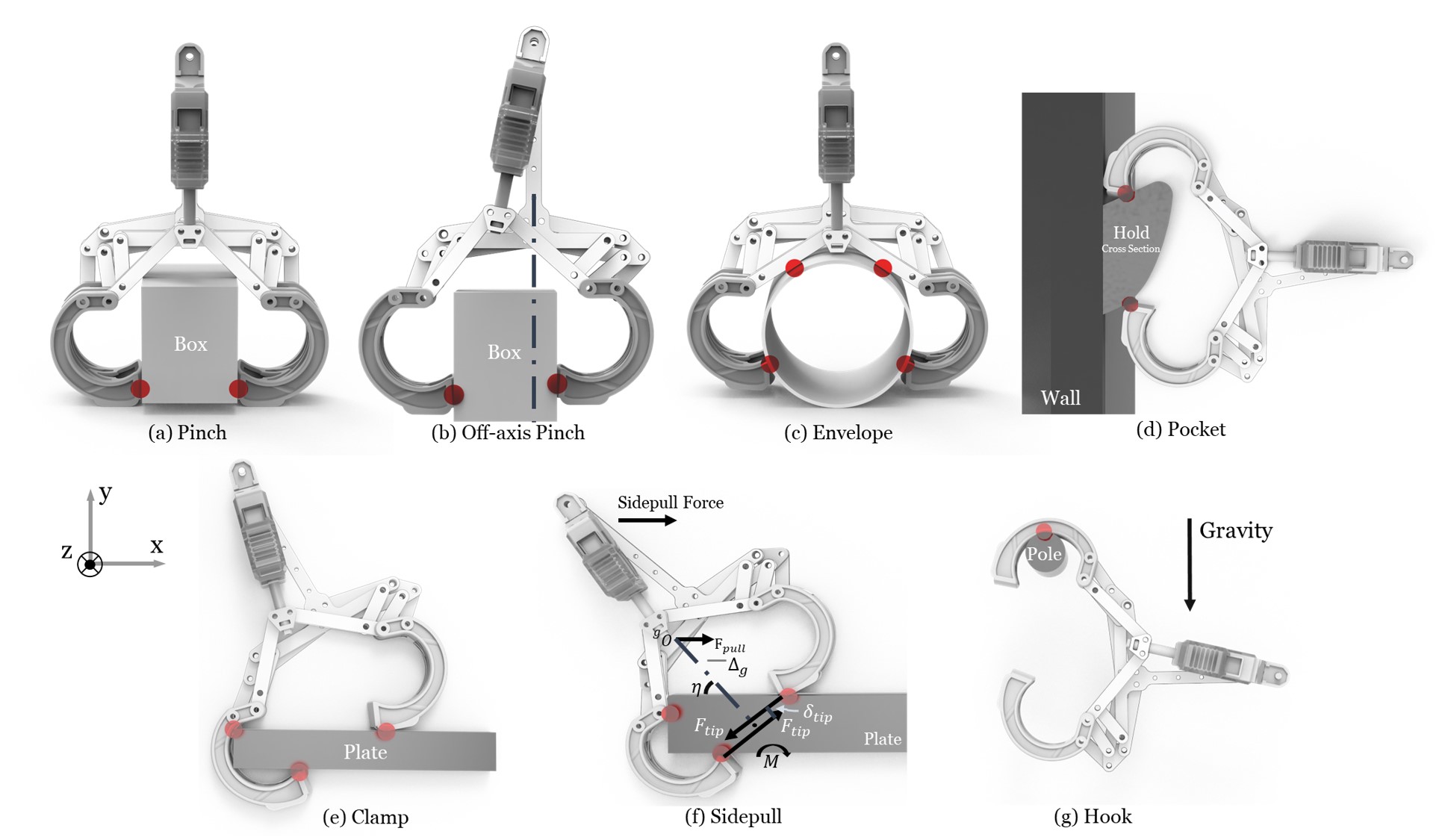}
     \caption{C-GOAT gripper's seven different grasping modes. Red dots represent primary points of contact that render grasping forces.
     \label{fig:pinch}}
\end{figure*}

\subsection{GOAT Gripper Fingertip\label{sec:goat_finger}}
GOAT grippers have two types of fingers depending on the surface and applications as follows:
\begin{enumerate}
    \item Spring-Loaded Spine Enhanced Fingers
    \item C-shaped Dry Adhesive Enhanced Fingers
\end{enumerate}
The spine tips are for rough and rocky surfaces, whereas the dry adhesive can grasp slippery terrains.

\subsubsection{Spring-Loaded Spine-Enhanced Fingers\label{sec:spinetip}}
The spring-loaded spine-enhanced finger GOAT gripper is rendered in \fig{fig:hayato-goat}.
Our spine cell design is based on \cite{spine_cell}. Each cell comprises fifty spines, each with a diameter of $\phi$ \SI{0.93}{\milli\meter}, and a \SI{5}{\meter\newton\per\milli\meter} spring. The cell surface is slanted so that the spines approach at an optimal angle. 
Strain gauges can measure normal-axis fingertip force, as demonstrated in \cite{alex_admittance}. The spine tips are more appropriate on textured and rough surfaces, such as rock, concrete, and bouldering holds \red{with} microcavities on their surface \cite{spine_array_uno}.

\subsubsection{C-shaped Dry Adhesive Enhanced Fingers\label{sec:dryadhesive}}
The \red{C-GOAT} gripper shown in \fig{fig:gas-goat} can realize \red{seven} different grasping modes by utilizing both actuated and underactuated DoFs, and the robot's whole body as shown in \fig{fig:pinch}. 
Because of finger geometries, the C-shaped finger has \red{an elastomer dry adhesive applied to} both the inside and outside C sections instead of spine arrays. The dry adhesive increases the friction coefficient, and thus, the robot can climb slippery terrains.

\subsection{Multi-Modal Grasping with \red{C-GOAT}\label{sec:goat_multi}}
\subsubsection{GOAT Gripper Multi-Modality}
Diverse grasping modes enable robots to employ an appropriate grasping method for various geometries. 
Paired with C-shaped fingers and one underactuated DoF, SCALER can embrace the potential for versatile and dexterous multi-modal grasp in climbing.
The list of modes and applicable objects is listed in Table \ref{tb:modes}.

\begin{table}[t]
\centering
\begin{threeparttable}
\caption{Types of grasping modes and their characteristics \label{tb:modes}}
\begin{tabular}{lcc}
\hline
Mode & Objects & Figure \\  \hline
Pinch & General Items in the range & \ref{fig:pinch}a \\
\rowcolor[HTML]{EFEFEF} 
Pinch Off-axis & Items placed off-centered &\red{\ref{fig:pinch}b} \\
Envelope & Cylindrical  items & \red{\ref{fig:pinch}c} \\
\rowcolor[HTML]{EFEFEF} 
Clamp & \begin{tabular}[c]{@{}c@{}} Plates \red{thinner} than the C-finger opening\end{tabular} & \red{\ref{fig:pinch}f} \\
Sidepull & \begin{tabular}[c]{@{}c@{}} Plates \red{thicker than the C-finger opening}\end{tabular} & \red{\ref{fig:pinch}g} \\
\rowcolor[HTML]{EFEFEF}     
Pocket & Items with pocket holes larger than fingertips & \red{\ref{fig:pinch}d} \\
Hook & Bars and pipes smaller than C-finger & \red{\ref{fig:pinch}g} \\ \hline
\end{tabular}
\end{threeparttable}
\end{table}

Pinch and envelope grasps \cite{robotiq} are standard in two-fingered grippers rendered in \red{\fig{fig:pinch}a,c}, respectively. The pinch grasp relies more on friction, while the envelope encompasses an object mechanically, \red{adding more kinematic constraints}. Thanks to the GOAT gripper's underactuation, it can pinch grasp an object off-center axis as in \fig{fig:pinch}b.

The plate clamp in \red{\fig{fig:pinch}e} resembles a crimp grip, \redrev{allowing the GOAT gripper to clamp} thin plates with the C-shaped fingers and underactuation. 
When encountering plates with a thickness beyond the C-shaped opening, SCALER can embrace a whole-body approach, \textit{sidepull}, to stabilize the grasping instead. This tactic is common for a human climber to overcome \redrev{ungraspable} bouldering holds \cite{beal2011bouldering}. 
SCALER can pull the gripper to the side in \red{\fig{fig:pinch}f, which ensures the third inner contact point occurs, adding physical constraint.}

\begin{table}[]
\centering
\begin{threeparttable}
\caption{Grasping Mode Contact Characteristics}\label{tb:contact_mode}
\begin{tabular}{ccccccc}
\hline
                   &  $\dot{x}^+$ &  $\dot{x}^-$ &  $\dot{y}^+$ & $\dot{y}^-$ &  $\dot{z}^+$ & $\dot{z}^-$ \\ \hline
\rowcolor[HTML]{EFEFEF} 
Pinch          &  $0$ &  $0$ &  $\dagger$ & $\dagger$ &  $\dagger$ & $\dagger$  \\
Pinch off-axis &  $0$ &  $0$ & $\ddagger$  & $\ddagger$   &  $\ddagger$ & $\ddagger$  \\
\rowcolor[HTML]{EFEFEF} 
Envelope        &  $0$ & $0$  &  $\ddagger\ddagger$ &  $0$ & $\ddagger\ddagger$  & $\ddagger\ddagger$  \\
Clamp          &  $0$ &  $\mathsection$ & $0$ &  $0$ & $\mathsection$  &  $\mathsection$ \\
\rowcolor[HTML]{EFEFEF} 
Sidepull       & $\ast$   &  $\mathsection$ & $0$  & $0$  &  $\mathsection$ & $\mathsection$  \\
Pocket         &  $0$ & $0$  &  $0$ & $0$ & $\dagger$  & $\dagger$  \\
\rowcolor[HTML]{EFEFEF} 
Hook / Encompass &  $0$ &  $\in \mathbb{R}$ &  $0$ &  $0$ & $\in \mathbb{R}$  & $\in \mathbb{R}$ \\
\hline
\end{tabular}
\begin{tablenotes}
      \footnotesize
      \item \red{Given $\eta:=\{x, y, z\}$,} \red{$\dot{\eta} = \dot{\eta}^+ - \dot{\eta}^-$, $\left(\dot{\eta}^+, \dot{\eta}^- \geq 0\right)$. $\dot{\eta}^+ = 0$ or $\dot{\eta}^- = 0$ implies no motion due to physical constraints in that respective direction.  $\mathcal{Q}_p$, $\mathcal{Q}_po$, $\mathcal{Q}_e$, $\mathcal{Q}_c$ are the limit surface of the pinch, pinch off-axis, envelope, and clamp, respectively, in \fig{fig:limit_surface}.}
      \item \red{$\dagger$: $0$ if $F \in \mathcal{Q}_p$. $\ddagger$: $0$ if $F \in \mathcal{Q}_{po}$. $\ddagger\ddagger$: $0$ if $F \in \mathcal{Q}_e$.}
      \item \red{$\mathsection$: $0$ if $F \in \mathcal{Q}_c$. $\ast$: $0$ if $F > F_{\text{sidepull}}$. Unrestricted if $\in \mathbb{R}$.}
\end{tablenotes}
\end{threeparttable}
\end{table}

\begin{figure*}[t!]
  \begin{subfigure}[t]{0.33\textwidth} 
    \centering
    \includegraphics[height=4.75cm, width=\textwidth]{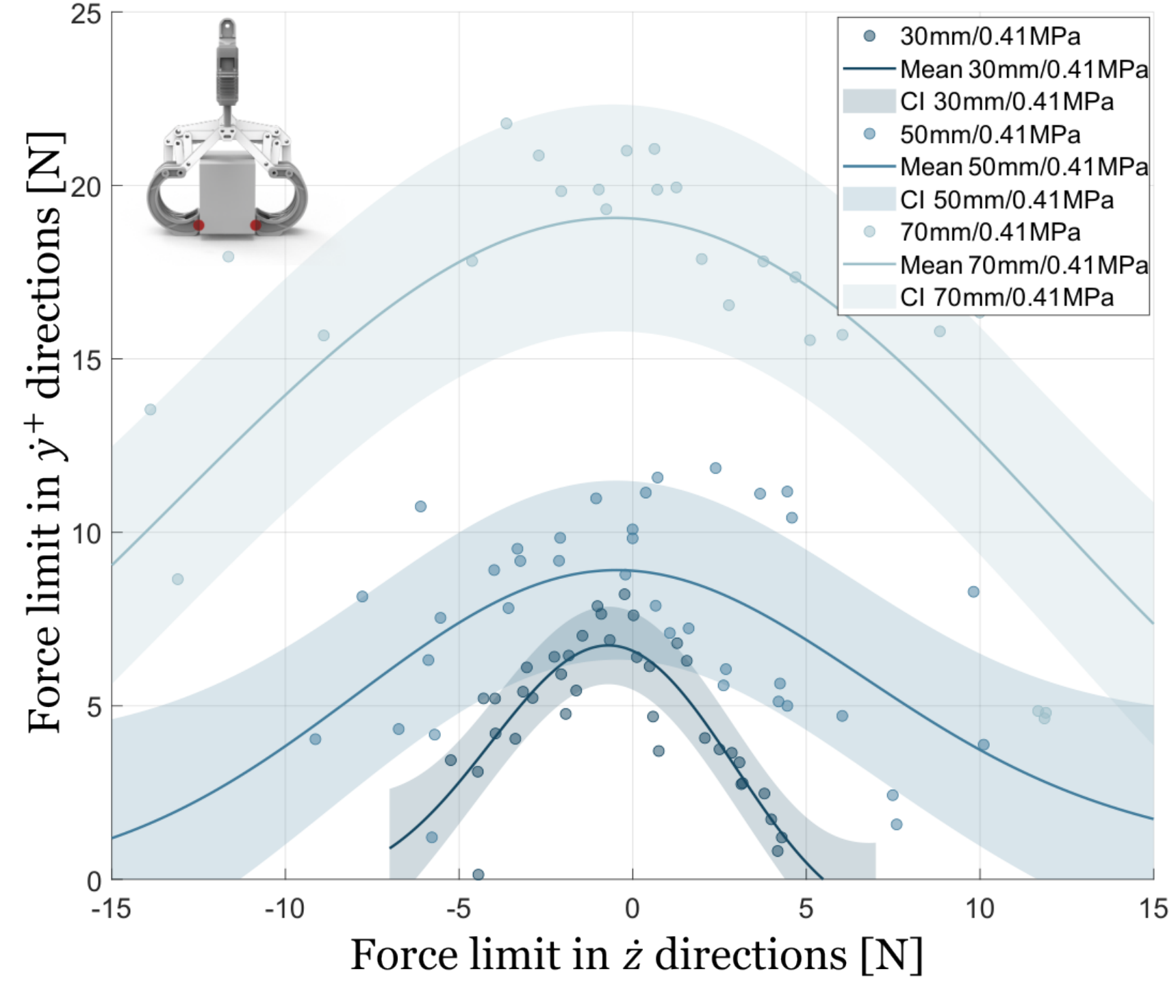}
     \caption{\red{Pinch limit surface for boxes $\in \{30,50,70\}$ mm and pneumatic actuator pressure at 0.41MPa, $\mathcal{Q}_p$.}}
     \label{fig:limit_pinch_60psi}
  \end{subfigure}
  \begin{subfigure}[t]{0.33\textwidth}
    \centering
    \includegraphics[height=4.75cm,width=\textwidth]{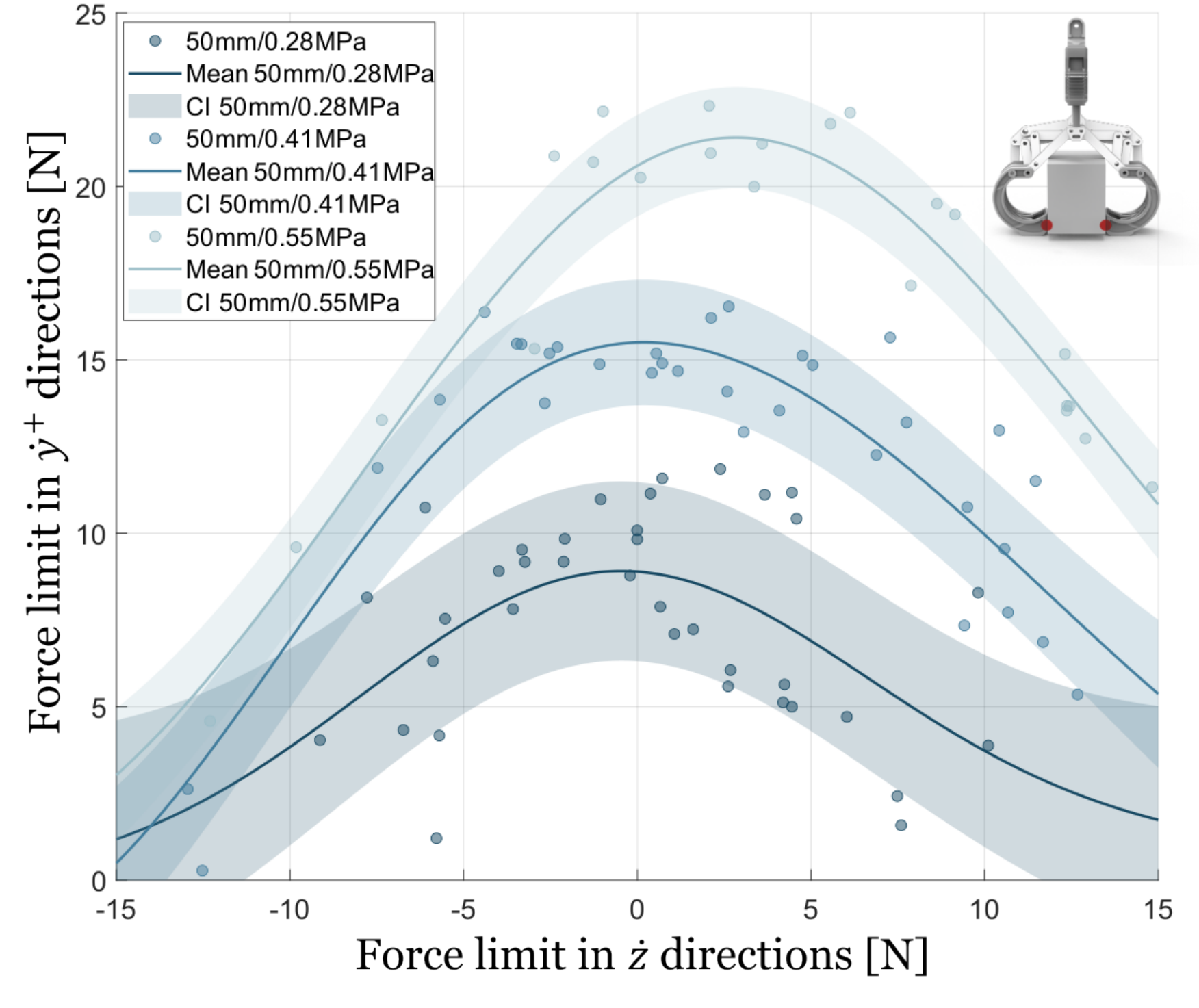}
     \caption{\red{Pinch limit surface for the $50$ mm box pneumatic actuator pressure at $\in \{0.28,0.41,0.55\}$ MPa, $\mathcal{Q}_p$.}}
      \label{fig:limit_pinch_50mm}
  \end{subfigure}
    \begin{subfigure}[t]{0.33\textwidth} 
    \centering
    \includegraphics[height=4.75cm,width=\textwidth]{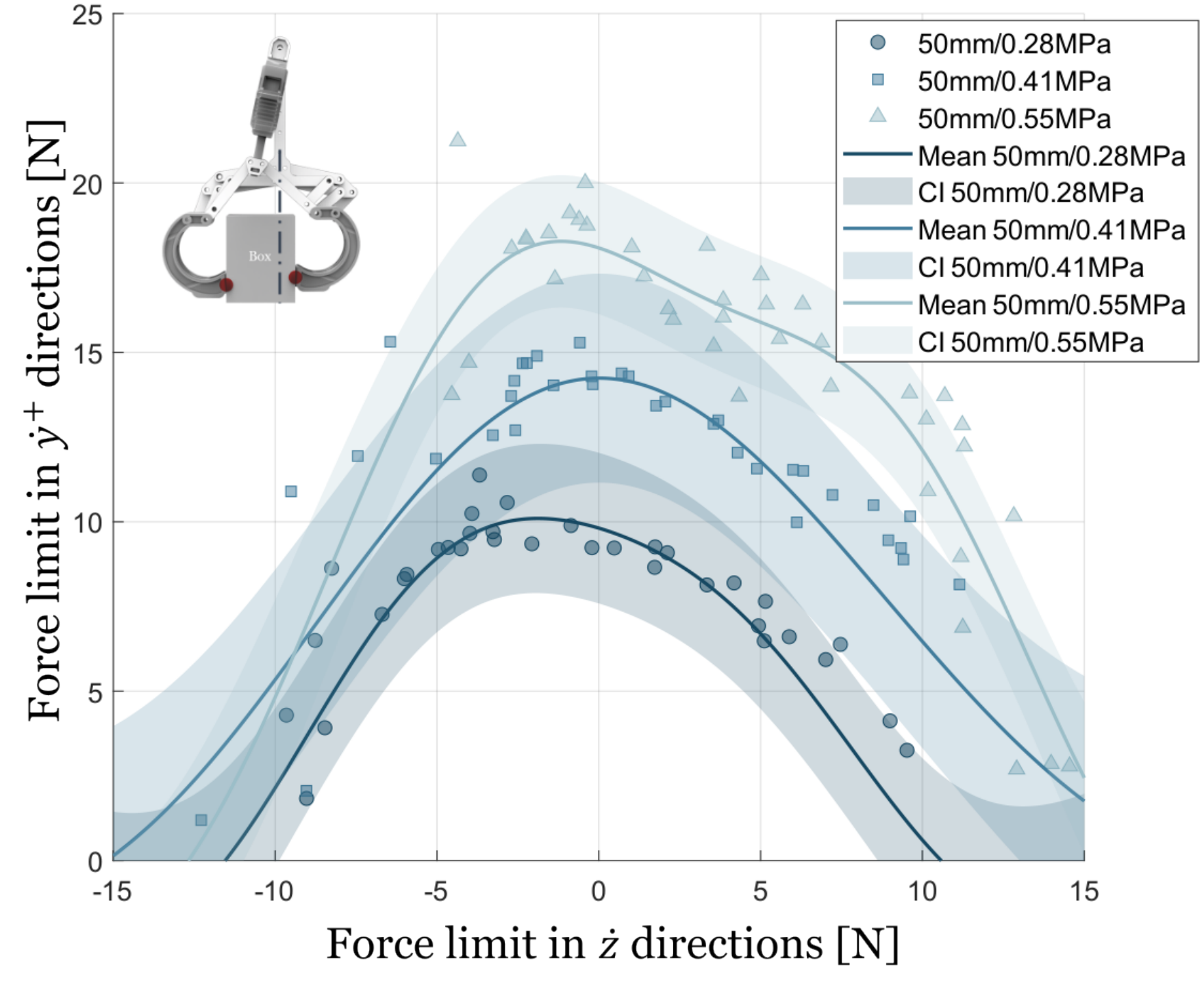}
     \caption{\red{Pinch off-axis limit surface,for the $50$ mm box pneumatic actuator pressure at $\in \{0.28,0.41,0.55\}$ MPa, $\mathcal{Q}_{po}$.}}
     \label{fig:limit_pinch_off}
  \end{subfigure}
    \begin{subfigure}[t]{0.33\textwidth}
    \centering
    \includegraphics[height=4cm,width=\textwidth]{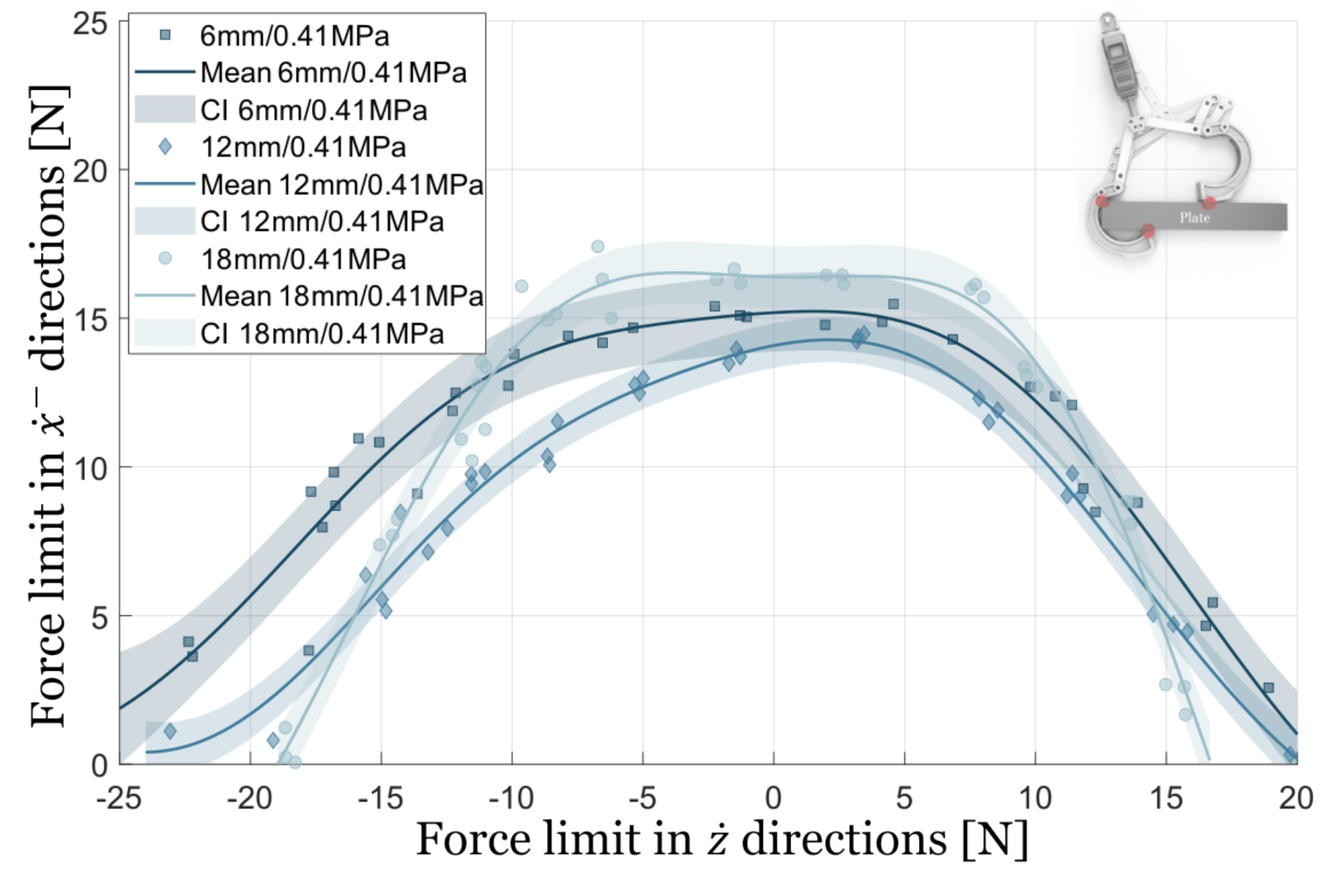}
     \caption{\red{Clamp limit surface for plates $\in \{6,12,18\}$ mm and pneumatic actuator pressure at 0.41MPa, $\mathcal{Q}_c$.}}
      \label{fig:limit_clamp_40psi}
  \end{subfigure}
  \begin{subfigure}[t]{0.33\textwidth} 
    \centering
    \includegraphics[height=4cm,width=\textwidth]{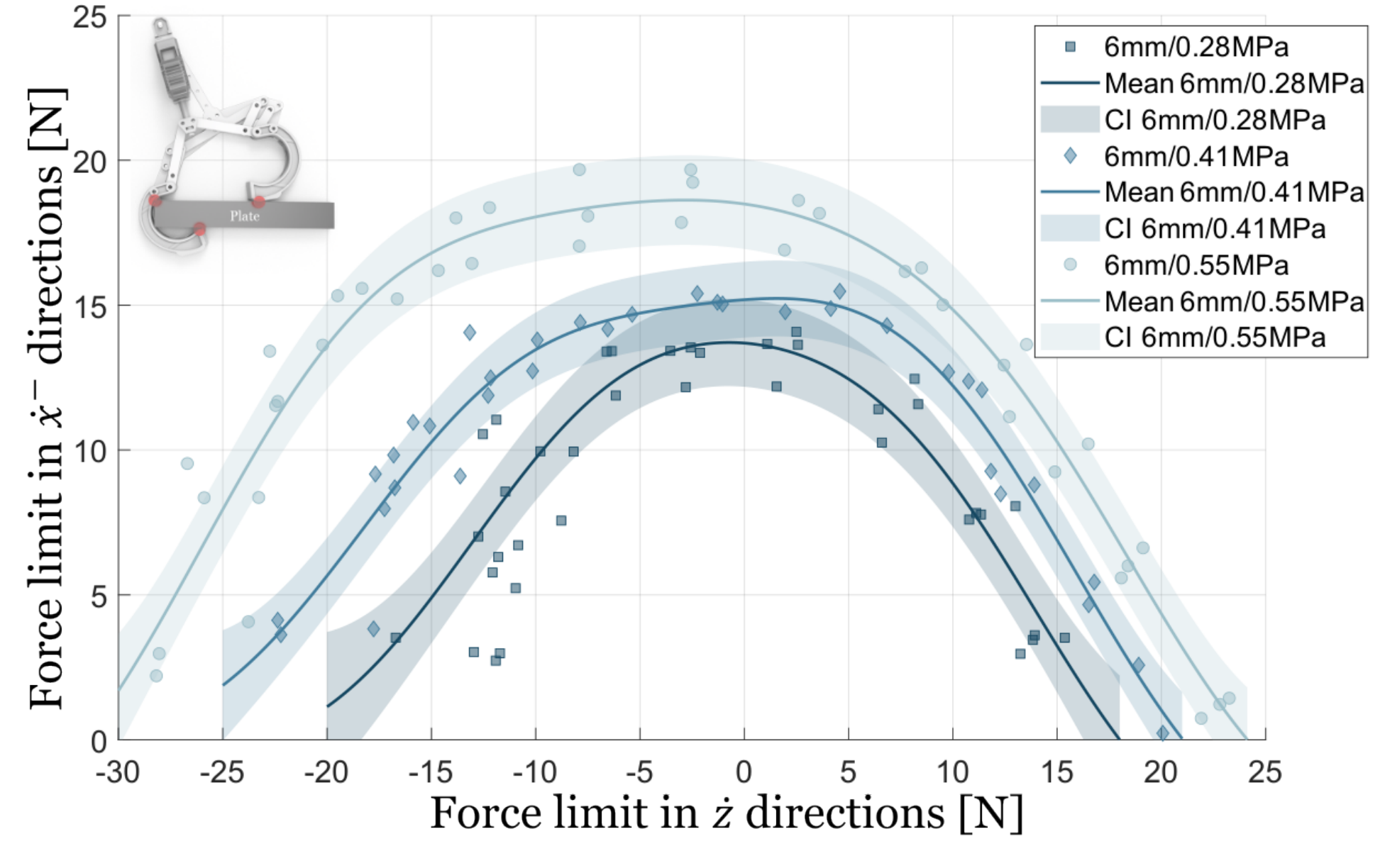}
     \caption{\red{Clamp limit surface for the $6$ mm plate pneumatic actuator pressure at $\in \{0.28,0.41,0.55\}$ MPa, $\mathcal{Q}_c$.}}
       \label{fig:limit_clamp_6mm}
  \end{subfigure}
    \begin{subfigure}[t]{0.33\textwidth} 
    \centering
    \includegraphics[height=4cm,width=\textwidth]{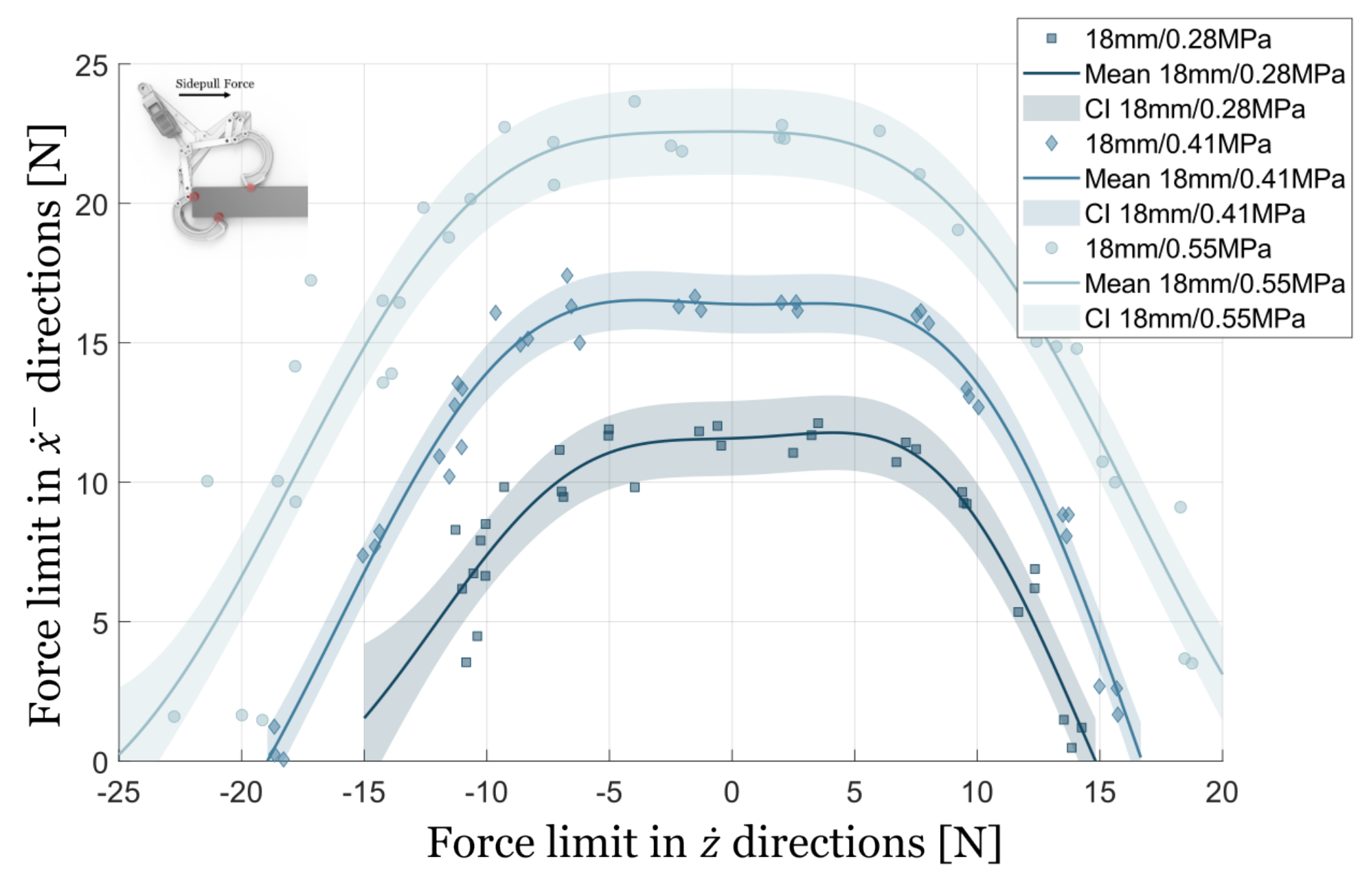}
     \caption{\red{Clamp limit surface for the $18$ mm plate pneumatic actuator pressure at $\in \{0.28,0.41,0.55\}$ MPa, $\mathcal{Q}_c$.}}
     \label{fig:limit_sidepull}
  \end{subfigure}
  \caption{\red{Limit surfaces of the multi-modal C-GOAT gripper. Gaussian Process Regression calculated the mean and confidence interval for each dataset. For pinch, $\dot{y}^-$ direction will be symmetric.
  Data is collected without dry adhesive due to the manipulator's force limit. The definition of $\eta \in {\dot{x}, \dot{y}, \dot{z}}$ in \tab{tb:contact_mode}.
  }} 
  \label{fig:limit_surface}
\end{figure*}

Pocket grasp \cite{beal2011bouldering}, and hook \cite{balaguer2005climbing} allow SCALER to hook onto narrow structures, such as \red{large cavity,} poles, and wire, as shown in \fig{fig:pinch}d, g. \red{These methods have more physical constraints and less reliance on frictional forces. Pocket grasp can generate a grasping force with the gripper actuator, whereas hook solely relies on kinematic constraints or external forces.}
Demonstrations of multi-modal grasping are explored in Section \ref{sec:multi_modal_climb} and \ref{sec:whole_body}.

\subsubsection{Sidepull Force \red{Requirements}\label{sec:sidupull_model}}
The fingertip's normal force will generate a moment when grasping a thick plate. Hence, the sidepull force must cause at least a greater torque as in \eqref{eq:sidepull} to have three contact points, and parameters visualized in \red{\fig{fig:pinch}f}. This becomes a pinch grasp if the torque is insufficient.
\begin{equation}
\left\{
\begin{aligned}
    & F_{\mathrm{tip},i} \\
    & F_{\mathrm{pull}}
\end{aligned}
\,\middle|\,
\begin{aligned}
    & \sum \mathcal{M} = \sum_{i=1}^{2} \delta_{\mathrm{tip}}\, F_{\mathrm{tip},i} + \Delta_g\, F_{\mathrm{pull}} \sin{\eta} \geq 0, \\  
    & F_{\mathrm{tip},i} \in \mathcal{Q}_\text{limit},\quad i=1,2
\end{aligned}
\right\}
\label{eq:sidepull}
\end{equation}

Here, $\mathcal{M}$ is a moment acting due to the fingertip force vectors denoted as $F_{\mathrm{tip},i}$. $\mathcal{\delta}_{\mathrm{tip}}$ is the Euclidean distance between $F_{\mathrm{tip},i}$, $F_{\mathrm{tip}}$. $\Delta_g$ is the perpendicular distance from the center of grasp and the gripper frame, $F_{\mathrm{pull}}$ is the sidepull force, and $\eta$ is the gripper angle with respect to the plate grasping. The sidepull whole-body technique is experimented with in Section~\ref{sec:whole_body}.

\subsection{\red{C-GOAT Limit Surface Analysis\label{sec:limit_surface}}}

Here, we discuss the contact dynamics of seven grasping modes with the C-GOAT gripper. The spine-enhanced GOAT gripper's analysis is in \cite{GOAT}. 
All modes have distinct contact characteristics, resulting in unique constraints and limit surfaces as listed in \tab{tb:contact_mode} and corresponding limit surfaces in \fig{fig:limit_surface}. 
The limit surfaces indicate the maximum force each grasping mode can withstand in certain directions, such as normal and tangential direction, over various object geometries as defined in \fig{fig:pinch}.

\begin{table}[t]
\caption{\red{Mean pulling force limit in envelope mode for $\dot{y}^+$ and $\dot{z}$ directions (mean of 3 and 6 trials, respectively).}}
\label{tb:envelope}
\centering
\begin{tabular}{c|ccc|ccc}
\multirow{2}{*}{\begin{tabular}[c]{@{}c@{}}Diameter\\/ \textbf{Pressure (\si{\mega\pascal})}\end{tabular}} 
& \multicolumn{3}{c|}{$\dot{y}^+$ direction force (\si{\newton})} 
& \multicolumn{3}{c}{$\dot{z}$ direction force (\si{\newton})} \\
& \textbf{0.28} & \textbf{0.41} & \textbf{0.55} & \textbf{0.28} & \textbf{0.41} & \textbf{0.55} \\ \hline
\rowcolor[HTML]{EFEFEF}
\SI{50}{\milli\meter} & 73.0 & 102.7 & 116.3 & 7.6 & 13.0 & 18.0 \\
\SI{60}{\milli\meter} & 68.4 & 92.4 & 118.6 & 8.5 & 13.2 & 17.2 \\
\rowcolor[HTML]{EFEFEF}
\SI{80}{\milli\meter} & 43.1 & 54.7 & 67.7 & 10.5 & 17.5 & 22.9
\end{tabular}
\end{table}

\subsubsection{\red{Kinematic and Force Constraints}}
\red{
Here, we analyze the C-GOAT seven grasping modes' kinematic and force constraints.
For all grasping modes except hook, the $z$ directional velocities, $\dot{z}$ in Table~\ref{tb:contact_mode}, \fig{fig:pinch} are constrained to $0$ if the force vector is in the respective limit surface. For the hook, the gripper cannot apply any normal force, and hence, there is no bound for $\dot{x}^-$, $\dot{z}^+$, and $\dot{z}^-$. For the envelope grasp, the motion in $\dot{y}^+$ direction requires backdriving the gripper actuator to detach the gripper.
The sidepull mode has physical constraints in $\dot{x}^+$ if the sidepull force condition is satisfied, as in Section \ref{sec:sidupull_model}. 
If this condition is unmet, the grasp collapses to a pinch, oriented \SI{90}{\degree} around the $z$-axis.
}

\subsubsection{\red{Grasping Mode and Limit Surface}\label{sec:limit_surface_modes}}
We collected the C-GOAT limit surface data \fig{fig:limit_surface} using the SCALER manipulator configuration shown in \fig{fig:scaler-m}, equipped with the F/T sensor.
The mean and \SI{95}{\percent} confidence interval are calculated using Gaussian Process Regression with a square exponential kernel. 
 The tests were conducted without dry adhesive due to the arm's pulling force limitations. 
We have run separate experiments to determine the effectiveness of the dry adhesive.  
With dry adhesive in \tab{tb:limit_dry}, the maximum pulling force in the $\dot{y}^+$ direction is on average $3.7$ times greater compared to the same grasp and object scenario in \fig{fig:limit_pinch_50mm}. Although this scaling does not necessarily apply linearly to the limit surfaces, \tab{tb:limit_dry} provides insights to the dry adhesive effectiveness.

We analyze the limit surface of the pinch, off-axis pinch, clamp, sidepull, and envelope grasps to understand their contact force limit in the directions not kinematically constrained in the \tab{tb:contact_mode}. 
The mode of grasping, the object geometries, and the actuator forces affect the limit surfaces as in \fig{fig:limit_surface} and \tab{tb:envelope}. 
This contact analysis is useful in the future for tasks such as grasping planning. 
The pocket limit surface is equivalent to the pinch or pinch off-axis, but with more kinematic constraints. The hook solely relies on finger geometries.

\begin{table}[]
\caption{The mean maximum pulling force in the $\dot{y}^+$ direction for C-GOAT gripper pinch grasp on \SI{50}{\milli\meter} box with and with no dry adhesive.}
\label{tb:limit_dry}
\centering
\begin{tabular}{cccc}
Pressure & \SI{0.28}{\mega\pascal} & \SI{0.41}{\mega\pascal} & \SI{0.55}{\mega\pascal} \\ \hline
\begin{tabular}[c]{@{}c@{}}With dry adhesive\\ $\dot{y}^+$ direction\end{tabular} & \SI{36.4}{\newton} & \SI{54.5}{\newton} & \SI{78.4}{\newton} \\
\rowcolor[HTML]{EFEFEF} 
\begin{tabular}[c]{@{}c@{}}With no dry adhesive\\ $\dot{y}^+$ direction\end{tabular} & \SI{8.9}{\newton} & \SI{15.5}{\newton} & \SI{21.4}{\newton}
\end{tabular}
\end{table}

\textbf{\textit{Pinch Grasp Analysis:}}
First, we compare the pinch grasping on different object sizes and actuator forces. 
The pinch grasp force limit in \fig{fig:limit_pinch_60psi} decreases for smaller objects due to the nonlinear force transmission of the GOAT mechanism \cite{GOAT}, and the gripper kinematics significantly changes with this width range. 
The higher actuator force, such as higher gas pressures in \fig{fig:limit_pinch_50mm}, increases the limit surface as the normal force at the contact rises. The off-axis pinch \fig{fig:limit_pinch_off} exhibits a slight (<\SI{2.5}{\newton}) skewness in the peak and asymmetry compared to the pinch in \fig{fig:limit_pinch_50mm}. This is because the fingertip's normal forces are unequal due to off-axis grasping.

\textbf{\textit{Clamp and Sidepull Grasp Analysis:}}
The clamp grasp, on the other hand, shows consistent limit surfaces across the different plate thicknesses, as seen in \fig{fig:limit_clamp_40psi}. This is because the gripper kinematics does not significantly change over these geometries, unlike the pinch grasp. The $\SI{18}{\milli\meter}$ plate is sidepull due to its thickness, and the $\SI{12}{\milli\meter}$ plate is close to the clamp thickness upper limit. 
Increasing the actuator force also expands the limit surface for clamp and sidepull in \fig{fig:limit_clamp_6mm}, \ref{fig:limit_sidepull} due to an increase in normal force.

\red{\textbf{\textit{Pinch and Clamp Grasp Analysis:}}
The $\dot{z}$ directional limit ($x$-axis in \fig{fig:limit_surface}) is higher for the clamp than for the pinch grasp. 
The clamp is a more constrained grasp, as in \tab{tb:contact_mode} with \redrev{an} extra contact point. 
As discussed in Section~\ref{sec:goat_multi}, the sidepull can collapse to a pinch grasp. The sidepull has a larger limit surface in all directions in \fig{fig:limit_sidepull} even though the object is smaller ($\SI{18}{\milli\meter}$) compared to the pinch
in \fig{fig:limit_pinch_60psi}.}

\textbf{\textit{Envelope Grasp Analysis:}}
The envelope in \tab{tb:envelope} was tested for the cylinder diameters $\in \{50,60,80$\}\si{\milli\meter}, which is close to the lower and upper bound of envelope graspable size. 
In the $\dot{y}^+$ direction force, the envelope grasp requires significantly higher forces than other limits in \fig{fig:limit_surface} since this backdrives the gripper actuator as seen in \fig{fig:pinch}c. As the diameter increases, this force limit decreases since less backdriving is required.
Whereas, the results in the $\dot{z}$ direction are consistent with those of the pinch grasp in \fig{fig:limit_surface}, because both grasps share similar kinematic configurations. Hence, the limit force decreases as the object diameter decreases for the same reason as the pinch. The increase in the actuation force raises the limits in both directions across all sizes tested.

\subsection{Grasping Force Controllers\label{sec:grasping_control}}
The grasping force, or the normal force at each fingertip, is crucial in stable climbing. 
Increasing grasping force can help stabilize the grasp during incipient slipping \cite{contact_rich}, \cite{donlon2018gelslim}.
However, consistent grasping can generate excessive heat, degrading the DC actuator over time; hence, grasping force control is necessary. 
The DC linear actuated GOAT gripper includes three controllers: 1) position control, 2) current-based force control, and 3) stiffness model-based force control.

The pneumatic actuator only has an on-off control, which lacks force feedback but does not overheat. More discussion on the tactile limitation is in Section~\ref{sec:limit_c_shape}.
The MightyZap DC linear actuator by IR Robot features position and current-based force controls. \redrev{The fingertip position is computed with the gripper's inverse kinematics.}
Stiffness model data is collected using a load cell while grasping different object sizes in \cite{alex_admittance}. 
The spine GOAT gripper's maximum withstanding force has been modeled using Gaussian process regression in \cite{GOAT}. The model is stochastic due to the spine tip \cite{spine_cell}.

\begin{figure}[t!]
    \centering
    \includegraphics[width=0.37\textwidth, trim={0cm 0cm 0cm 0cm},clip]{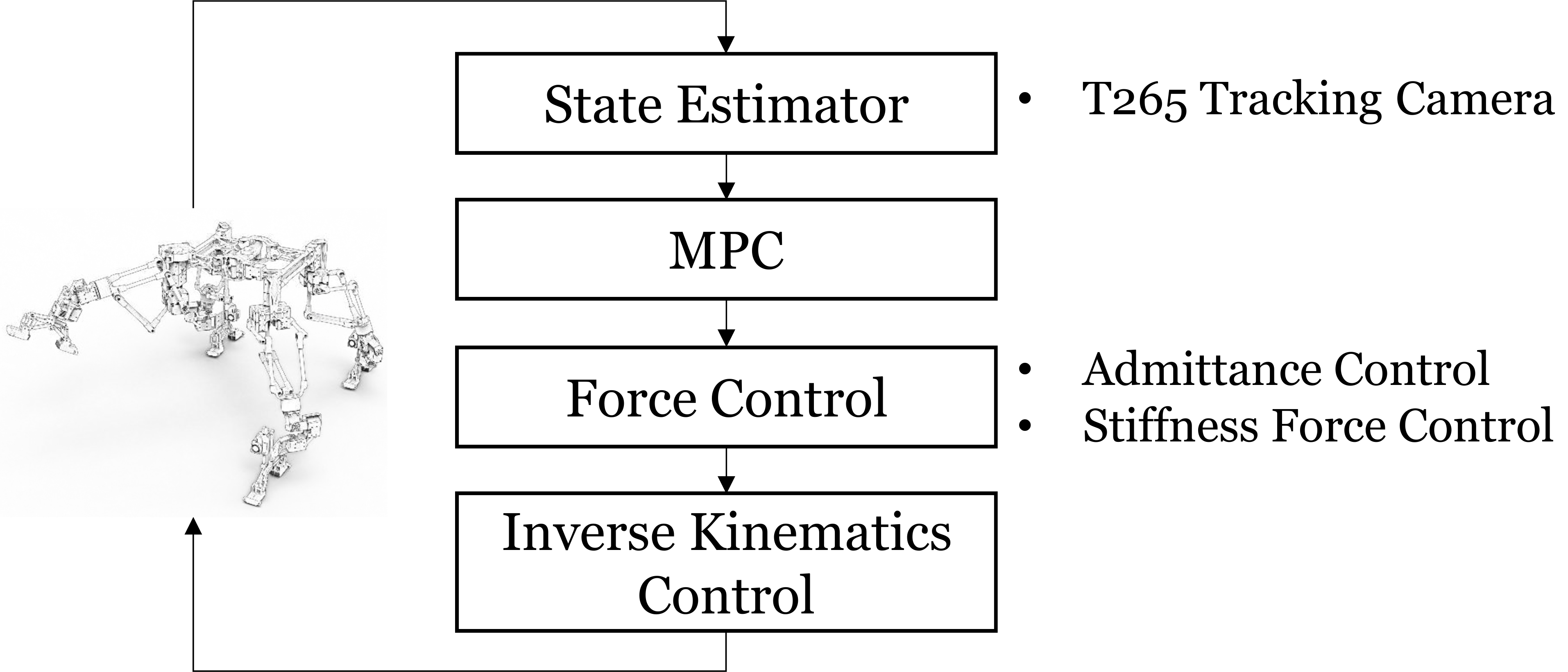}
     \caption{SCALER software flowchart.}
\label{Fig:software}
\end{figure}

\section{SCALER Software \label{sec:software}}
\red{This section discusses} SCALER's software, sensors, and control frameworks in \fig{Fig:software}.

\subsection{Sensor-Actuator Interface and Frequencies}
SCALER is equipped with various sensors: four F/T sensors on the wrists, a localization camera, a body IMU sensor, Dynamixel actuator encoders, and sensors on the GOAT gripper. 
The sensors and actuators operate at frequencies sufficiently higher than the controller modules. The Bota Rokubi F/T sensors are at \SI{800}{\hertz}, the Intel RealSense T265 localization camera is at \SI{200}{\hertz}, the Microstrain 3DM-GX4-25 IMU runs at \SI{500}{\hertz}, and Dynamixel at \SI{400}{\hertz}.
Each limb features a redundant communication chain for robustness and safety. The wrist and gripper actuators are connected to the shoulder power and communication board via an M12 cable. 

\subsection{State Estimation\label{sec:lse}}
SCALER employs the vision SLAM T265 camera to track the trunk state with respect to the world reference frame. Kinematics and joint encoders are used to determine all four end-effector states. The T265 camera tracking is analyzed in \cite{optistate}. For bouldering wall holds, \cite{yusuke_scaler_2022} demonstrated primitive-based sparse object mapping, which estimates the hold's pose and shape as an ellipsoid. The reported hold pose and shape estimation errors are $\pm\SI{4}{\milli\meter}$ and \SI{12}{\percent} of their size \cite{yusuke_scaler_2022}. In this paper's experiments, we have not utilized this mapping framework due to limitations such as cases where the limbs are in view and disturbing the depth images.

\subsection{Force Admittance Control Framework}
Admittance control tracks desired operational space force profiles in contact-rich tasks.
Wrench or force tracking is essential to sustain dynamic stability and mitigate disturbance due to premature or unintended contacts. 
The control framework \eqref{eq:admittance_control} extends the formulation in \cite{admittance} to accommodate the wrench reference trajectories. Each leg has an independent admittance controller.
\begin{equation}
    K_f(\mathbf{W}-\mathbf{{W}}_{\mathrm{ref}}) = M_d \ddot{\mathbf{x}} + D_d \dot{\mathbf{x}}+K_d(\mathbf{x}-\mathbf{x}_{\mathrm{ref}})
    \label{eq:admittance_control}
\end{equation}
Here, $K_f$, $M_d$, $D_d$ and $K_d$ are the user-defined diagonal admittance control gain in $\mathbb{R}^6$, $\mathbf{x}$. $\dot{\mathbf{x}}$ and $\ddot{\mathbf{x}}$ are the end effector 3D pose and its first and second derivative. $\mathbf{W}$ is a 3D wrench vector. $\mathbf{W}_{\mathrm{ref}}$ and $\mathbf{x}_{\mathrm{ref}}$ are the desired wrench and pose. The control input is the double integral of $\ddot{\mathbf{x}}$. The $\mathbf{W}$ is estimated with the F/T sensors and the method in \cite{alex_admittance}. The $\mathbf{x}$ and $\dot{\mathbf{x}}$ are measurements.
Due to gear backlash, the controller struggles to track rapidly changing force profiles \cite{alex_admittance}.
This control bandwidth suffices for quasi-static climbing that experiences less frequent force reference changes. The admittance controller experiments are shown in \cite{contact_rich} and \cite{alex_admittance}.


\begin{figure}[t!]
    \centering
    \includegraphics[width=0.49\textwidth,trim={0.5cm 0cm 0cm 0cm},clip]{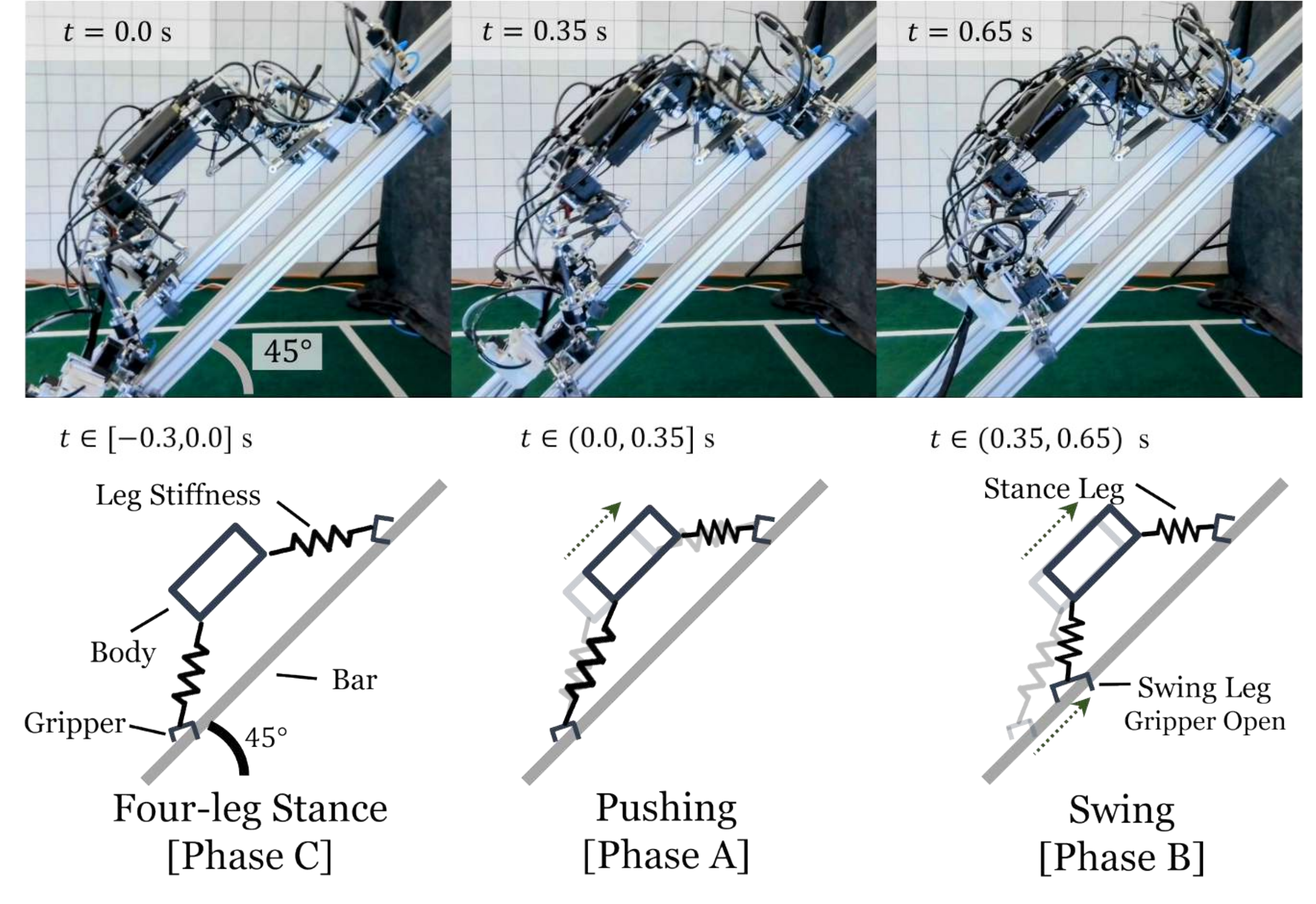}
     \caption{SCALER modified trot gait on the \SI{45}{\degree} slope. SCALER starts from the end of \textbf{Phase C}. For \textbf{Phase A} $t \in (0.0, 0.35]$~\si{\second} SCALER \red{push the ground and stiffen} the leg to increase the support force on the stance leg. For \textbf{Phase B} $t\in(0.35, 0.65]$~\si{\second}, two diagonal RB/LF legs \red{pair here} swings to the next footsteps. For \textbf{Phase C} $t\in(0.65, 0.95]$~\si{\second} SCALER closes grippers and is in 4 leg stance. This test was performed in open-loop.
     The kinematic drawing illustrates the two legs and the body motions with imaginary linear leg springs for visualization purposes.
    \label{fig:trot_45}} 
\end{figure}

 \begin{figure*}[t!]
    \centering
    \includegraphics[width=0.99\textwidth,trim={0.8cm 0cm 0.8cm 1cm},clip]{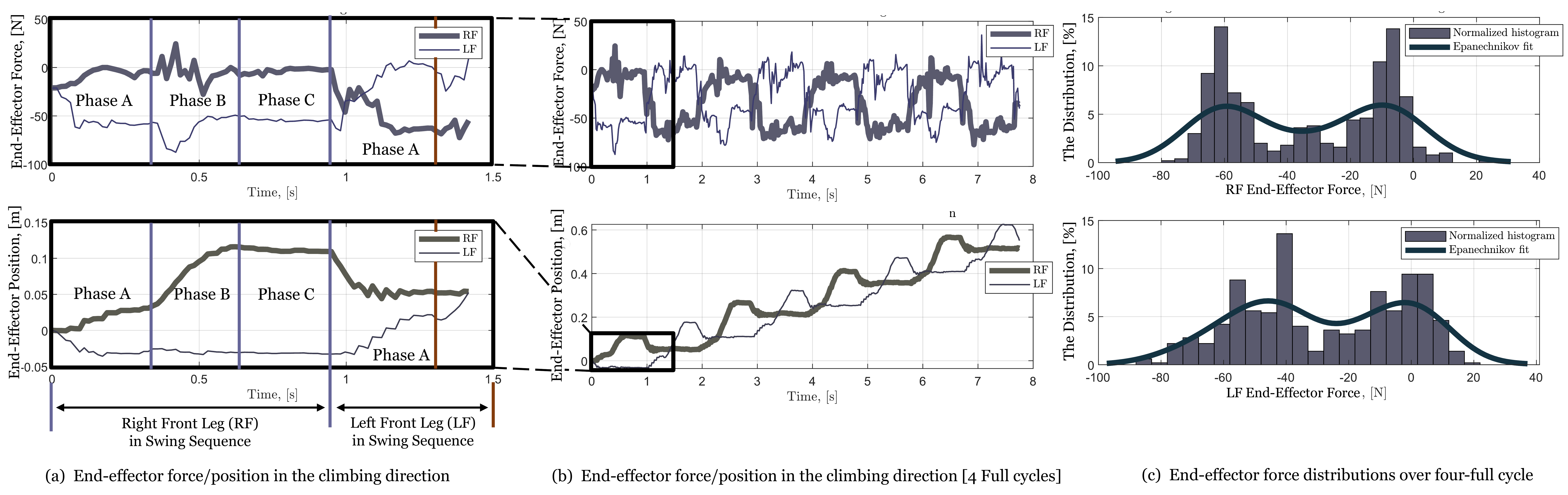}
     \caption{Foot force and position graphs over time, and force distribution over four full cycles of the modified trot gait. SCALER end effector climbing direction force and estimated foot position were recorded at $64.3$ Hz \red{in Section \ref{sec:dynamic_trot}}. SCALER swings LF first and then RF. The first sequence was half the stride length compared to the rest. (a) End-effector force and position of RF and LF for \red{the first} sequence in \fig{fig:trot_45}. (b) The four-full gait cycle end-effector force and position.  (c) Histogram of the RF and LF force over the four-full gait cycle with Epanechnikov kernel smoothed curve fitting. 
    \label{fig:trot_graph}}
\end{figure*}

\subsection{Model Predictive Control\label{sec:soft_mpc}}
Model Predictive Controller (MPC) disturbance rejection becomes more important when SCALER conducts dynamic motion with the pneumatic actuators since the gripper induces disruption. 
Our MPC employs linearized discrete-time dynamics based on a single rigid body model affected by ground reaction forces at the patch contacts in \eqref{eq:mpc} based on \cite{MPC}. 
\redrev{The same assumption on the trunk orientation \cite{MPC} was made with respect to the wall frame.} $A_k$ and $B_k$ are the discrete-time linearized dynamics models at a discrete time, $k$, and shown in equations \eqref{A_matrix} and \eqref{B_matrix}. $\boldsymbol{\mathcal{X}}^{\mathrm{mpc}}_{k} \in \mathbb{R}^{12}$ is the trunk state, \redrev{with Euler angles, positions, angular and linear velocity, $\mathbf{x}_{k}^{n}$ is the 3D pose of the end-effector for limb $n$, $\mathbf{f}_k^{\mathrm{mpc}}$ is a ground reaction force, and $\mathbf{\hat{I}}$ is the inertia tensor matrix all in world frame.} $m$ is the total mass of the robot's body, $[\bullet]_\times$ is defined as the skew-symmetric matrix, and $\mathbf{I}_3 \in \mathbb{R}^{3 \times 3}$ is an identity matrix, $O_{u \times w} \in \mathbb{R}^{u \times w}$ is a zero matrix.
\begin{equation}
    \boldsymbol{\mathcal{X}}_{k+1}^{\mathrm{mpc}} = A_k \boldsymbol{\mathcal{X}}_{k}^{\mathrm{mpc}} + B_k \mathbf{f}_k^{\mathrm{mpc}} + \mathbf{G} \label{eq:mpc}
\end{equation}
\begin{equation}
\label{A_matrix}
{A}_{k}=\left[\begin{array}{cccc}
O_{3 \times 6} & \mathbf{R}_{b,k}^{w\top} & O_{3 \times 3} \\
O_{3 \times 6}&O_{3 \times 3} & \mathbf{I}_3 \\
O_{6 \times 6}& O_{6 \times 3} & O_{6 \times 3}\\
\end{array}\right]
\end{equation}
\begin{equation}
\label{B_matrix}
{B}_{k}=\left[\begin{array}{ccc}
O_{6 \times 3} & . . . & O_{6 \times 3} \\
\hat{\mathbf{I}}^{-1}\left[\mathbf{x}^{1}_{k}\right]_{\times} & . . . & \hat{\mathbf{I}}^{-1}\left[\mathbf{x}^{n}_{k}\right]_{\times} \\
\mathbf{I}_3/m & . . . & \mathbf{I}_3/m
\end{array}\right]
\end{equation}
The gravity vector, $\mathbf{G}$ in \eqref{eq:gravity_vector} \red{changes based} on the climbing environment, such as a slope, vertical, or ceiling, though $\mathbf{G}$ is still a constant given a known slope. $R^{\text{Wall}}$ is the wall rotation matrix in $SO(3)$ with respect to the Earth's local tangent plane frame, and $g \in \mathbb{R}^1$ is a gravitational acceleration constant. 
\begin{equation}
\mathbf{G} = \begin{bmatrix}
O_{9 \times 9} & O_{9 \times 3} \\
O_{3 \times 9} & R^{\text{Wall}}
\end{bmatrix}
\begin{bmatrix}
O_{11 \times 1} \\
g
\end{bmatrix}
\label{eq:gravity_vector}
\end{equation}
More details are described in \cite{alex_admittance}.
We only used this closed-loop MPC for the untethered experiments with the modified trot gait. The lower control bandwidth of the admittance controller bounds the MPC performance, running at \SI{100}{\hertz}.

\section{Hardware Experiment\label{sec:experiment}}

In this section, the performance of SCALER in various climbing and ground locomotion tasks is evaluated to address the following questions:    

\begin{enumerate}
\item SCALER's capabilities for dynamic locomotion on the ground and dynamic climbing on the slope.
\item Its power-intensive motion capacities, e.g., payload, in each of these scenarios.
\item Performance in inverted environments (e.g., ceiling), including slippery terrains. 
\item Multi-modal and whole-body grasping in free-climbing.
\end{enumerate}

During these experiments, an off-board computer and external power supply were used to operate SCALER as precautionary measures. The fully untethered operation is covered in \red{Section \ref{sec:untethered}. The energy consumption stochastically fluctuates from \SIrange{50}{220}{\watt} for climbing. The highly constrained scenario, such as SCALER grasping environments, can cause internal forces, which temporarily increase power consumption. All reference trajectories in the experiments were predefined manually. The details on fully autonomous contact-rich free-climbing planning are in \cite{contact_rich}.} 

 \begin{figure}[t!]
    \centering
    \includegraphics[width=0.4\textwidth,trim={3.5cm 1cm 4cm 3.5cm},clip]{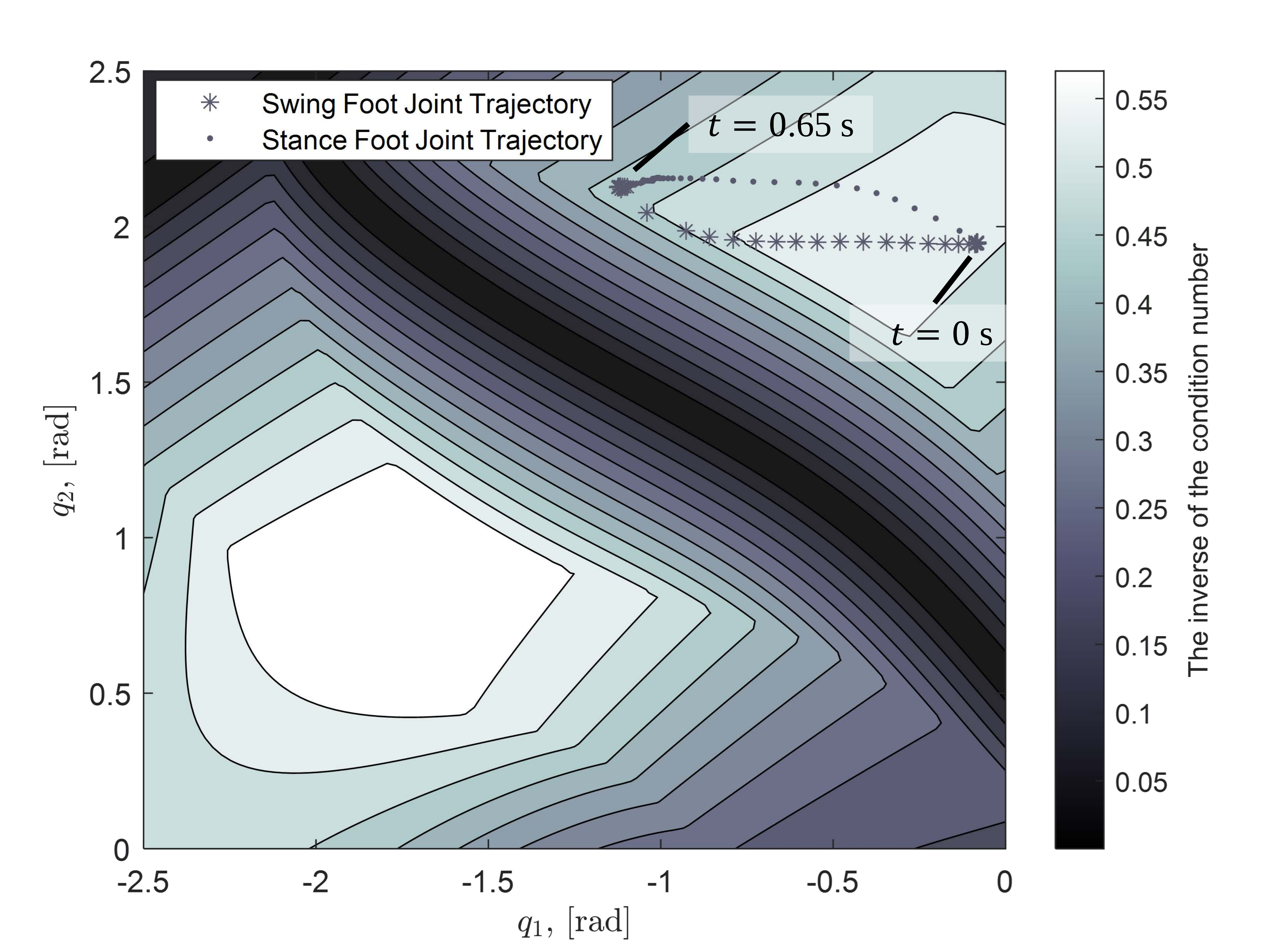}
     \caption{SCALER leg stiffness map. The inverse of the stiffness condition number at $q_0=0$ rad, where $q_i, i \in \{0,1,2\}$ are the shoulder, back, and front linkage motor joints in \fig{fig:leg_design}. The joint trajectory is from RF in the modified trot gait in Section \ref{sec:dynamic_trot}. The RF leg is in stance sequence, and the corresponding configuration is at $t=0.0~\si{\second}$ and $t=0.65~\si{\second}$ in \fig{fig:trot_45}. 
     \label{fig:stiffness_matrix}}
\end{figure}

\subsection{Dynamic Locomotion}
\subsubsection{Trot Gait on the Ground\label{sec:ground_trot}}
First, we evaluate SCALER's dynamic locomotion capability on the ground.
SCALER trotted at \SI{0.56}{\meter\per\second}, or a normalized body velocity of \SI[per-mode=reciprocal]{1.87}{\per\second}, normalized based on the body length. 
This trot velocity was recorded based on the video frame.
SCALER 3-DoF walking configuration demonstrated that SCALER can trot at a \red{comparable normalized} velocity to ground quadruped robots \red{as in Table \ref{tb:comparision}}.

\begin{figure}[t!]
  \begin{subfigure}[t]{0.24\textwidth}
    \centering
    \includegraphics[height=4cm,width=\textwidth,trim={0cm 0cm 0cm 0cm},clip]{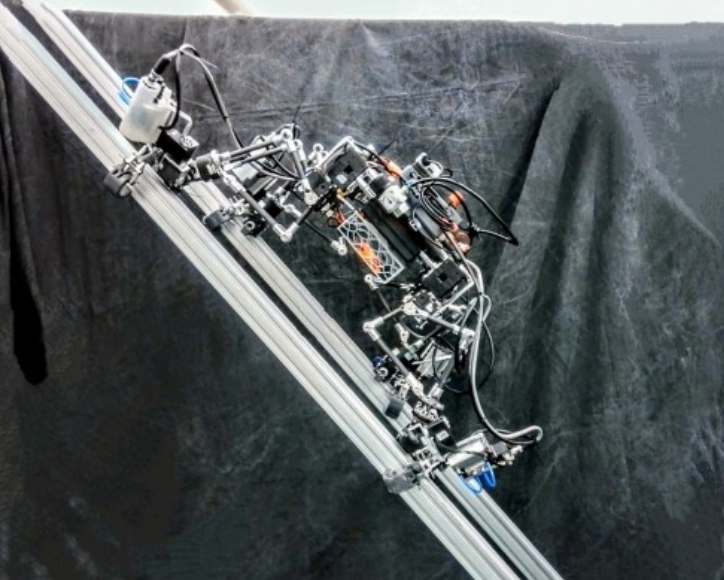}
    \caption{SCALER untethered dynamic climbing on the $45^\circ$ slippery bars slope.}
    \label{fig:untethered_pic}
  \end{subfigure}
  \hfill
  \begin{subfigure}[t]{0.24\textwidth}
    \centering
    \includegraphics[height=4cm,width=\textwidth,trim={0cm 0cm 0cm 0cm},clip]{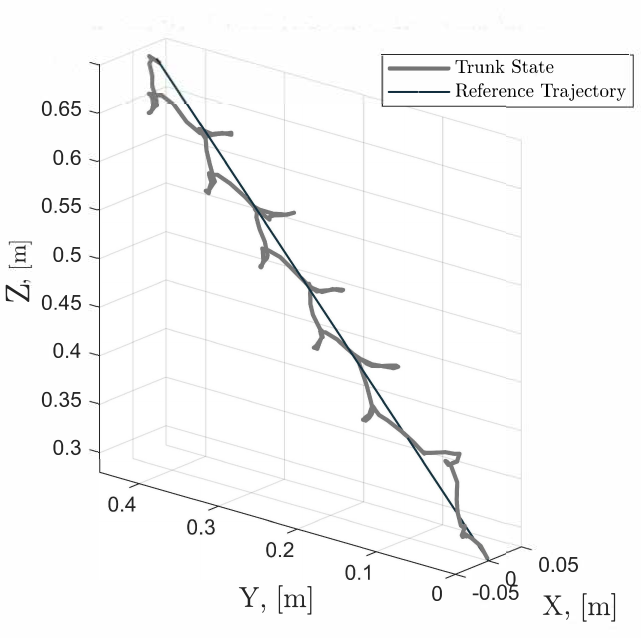}
    \caption{Trunk and reference trajectories.}
    \label{fig:untethered_3d}
  \end{subfigure}
  \caption{\redrev{Untethered SCALER modified trot gait with MPC.}}
  \label{fig:untethered_graph}
\end{figure}

\subsubsection{Dynamic Climbing with Modified Trot Gait\label{sec:dynamic_trot}}
In this experiment, we demonstrate SCALER's dynamic climbing capability using the proposed modified trot gait on the hardware. 
Dynamic climbing introduces a new set of challenges for SCALER, as mentioned in Section \ref{sec:modified_trot}. 
SCALER equipped with \red{C-GOAT} grippers performed \red{the} modified trot gait on the \SI{45}{\degree} sloped aluminum bars, as seen in \fig{fig:trot_45}. This terrain is slippery and has low friction. 

SCALER reached \SI{0.077}{\meter\per\second} or a normalized velocity of \SI[per-mode=reciprocal]{0.27}{\per\second}.
This was $13.8$ times faster than demonstrated in \cite{yusuke_scaler_2022}. \red{We observed consistent climbing overall for this environment, which was successful for $36$ consecutive steps over three separate runs. We observed no failure cases for this environment, though the dry adhesive showed wear over time.} 
The fastest climbing with a DC-actuated GOAT gripper is in Section \ref{sec:bouldering}, during which the robot was idle for \SI{57.3}{\percent} of the time because the gripper takes \SI{8.6}{\second} for the opening and closing cycle. This bottleneck is overcome by the C-GOAT grippers, which take \SI{0.2}{\second} for the cycle.

\subsubsection{Efficacy in Sag-Down Mitigation with Modified Trot Gait\label{sec:modified_trot_test}} 
This section validates the modified trot gait's effectiveness in reducing sag-down during dynamic climbing. Climbing direction, end effector force, and position were measured using F/T sensors and the T265 camera. 
The gait sequence details are in Section \ref{sec:modified_trot}, and each phase is visualized in \fig{fig:trot_45}.

 \begin{figure}[t!]
    \centering
    \includegraphics[width=0.49\textwidth,trim={0cm 0cm 0cm 0cm},clip]{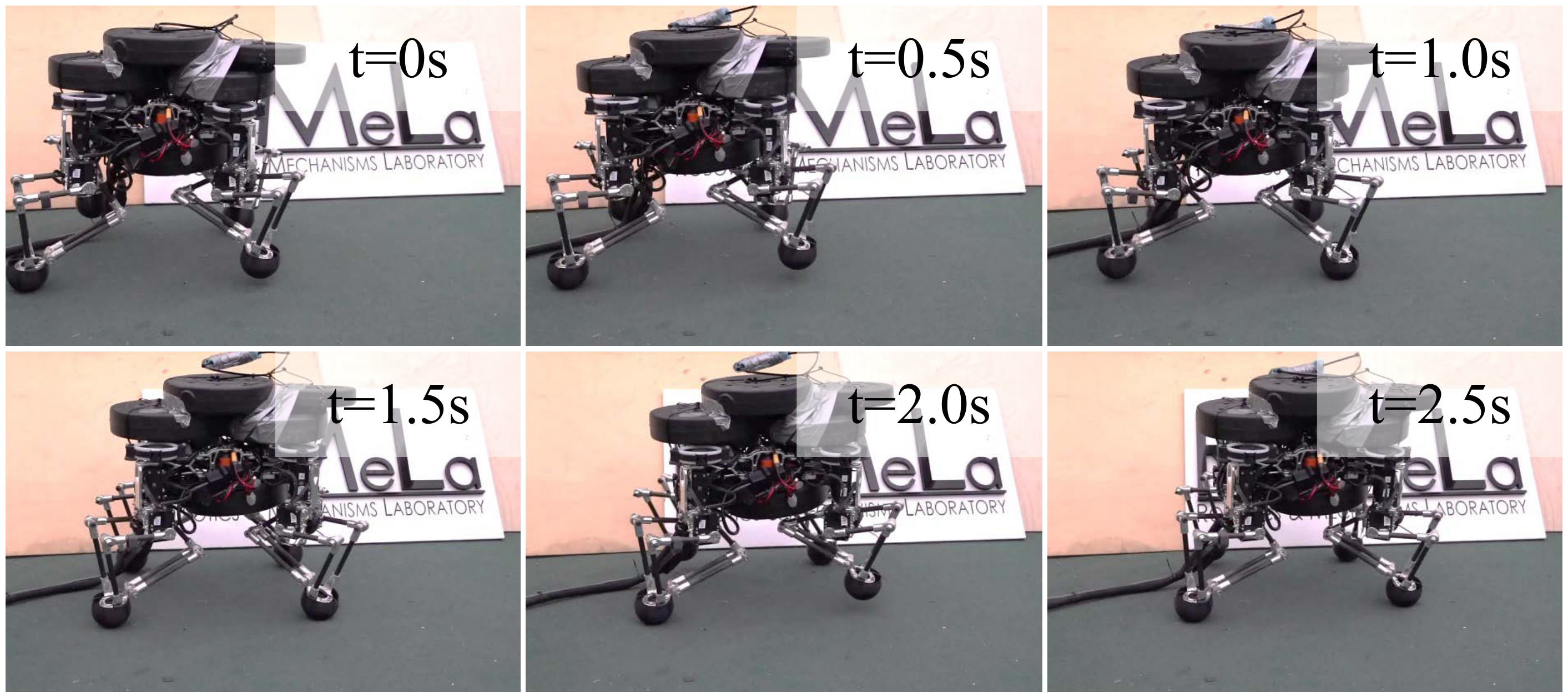}
     \caption{SCALER 3-DoF leg walking configuration with a \SI{14.7}{\kilogram} payload, equivalent to \SI{233}{\percent} of its weight.
    \label{fig:trot_weight}}
\end{figure}

In \fig{fig:trot_graph}a shows forces and positions in the climbing direction (i.e., the direction of the aluminum bars) for the first 4 phases, and \fig{fig:trot_graph}b is data for 4-full cycles of the gait. \fig{fig:trot_graph}c is the normalized force histogram over the 4-full gait cycle. \red{At $t=0$, the RF leg is in the swing sequence.}

In \textbf{Phase A}, the RF force drops to nearly zero, while that of the LF's doubles due to the \red{pushing} action, indicating successful support force and leg transition. This causes a slight position 'drop,' which is interpreted as an 'imaginary slip.'
The T265 camera estimator assumes rigid legs where the actual limb \red{has compliance}, which detects the apparent but non-actual slip. 
\textbf{Phase B} sees RF gripper force disturbances as it swings forward. \textbf{Phase C} closes grippers and maintains a 4-leg stance for stability.

The distinct force/position pattern repeats over four trot gait cycles. The two peaks in \fig{fig:trot_graph}c represent the forces during stance and swing sequences. 
Skewness in these peaks between RF and LF suggests the robot tilts toward RF.

In summary, the modified trot gait has achieved the intended gait design objectives, confirmed by both experimental data and an attached video showcasing SCALER's successful climb, compared with a failed conventional trot gait case.

 \begin{figure}[]
    \centering
    \includegraphics[width=0.99\linewidth,clip]{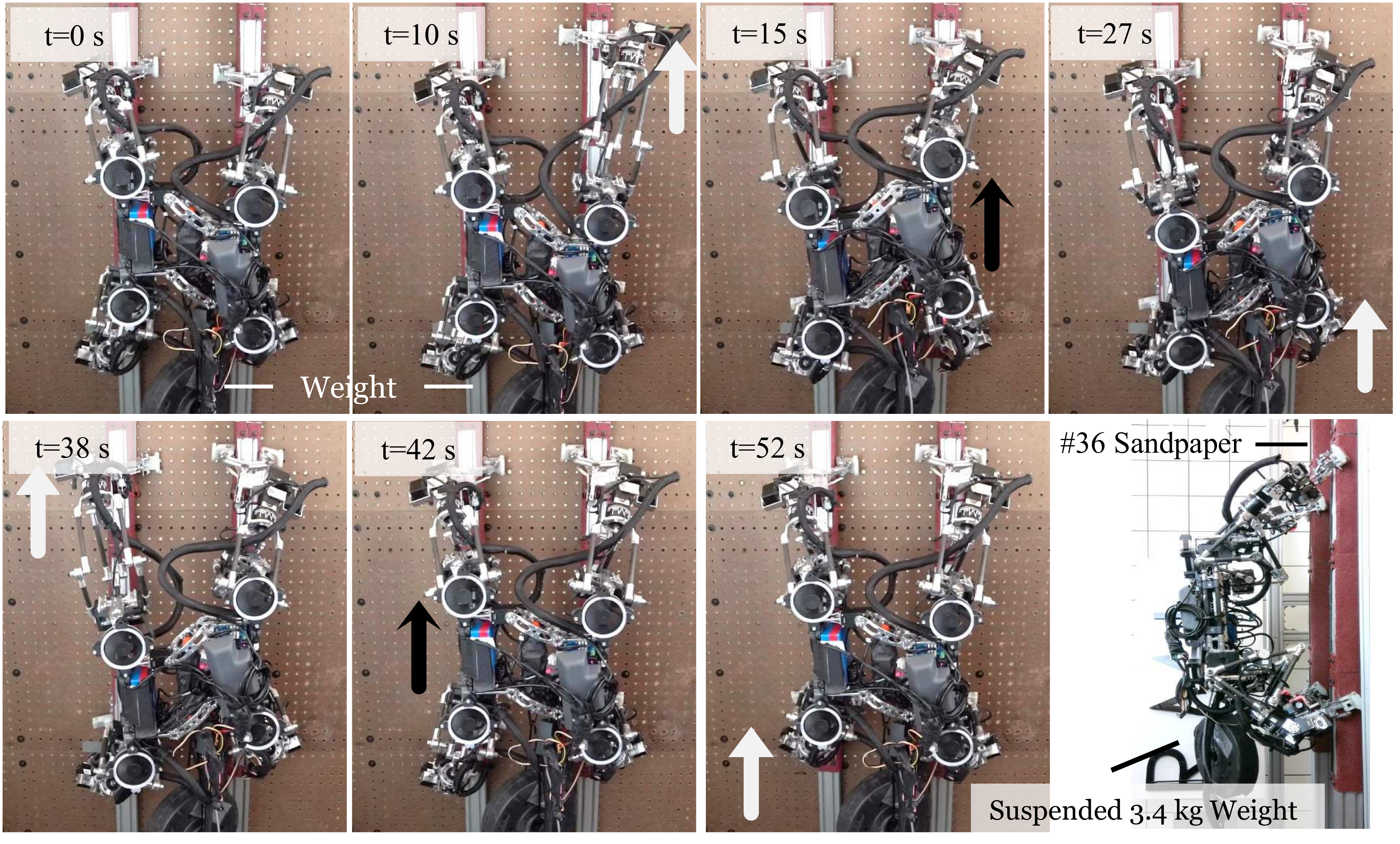}
     \caption{
     SCALER climbs a vertical wall with a \SI{3.4}{\kilogram} payload using the SKATE gait. White and black arrows indicate leg swing and body motion, respectively. The gait schedule is illustrated in \fig{fig:gait_skate}. Phase~0 begins at \SI{10}{\second}, followed by Phase~1 at \SI{15}{\second} and Phase~2 at \SI{27}{\second}, while the left side of the legs remains stationary. Then, the LF enters Phase~0. The SKATE gait sequence took \SI{52}{\second} to travel \SI{0.14}{\meter}, achieving a speed of \SI{0.16}{\meter\per\minute}.
     \label{fig:vertical_weight}}
\end{figure} 

 \begin{figure}[t!]
 \centering
          \begin{subfigure}{0.24\textwidth}
    \centering
\includegraphics[height=3.5cm, width=\textwidth,trim={0cm 0cm 0cm 0cm}, clip]{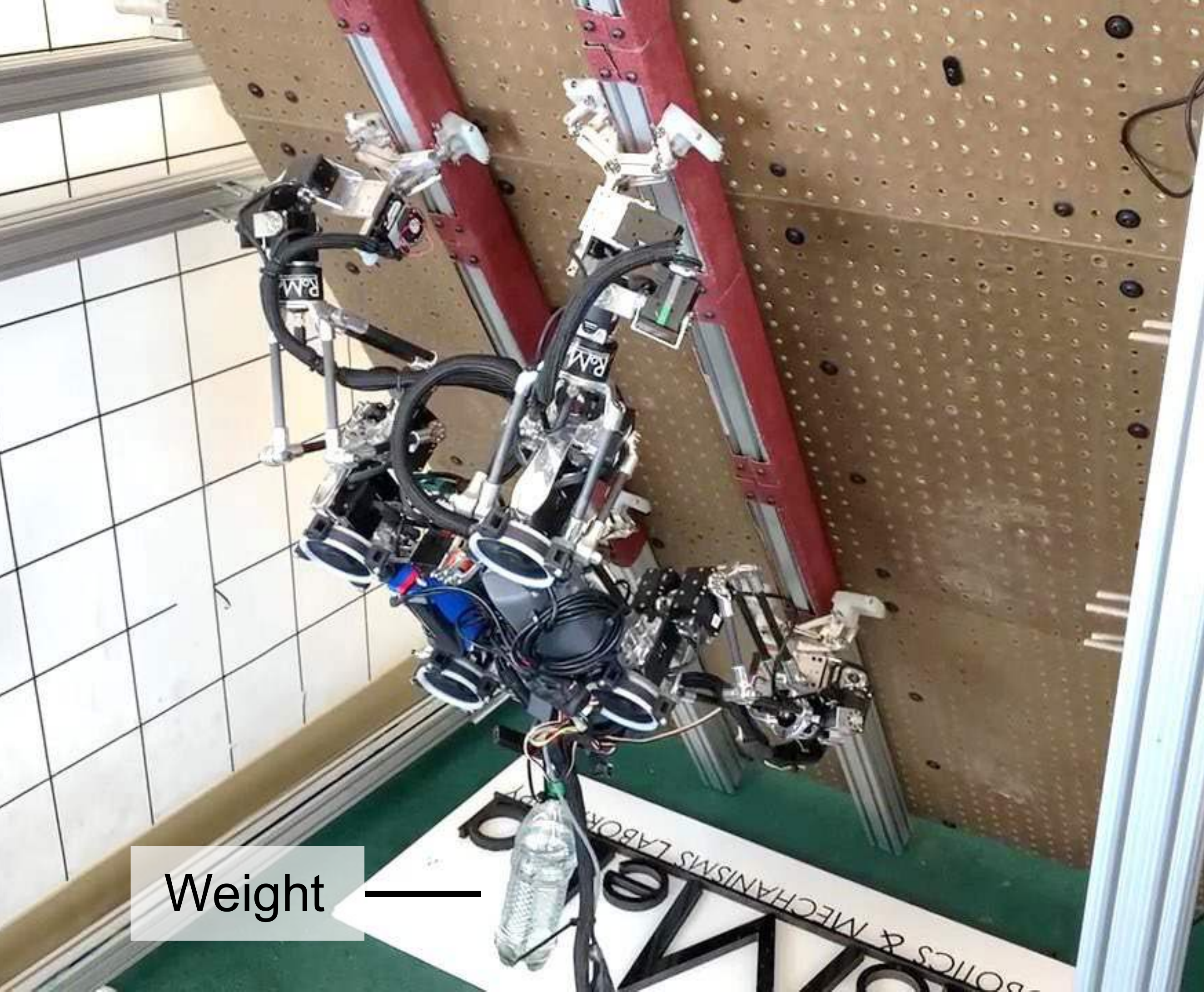}
\caption{Climbing on an overhang wall at \SI{125}{\degree}.\label{fig:overhang}}
    \end{subfigure}
         \begin{subfigure}{0.24\textwidth}
\includegraphics[height=3.5cm,width=\textwidth,trim={0cm 0cm 0cm 0cm}, clip]{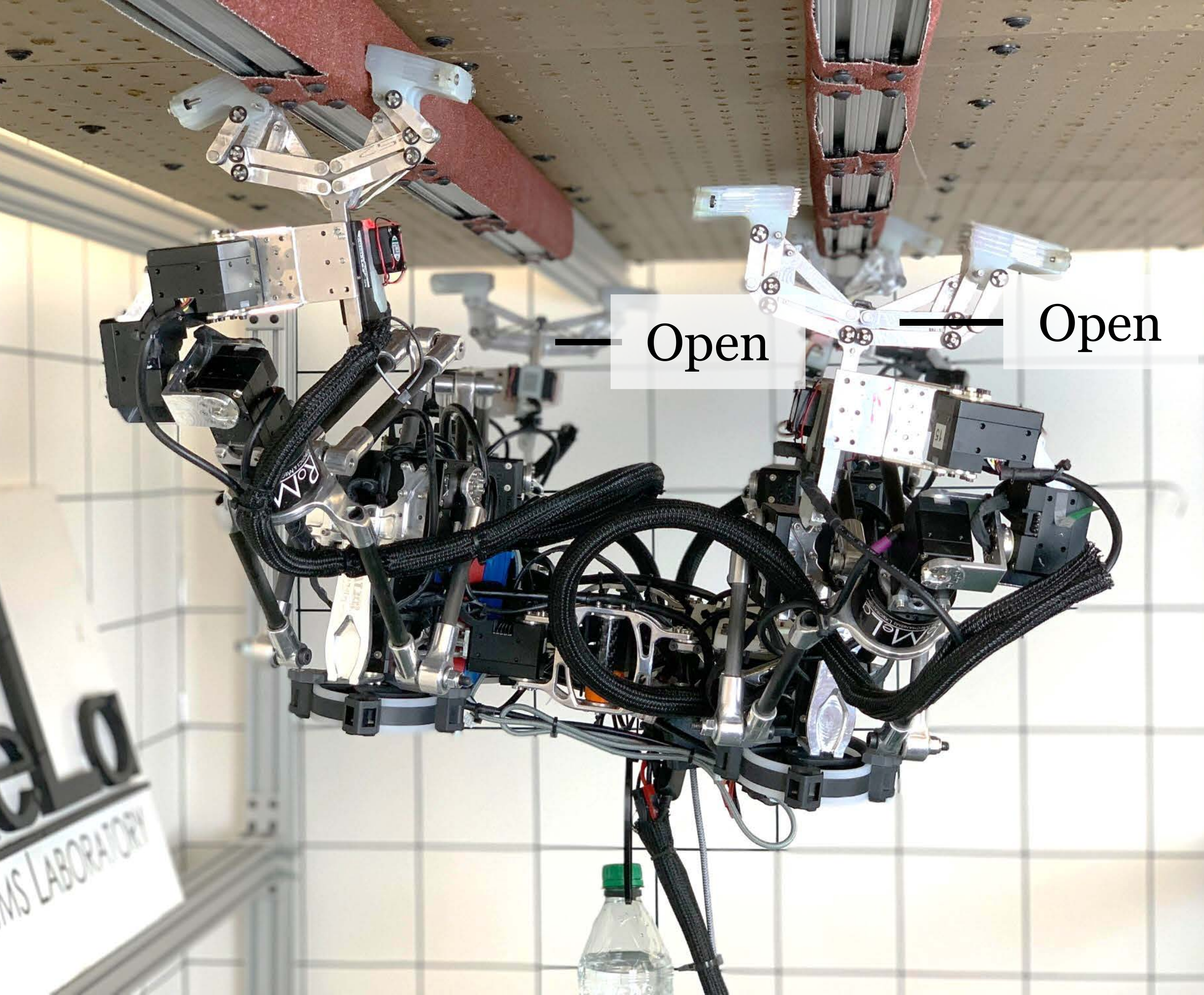}
\caption{Two finger support on the ceiling.\label{fig:ceiling_two}}
    \end{subfigure}
     \caption{SCALER under inverted environments. A weight hangs from the robot to indicate the direction of gravity. 
     }\label{fig:inverted}
\end{figure}

\subsubsection{Limb \red{Stiffness} Model and Force Analysis\label{sec:compliance_model_test}}
The VJM model in Section \ref{sec:compliance_model} is calculated, and the stiffness map is in \fig{fig:stiffness_matrix}, which is the inverse of the stiffness condition number.
The valley in the stiffness map visualizes the stiffest regions \cite{stiffness_identification}. 
The joint trajectory recorded from the experiment in Section \ref{sec:dynamic_trot} is plotted on the stiffness map in \fig{fig:stiffness_matrix}. \red{This indicates that the leg stiffens when pushing in the modified trot gait.}
The stiffness of each joint was measured using a Mitutoyo HD-12 AX, which applied a force on the F/T sensors to deform the joint.

The \red{pushing} force during \textbf{Phase A} in Section \ref{sec:dynamic_trot} is estimated to be \SI{31.7}{\newton} using VJM, given the commanded \SI{0.16}{\meter} pushing distance in the climbing direction. The \red{pushing} force generated from the LF in \textbf{Phase A} in \fig{fig:trot_graph}a is \SI{37.3}{\newton} on average. 
Our stiffness model provides adequate stiffness estimation for the SCALER's climbing tasks, with an average of \SI{17.6}{\percent} modeling error for this modified trot gait case.

 \begin{figure}[t!]
    \centering
    \includegraphics[width=0.4\textwidth,trim={0cm 0cm 0cm 0.8cm},clip]{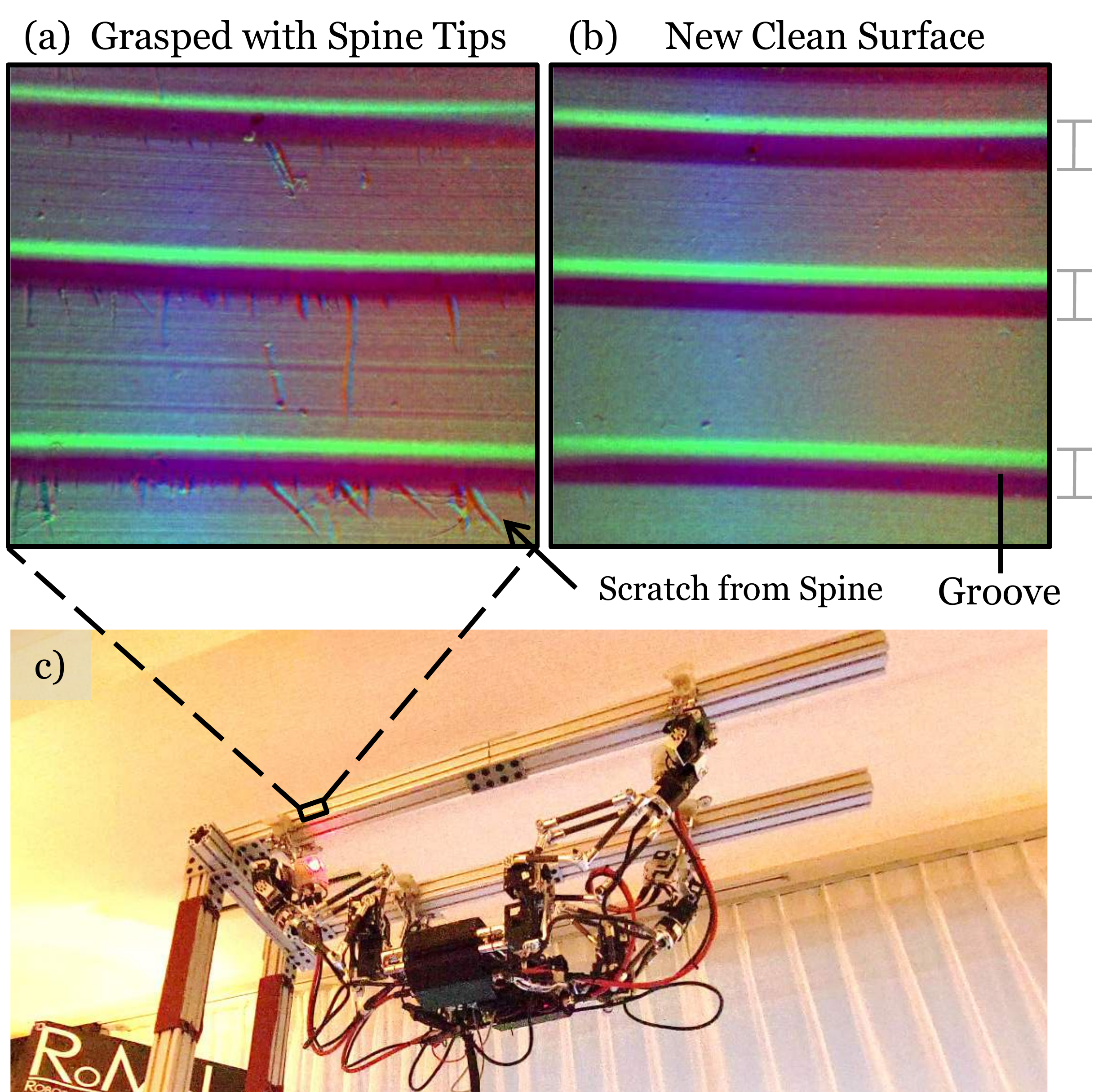}
     \caption{
     Tactile images capturing the surface of the T-slotted aluminum bar and its interaction with SCALER's spine tips. (a) The tactile image shows scratches from spines compared to the new aluminum bar shown in (b). (c) SCALER with spine tip GOAT grippers can travel upside down on a slippery surface due to this spine effect. 
     \label{fig:ceiling_bar}}
\end{figure}

 \begin{figure}[]
 \centering
             \begin{subfigure}{0.33\linewidth}
\includegraphics[height=2.cm, width=\textwidth, trim={0cm 0cm 0cm 0cm}, clip]{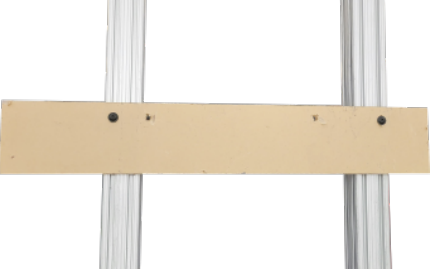}
\caption{An obstacle plate placed across the two aluminum bars on the \SI{45}{\degree} slope.\label{fig:multi-modal-env}}
    \end{subfigure}
    \hfill
    \begin{subfigure}{0.65\linewidth}
\includegraphics[height=2.5cm, width=\textwidth, trim={0cm 0cm 0cm 0cm}, clip]{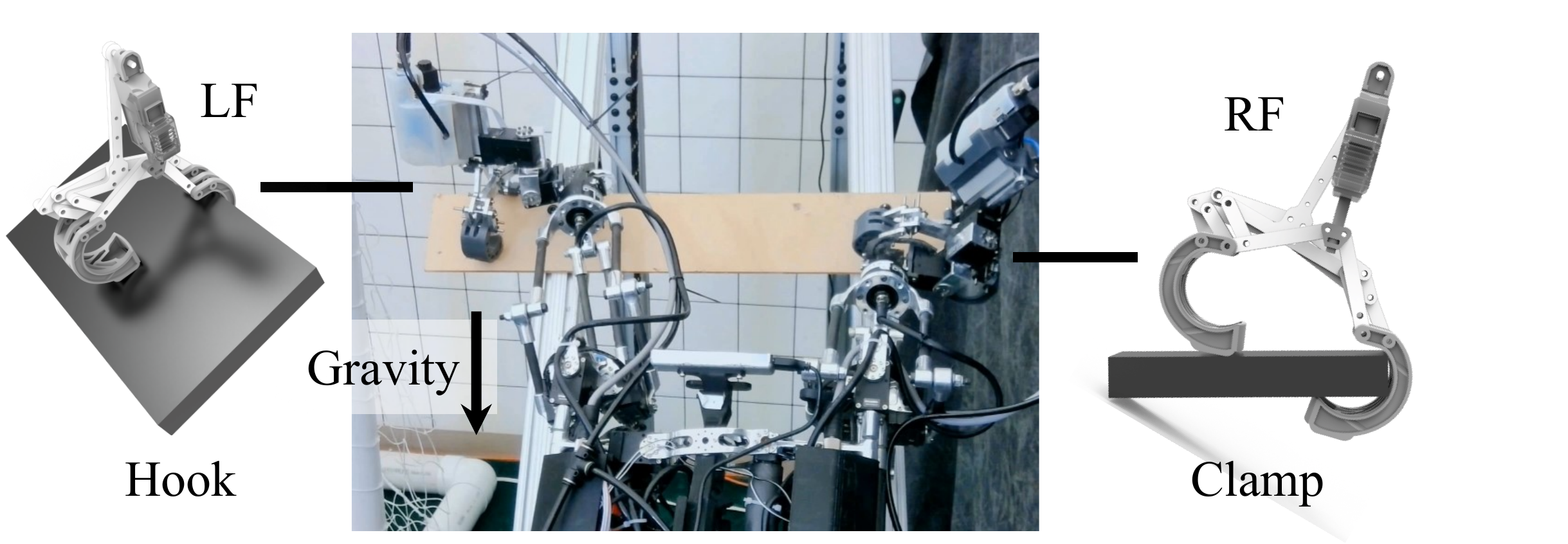}
\caption{A top view when SCALER grasps the obstacle while climbing the environment.   
\label{fig:multi-modal-top}}
    \end{subfigure}
     \caption{The environment and the top view for SCALER climbing using three modes of grasping: pinch, hook, and clamp to overcome a thin plate obstacle. 
     }\label{fig:multi-modal-env-top}
\end{figure}

 \begin{figure*}[t!]
    \centering
    \includegraphics[width=0.9\textwidth,trim={0cm 1.3cm 1cm 0.7cm}, clip]{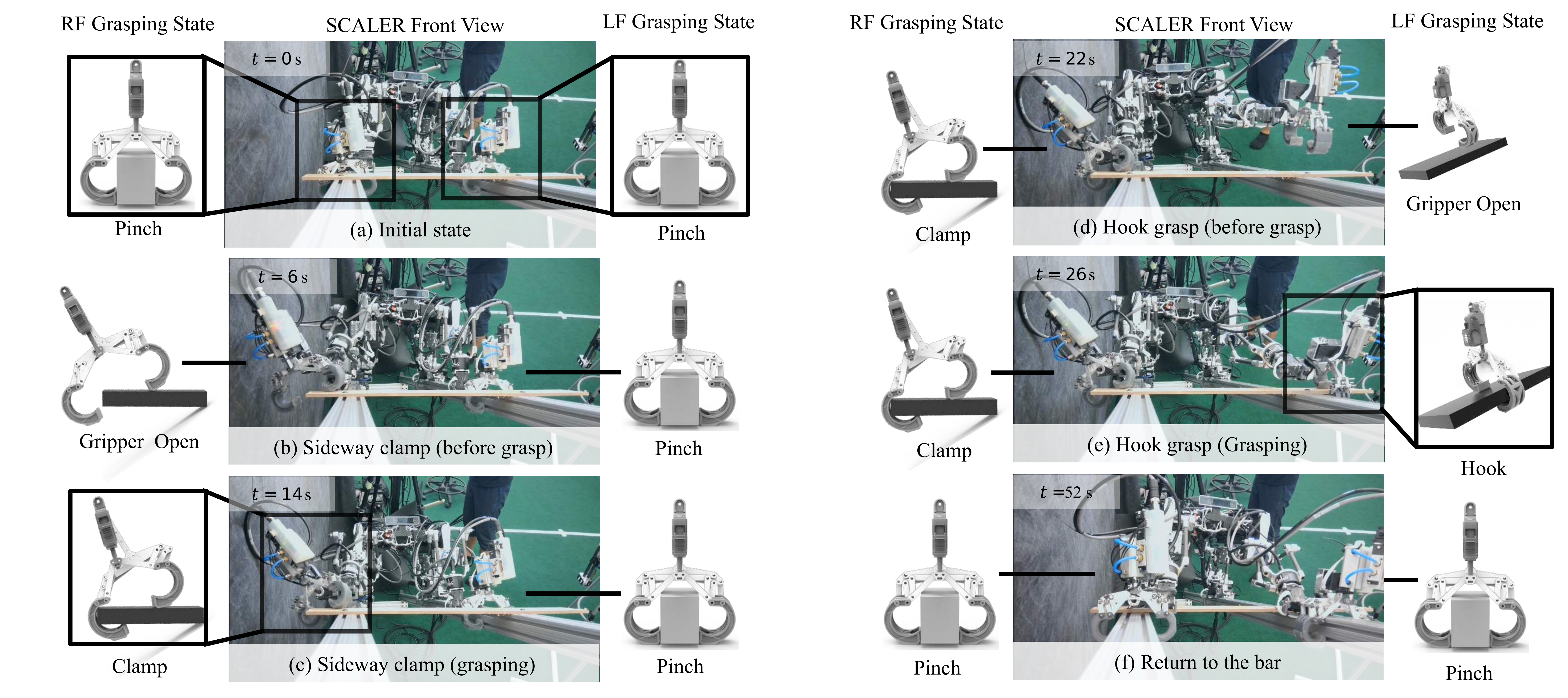}
     \caption{Front views of SCALER climbing the environment with an obstacle, a thin plate in the way, shown in \fig{fig:multi-modal-env-top}. SCALER utilized pinch, clamp, and hook grasp modes to overcome the obstacle so that it can continue to climb. RF and LF grasping states render the SCALER's gripper grasping modes.
     (a) SCALER grasping aluminum bars before entering the obstacle region. (b), (c) SCALER clamped the thin plate from the side. (d), (e) SCALER hooks the gripper on the thin obstacle plate. (f) SCALER moved both front arms back to the bar. It took \SI{60}{\second} to reach the state (f) from (a). 
     \label{fig:multi-modal}}
\end{figure*}

\subsubsection{Untethered Climbing \red{using C-GOAT} with MPC\label{sec:untethered}} 
SCALER can operate fully untethered when equipped with the computing and power modules. This represents SCALER's ability to go off-grid and climb with no range limit due to the tethering. \fig{fig:untethered_graph}a shows the SCALER climbing on a \SI{45}{\degree} slippery slope untethered with MPC presented in Section \ref{sec:soft_mpc}, and \fig{fig:untethered_graph}b is the trunk states from the T265 camera. \red{The experiment was repeated once. MPC showed improvement in the tracking in the $x$-axis in \fig{fig:untethered_graph}b particularly (Open-loop \SI{35}{\milli\meter}, MPC \SI{1.6}{\milli\meter} offset after $12$ steps).}

\subsection{Payload Capacity}
\subsubsection{Ground Trot with Payload\label{ground_payload}}
To benchmark SCALER's load capacity compared to other quadruped robots, we attached a payload of a \SI{3.4}{\kilogram} weight to its belly and an \SI{11.3}{\kilogram} weight on the top, corresponding to $2.33$ times its own weight.
The trotting velocity of SCALER with the \SI{14.7}{\kilogram} payload was at \SI{0.13}{\meter\per\second}, or a normalized velocity of \SI[per-mode=reciprocal]{0.33}{\per\second}.
This payload was the maximum SCALER can trot, but it struggled to move in a straight direction. 
A payload of \SI{10.2}{\kilogram} yielded a speed of \SI{0.25}{\meter\per\second}, and SCALER was able to trot in a straight line.
A normalized work capacity, defined as $\mathcal{WC} = \mathcal{V} \times \mathcal{P}$, where $\mathcal{V}$ and $\mathcal{P}$ represent normalized maximum body velocity and payload, respectively, provides a quantitative metric for SCALER's mechanical efficiency of legged robots \cite{quad_comparison}.
SCALER outperformed other robots with the normalized workload capacity of $3.88$ \cite{quad_comparison}.

\subsubsection{Vertical Wall SKATE Gait Climbing with Payload\label{sec:vertical_payload}}
In this experiment, we evaluated SCALER's mobility in vertical climbing with a payload. 
We tested SCALER on a vertical wall outfitted with a straight \SI{0.05}{\meter} wide rail covered by $\#36$ sandpaper to simulate rock surfaces. Each gait sequence lifted the body by \SI{0.075}{\meter} and achieved a speed of \SI{0.16}{\meter\per\minute} while carrying a \SI{3.4}{\kilogram} payload. This accounts for \SI{35}{\percent} of SCALER's total weight. Suspended payloads pose challenges due to pendulum dynamics. The black and white arrows in \fig{fig:vertical_weight} indicate which leg and body motions are compared to the previous frame, respectively. The same experiments were conducted twice for consistency. SCALER demonstrated vertical climbing with a payload, indicating the SCALER's power-intensive motion and the GOAT gripper's capabilities.

\subsection{Inverted Environments\label{sec:inverted}}

In inverted environments, such as overhangs and ceilings, the effectiveness of the grippers is crucial, as they must exert a sufficient contact force to counteract the gravitational pull.

\subsubsection{Overhang Climbing\label{sec:overhang}}
SCALER climbed on an overhanging wall tilted at \SI{125}{\degree} towards the robot in \fig{fig:overhang}. A weight of \SI{0.5}{\kilogram} was suspended from the robot's body to indicate the direction of gravity. 
We conducted overhang and vertical climbing in this environment three times each.
In this environment, gripper weight helped the spine fingertips break contact when lifting limbs. Although there was no failure, there was a case in vertical climbing where the back limb gripper's finger stuck due to the spine, and the stride distance for this limb was shorter than the commanded stride.

\subsubsection{Upside-down Ceiling Walk On Rough Surfaces\label{sec:upsidedown}}
When SCALER operates in an upside-down orientation on the ceiling, SCALER uses spine fingertip GOAT grippers to \red{grasp the rails on} the ceiling in \fig{fig:fig1}a. The $0.5$ kg weight is suspended from SCALER. \red{This experiment was repeated twice.} 

The effects of the fingertip spine enhance gripping as data presented in \cite{GOAT} for the GOAT gripper. These effects enable SCALER to maintain a stable grip on the ceiling, even when only two grippers are grasping, as shown in \fig{fig:ceiling_two}. 
Furthermore, these finger spines allow SCALER to remain attached to the ceiling or a vertical wall even when the system is powered off. The spine effects can generate sufficient shear force with minimal normal force as long as the needle tips remain engaged with the surface microcavities. 

\begin{figure*}[]
    \centering
    \includegraphics[width=0.99\textwidth,clip]{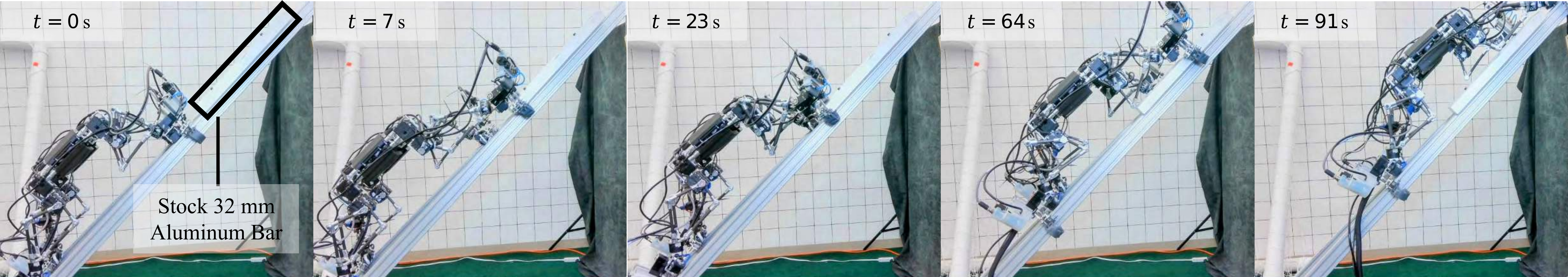}
     \caption{The whole-body approach with side-pull techniques. The obstacle plate, marked in a black box, is too thick to clamp, but with the whole-body approach, SCALER can side-pull to stabilize itself. The force graph indicates a significant sideways force generated by SCALER in \fig{fig:whole-body-graph}. SCALER started from the bars and conducted the sidepull whole-body climbing from $t=7~\si{\second}$. SCALER reached the top of the obstacle and returned to the bar at $91~\si{\second}$.
     \label{fig:whole-body}}
\end{figure*}

 \begin{figure}[]
    \centering
    \includegraphics[width=0.99\linewidth,clip]{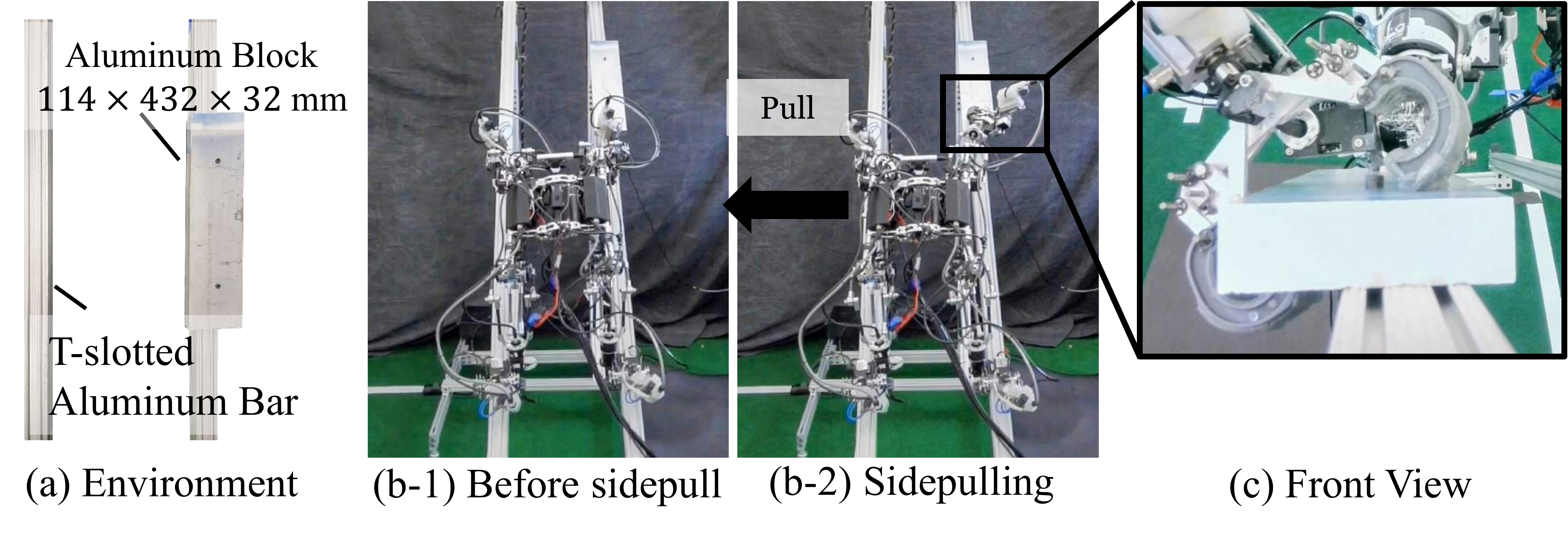}
     \caption{SCALER climbing the aluminum bars with an obstacle plate in the way. (a) The test environment has a stock aluminum plate attached to one side of the bars. (b-1) A top view before sidepull. (b-2) when sidepulling. (c) A front view of the sidepulling GOAT gripper.
     \label{fig:whole-body-top}}
\end{figure} 

\subsubsection{Upside-down Ceiling Walk on Slippery Metal Surface\label{sec:upsidedown_low_friction}}
In this experiment, we evaluate the spine GOAT grippers on a slippery aluminum surface as pictured in \fig{fig:ceiling_bar}a,b with Gelsight Mini tactile sensors \cite{yuan2017gelsight}.
SCALER successfully walked on a T-slotted aluminum bar upside down in \fig{fig:ceiling_bar}c. The spines engaged with micro textures on the aluminum bar surface, generating enough shear force to support SCALER upside down on the slippery surface.
The tactile images shown in \fig{fig:ceiling_bar}a,b compare the aluminum bar surface grasped with the spine and a new, unused, identical aluminum bar.
The spines caused scratches at the bottom edge of \red{the} grooves \red{in \fig{fig:ceiling_bar}a}. These etches indicate that the spines are anchored in the grooves to generate shear forces, whereas spines making contact on the flat aluminum surfaces do not contribute to this shear force significantly. 
 \red{Since the environment is very slippery, there was no case when the gripper stuck like with sandpaper environments in Section~\ref{sec:overhang} over four episodes.}

\subsection{Multi-Modal and Discrete Terrains Free-Climbing\label{sec:obstacle_traverse}}

This section focuses on SCALER's loco-grasp free-climbing capabilities in directionally continuous (\fig{fig:environment}c) and discrete environments (\fig{fig:environment}b).  

\subsubsection{Climbing over Obstacles with Multi-Modal Grasping\label{sec:multi_modal_climb}}
We experiment with SCALER's ability to overcome obstacles while climbing through multi-modal grasping. SCALER leverages the \red{C-GOAT} grippers to climb by grasping obstacles.
A \SI{6.5}{\milli\meter} plate obstacle was placed across the T-slotted aluminum bar oriented at \SI{45}{\degree} as shown in \fig{fig:multi-modal-env}. SCALER clamps and hooks the obstacle plate from the side and top to continue climbing, as captured in \fig{fig:multi-modal-top}. This capability enhances the SCALER's traversability since it can utilize obstacles instead of avoiding them. 

\fig{fig:multi-modal} shows the front view of SCALER conducting multi-modal grasping with rendering of the RF and LF grasping states. 
\fig{fig:multi-modal}b, d pictures that SCALER orients and aligns the grippers to clamp and hook the plate. \red{Obstacle traversing behavior is done with manual trajectory here. Out of $10$ trials, the clamp and hook were successful $9$ and $4$ times, respectively. The hook tended to fail because the back of the finger caused unexpected contact with the plate, and one episode was fully successful in one shot.}

The success of the multi-modal grasping method signifies SCALER's dexterous capabilities, making it more versatile in traveling complex terrains than traditional climbing robots.

 \begin{figure}[t!]
    \centering
    \includegraphics[width=0.4\textwidth, trim={6cm 0cm 6cm 2cm},clip]{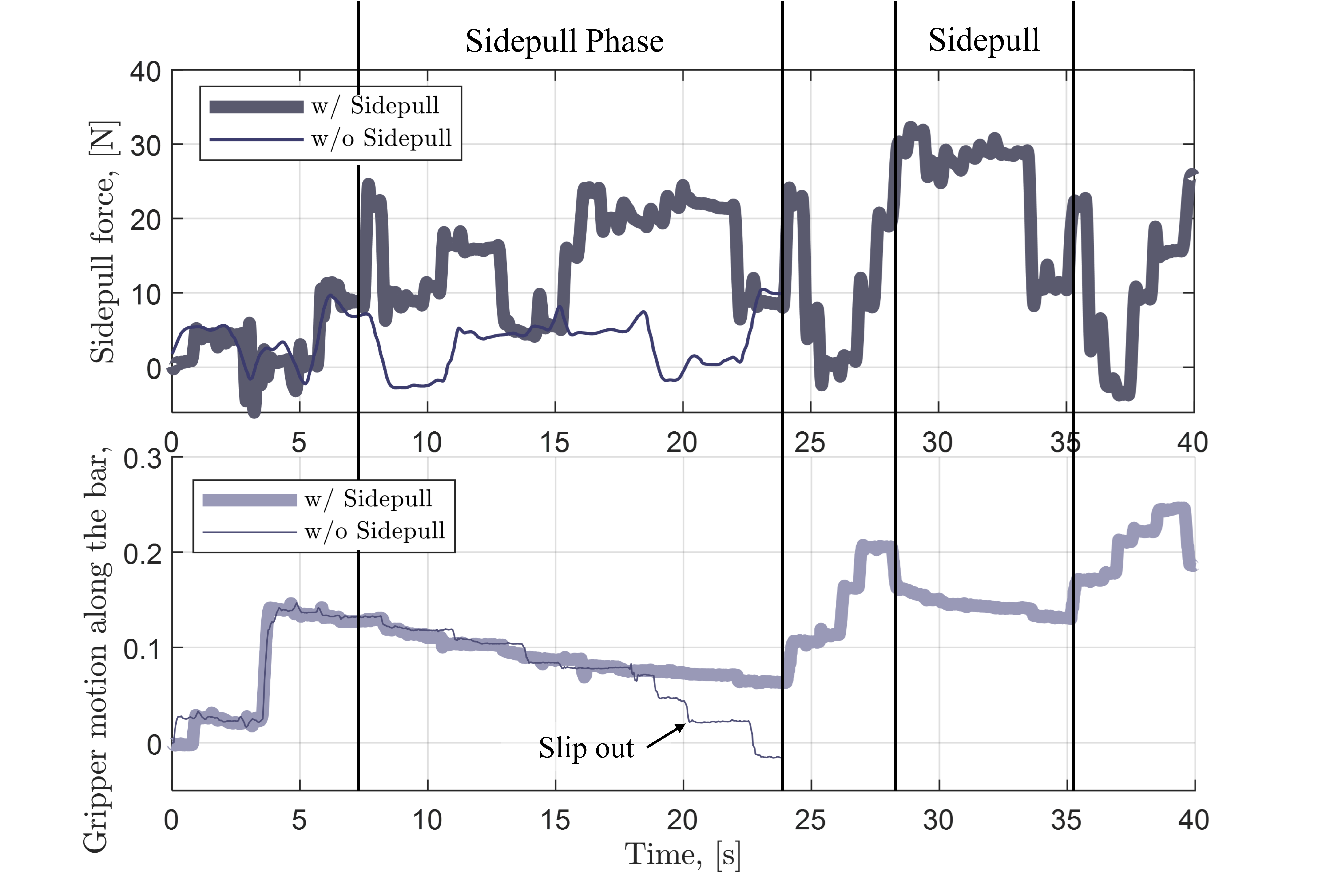}
     \caption{Sidepull forces and sidepulling gripper motions in the climbing direction. Without sidepull, the gripper slid out, and SCALER could not continue climbing. With sidepull, SCALER climbed and back to the bar at $91~\si{\second}$ in \fig{fig:whole-body}.  
     \label{fig:whole-body-graph}}
\end{figure} 

 \begin{figure}[t!]
    \centering
    \includegraphics[width=0.49\textwidth,clip]{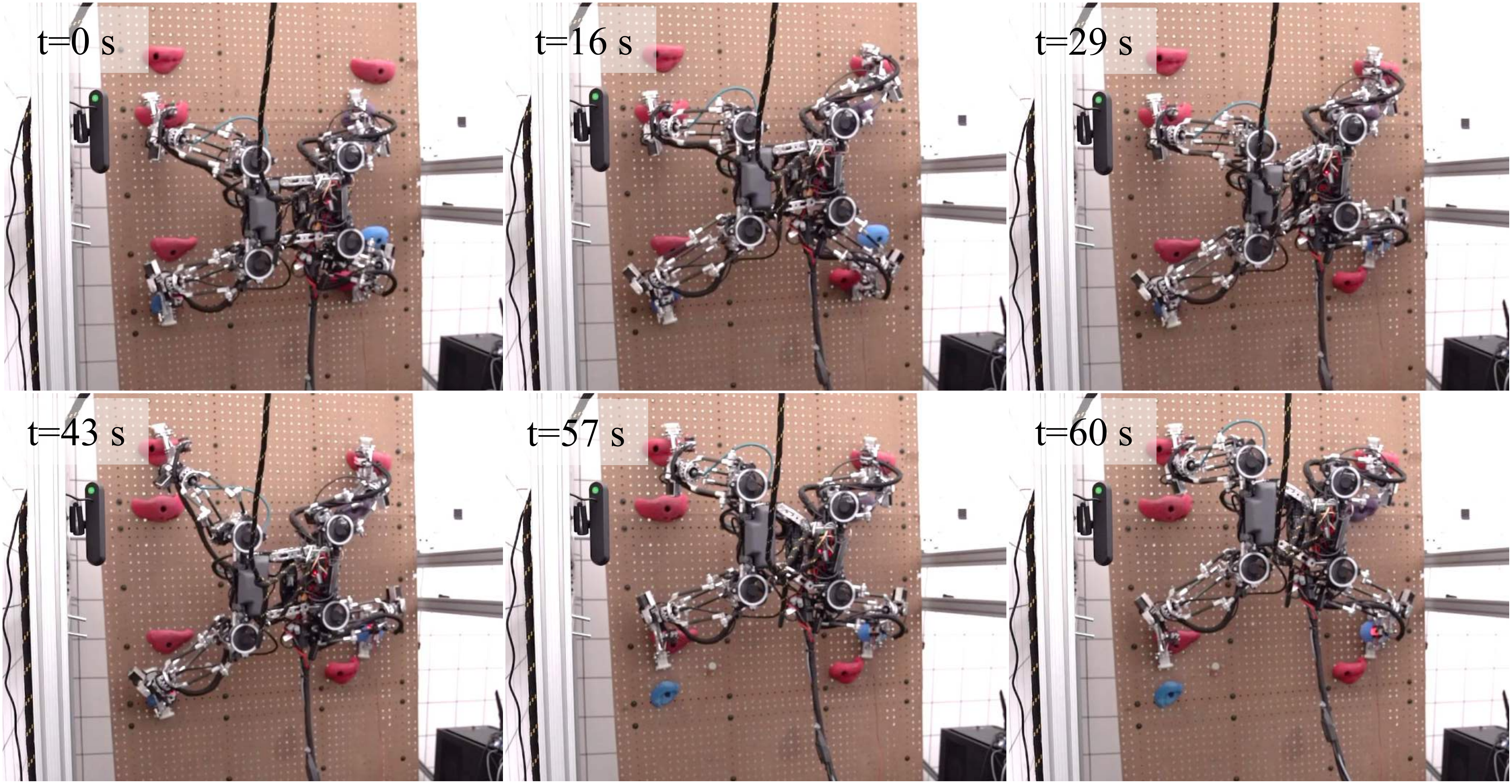}
     \caption{
     SCALER climbs a bouldering wall after each leg moves forward. The wall is at \SI{90}{\degree}. SCALER climbed up \SI{0.35}{\meter} at \SI{0.35}{\meter\per\minute}.
     \label{fig:bouldering}}
\end{figure}

\begin{table*}[t!]
\centering
\begin{threeparttable}
\caption{Performance of quadruped robots on the ground comparisons. \cite{quad_comparison}. \label{tb:comparision}}
\begin{tabular}{lcccccccc}
\hline
\multicolumn{1}{c}{Robot} & \begin{tabular}[c]{@{}c@{}}DoF \\ per Limb\end{tabular} & \begin{tabular}[c]{@{}c@{}}Weight \\ (\si{\kilogram})\end{tabular}& \begin{tabular}[c]{@{}c@{}}Normalized Payload\\ (Unitless)\end{tabular} & \begin{tabular}[c]{@{}c@{}}Max Payload\\ (\si{\kilogram})\end{tabular} & \begin{tabular}[c]{@{}c@{}}Normalized Speed\\ (\si[per-mode=reciprocal]{\per\second})\end{tabular} & \begin{tabular}[c]{@{}c@{}}Ground Velocity\\ (\si{\meter\per\second})\end{tabular} & \begin{tabular}[c]{@{}c@{}}Normalized Workload\\ (Unitless)\end{tabular} \\ \hline
SCALER (Walking) & $3$ & $6.3$ & $2.33$ & $14.7$ & $1.87$ & $0.56$ & $3.88$ \\
ANYmal & $3$ & $30$ & $0.33$ & $10$ & $1.0$ & $0.8$ & $0.33$ \\
Stanford Doggo \cite{stanford_doggo} & $2$ & $4.8$ & N/A & N/A & $2.14$ & $0.9$ & $3.2$ \\
Titan XIII & $3$ & $5.7$ & $0.89$ & $5$ & $4.29$ & $0.9$ & $3.79$ \\
SPOT & $3$ & $30$ & $0.47$ & $14$ & $1.45$ & $1.6$ & $0.68$ \\
Mini Cheetah & $3$ & $9$ & *$1$ & *$9$ & $6.62$ & $2.45$ & N/A \\ \hline
\end{tabular}
\begin{tablenotes}
      \footnotesize
      \item Normalized Velocity: body length divided by the body velocity. Normalized Payload: body weight divided by payload. Normalized Work Capacity: normalized speed $\times$ normalized payload, a mechanical efficiency metric \cite{quad_comparison}. \red{*The value is theoretical.}
\end{tablenotes}
\end{threeparttable}
\end{table*}


\begin{table*}[t!]
\centering
\begin{threeparttable}
\caption{Performance of multi-legged robots for climbing comparisons. \label{tb:climbing}}
\begin{tabular}{lccccccc}
\hline
\multicolumn{1}{c}{Robot} & \begin{tabular}[c]{@{}c@{}}DoF per Leg \\ Leg / Gripper \\ (Passive DoF)\end{tabular} & \begin{tabular}[c]{@{}c@{}}Climbing\\ Gravity\end{tabular} & \begin{tabular}[c]{@{}c@{}}Climbing\\ Velocity\\ \si{\meter\per\minute}\end{tabular} & \begin{tabular}[c]{@{}c@{}}Normalized\\ Velocity\\\si[per-mode=reciprocal]{\per\minute}\end{tabular} & \begin{tabular}[c]{@{}c@{}}Climbing\\ Payload\\ \si{\kilogram} \end{tabular} & Gripper & Environment \\ \hline
\begin{tabular}[c]{@{}l@{}}SCALER\end{tabular} & $6 / 1$ & \begin{tabular}[c]{@{}c@{}}Earth, $g$\end{tabular} & $4.62$ & $13.2$ & \begin{tabular}[c]{@{}c@{}}$3.4$ \end{tabular} & \begin{tabular}[c]{@{}c@{}}Two finger\\ Spine, Dry adhesive \\ GOAT Gripper\end{tabular} & \begin{tabular}[c]{@{}c@{}}Discrete \fig{fig:environment}b,\\ Overhang \fig{fig:environment}c,\\ Ceiling \fig{fig:environment}c\\ Slippery surface\end{tabular} \\
\rowcolor[HTML]{EFEFEF} 
\begin{tabular}[c]{@{}l@{}}LEMUR 3 \cite{lemur3}\end{tabular} & $7 / 2$ & \begin{tabular}[c]{@{}c@{}}Mars, Moon, Zero\\ $0.38$, $0.17$, $0$ $g$\end{tabular} & $2.7e^{-3}$& $6.7e^{-3}$ & N/A & \begin{tabular}[c]{@{}c@{}}Radial Velcro-like\\ Spine Gripper\end{tabular} & \begin{tabular}[c]{@{}c@{}}Continuous \fig{fig:environment}a\\ Rough\end{tabular}\\
\begin{tabular}[c]{@{}l@{}}HubRobo \cite{hubrobo}\end{tabular} & \begin{tabular}[c]{@{}c@{}}$3(3)/ 1$\end{tabular} & \begin{tabular}[c]{@{}c@{}}Mars\\ $0.38$ $g$\end{tabular} & $0.17$ & $0.57$ & N/A & \begin{tabular}[c]{@{}c@{}}Radial Passive \\ Spine Gripper\end{tabular} & Discrete \fig{fig:environment}b\\
\rowcolor[HTML]{EFEFEF} 
\begin{tabular}[c]{@{}l@{}}Slalom \cite{slalom}\end{tabular} & \begin{tabular}[c]{@{}c@{}}$4(3)/ 0$\end{tabular} & \begin{tabular}[c]{@{}c@{}}Slope: $30^\circ$ \\ $\sim 0.5$ $g$\end{tabular} & $6$\ & $17.1$ & N/A & \begin{tabular}[c]{@{}c@{}}Dry Adhesive\\ EPDM Rubber\end{tabular} & Flat Solid/Soft \fig{fig:environment}a\\ 
RiSE \cite{rise_bd} & \begin{tabular}[c]{@{}c@{}}$2 / 0$\end{tabular} & \begin{tabular}[c]{@{}c@{}}Earth, $g$\end{tabular} & $15.0$ & $40.0$ & $1.5$ & Spine Array &  \begin{tabular}[c]{@{}c@{}}Continuous \fig{fig:environment}a\\ Rough\end{tabular} \\
\rowcolor[HTML]{EFEFEF} 
Bobcat \cite{bobcat} & \begin{tabular}[c]{@{}c@{}}$2 / 0$\end{tabular} & \begin{tabular}[c]{@{}c@{}}Earth, $g$\\ on strap\end{tabular} & $10.5$ & $22.8$ & N/A & Spine Array & \begin{tabular}[c]{@{}c@{}}Continuous \fig{fig:environment}a\\ Wire Mesh\end{tabular} \\
\begin{tabular}[c]{@{}l@{}}Climbot \cite{6094406}\end{tabular} & \begin{tabular}[c]{@{}c@{}}$2 / 1$ \end{tabular} & \begin{tabular}[c]{@{}c@{}}Earth, $g$\end{tabular} & $2.2$& N/A & N/A & two-fingered Gripper & \begin{tabular}[c]{@{}c@{}}Pipe \fig{fig:environment}c\end{tabular} \\
\rowcolor[HTML]{EFEFEF} 
\begin{tabular}[c]{@{}l@{}}MARVEL \cite{marvel}\end{tabular} & \begin{tabular}[c]{@{}c@{}}$3(3)/ 0$\end{tabular} & \begin{tabular}[c]{@{}c@{}}Earth, $g$\end{tabular} & $42.0$ & $127.2$ & $2$ & \begin{tabular}[c]{@{}c@{}}Electropermanent\\ Magnetic\end{tabular} & \begin{tabular}[c]{@{}c@{}}Continuous \fig{fig:environment}a\\ Ferromagnetic w/ Paint\end{tabular} \\ \hline
\end{tabular}
\begin{tablenotes}
      \footnotesize
      \item Climbing Gravity: simulated gravity used in their robot climbing tests, where $g \in \mathbb{R}^1$ is a gravitational constant. 
      Climbing velocity: the robot's body velocity while climbing. Normalized velocity is the climbing velocity normalized by the robot's body length. 
      * The body length of LEMUR $3$ is approximated \red{as the half} track length in \cite{lemur3}. 
      EPDM stands for Ethylene Propylene Diene Monomer.
    \end{tablenotes}
\end{threeparttable}
\end{table*}

\subsubsection{Whole-body Approach in Climbing\label{sec:whole_body}}
We explore SCALER's whole-body climbing strategy for overcoming difficult-to-grasp obstacles in \fig{fig:whole-body} and \ref{fig:whole-body-top}. \fig{fig:whole-body-graph} compares the RF gripper's sidepull forces and positions in the climbing direction with and without sidepull. The whole-body approach embraces the capabilities of GOAT grippers and SCALER.

When confronted with a thick, slippery obstacle plate, as depicted in \fig{fig:whole-body-top}a, SCALER employed a sidepull technique. The plate's dimensions \red{made it not possible to employ a pinch or a clamp grasp in a stable manner,} necessitating the whole-body \red{sidepull approach.} 
\fig{fig:whole-body-top}b-1 and \ref{fig:whole-body-top}b-2 visualize SCALER pushing itself to the left \red{and employing a} sidepull using the RF GOAT gripper. The close-up gripper front view in \fig{fig:whole-body-top}c resembles the sidepull rendered in \fig{fig:pinch}f. \red{Successively, $64$ steps were conducted over two experiments.}

In the sidepull phase shown in \fig{fig:whole-body-graph}, SCALER applied a lateral force to the RF gripper using a creep gait (\fig{fig:gait_on_leg}), enabling stable grasping. The minimum required sidepull force was estimated at \SI{15.8}{\newton} based on \eqref{eq:sidepull}. 
The cases, with and without sidepull, slipped at a similar rate until $t = \SI{16}{\second}$ as the sidepull force was insufficient. The case without sidepull couldn't meet this force requirement and slipped completely by $t = \SI{20}{\second}$. The case with sidepull stabilized after reaching \SI{20}{\newton}. 
This outcome aligned with the analytical prediction. SCALER successfully traveled over the \SI{432}{\milli\meter} obstacles and returned to the aluminum bar. The mean supporting force, the magnitude of the forces normal and parallel to the bar, is \SI{72}{\percent} higher than in the failure case, confirming that the whole-body approach using the C-GOAT gripper improved stability.

This whole-body strategy enhances SCALER's traversability further and demonstrates the symbiotic potential between the robot and its gripper in climbing.

\subsubsection{Bouldering Vertical Free-Climbing\label{sec:bouldering}}
This experiment demonstrates SCALER's ability to climb in discrete environments, particularly on a vertical bouldering wall. GOAT grippers have to grasp bouldering holds with a non-convex shape.
Conventional polymer-made bouldering holds were installed on a \SI{90}{\degree} wall, and SCALER was operated using a predefined manual trajectory. 
SCALER successfully climbed up four steps in \fig{fig:bouldering}, and each leg moved to the next bouldering hold at
$\SI{0.35}{\meter\per\minute}$ or a normalized speed of $\SI[per-mode=reciprocal]{1.0}{\per\minute}$. Over five trials, $12$ out of $16$ steps were successful, and one episode was fully successful with an open-loop trajectory.
SCALER stretched the legs farther with the torso mechanism, such as at $t = \SI{29}{\second}$ and $t = \SI{57}{\second}$ in \fig{fig:bouldering}, representing the kinematics benefit of SCALER's torso DoF.

SCALER's performance in the bouldering wall environment illustrates its potential in applications where maneuvering in discrete, unstructured environments is necessary. More discrete environment climbing with obstacles \red{was tested} in \cite{contact_rich}.

\section{Discussion and Limitation}
In this section, we discuss key takeaways and limitations from our experiments.
Results from hardware experiments are compiled and summarized in \tab{tb:comparision}, \ref{tb:climbing}.
\tab{tb:comparision} contrasts the SCALER 3-DoF walking configuration with other legged robots such as ANYmal and SPOT, revealing that SCALER achieves comparable normalized speeds.
SCALER surpasses all other quadrupeds in the payload capacity over twice its weight and normalized workload, as in \cite{quad_comparison}.

Table \ref{tb:climbing} compares SCALER to other state-of-the-art climbing robots regarding their operation environments, end-effector, and capabilities. SCALER has successfully shown traversability in various directions of gravity and versatile climbing performances across different climbing holds. 
We can distinguish climbing robots based on whether they have multi-finger grippers or not. This is a notable difference since the adhesive type of end-effectors can reduce the climbing problem down to locomotion under different gravitational force directions. Suction \cite{moclora}, magnetic \cite{marvel}, and gecko or Van der Waals force-based end-effectors generate normal forces regardless of the direction of gravity. 
Therefore, if such an end-effector can provide sufficient adhesive forces, the robot does not need to consider dexterity or grasping explicitly.

\subsection{Limitation}
\subsubsection{Rigidity and Compliance\label{sec:limit_compliance}}
Since SCALER is a loco-grasping platform, balancing conflicting requirements presents design challenges.
Rigidity in joints and linkages provides the benefit of accuracy and repeatability in position-controlled manipulation under load \cite{Wang2020}, while compliance offers mechanical adaptability \cite{HRP2_ladder}. Such adaptability is useful for compensating for system and environment \red{uncertainties}. 

Nonetheless, we observed failure cases in experiments due to leg compliance in high-friction environments such as sandpaper in Section \ref{sec:overhang} and in \cite{contact_rich}.
When SCALER approaches an object with a high-friction surface, it can fail to grasp the object if one of its fingers makes an undesired contact. In position control settings, the leg stiffness limits the force required to break this undesired contact unless the joint reaches its torque limit. This limitation arises because the deflection is solely due to the difference between the intended goal position and the point where the undesired contact occurs.

SCALER leg consists of three major compliance sources: parallel link mechanisms, backlash, and wear. The passive joints and carbon fiber tubes are relatively compliant, though the SCALER's two redundant front linkages enhance rigidity.
The SCALER's actuators use a spur gear with \SI{0.25}{\degree} backlash, translating to the maximum \SI{1.65}{\milli\meter} offset given leg length. 
Over time, the gear and output shaft wear increase backlash. 
The model in Section \ref{sec:compliance_model} helps us understand the characteristics of the SCALER leg compliance \red{and we utilized it for stiffness force control}. 
Leg compliance has been accounted for in planning as demonstrated in \cite{multi-surface} and \cite{risk_aware}.

\subsubsection{Limb Dexterity}
Another metric for evaluating grasping capability is the dexterous workspace. 
While dexterity can be improved by eliminating the end-effector offset from the spherical wrist, practical considerations such as gripper geometries and potential collisions with environments must also be accounted for.
Although the variants of the SCALER wrist in \fig{fig:wrist} have sufficient joint range for our designated tasks, the workspace can be constrained by self or environmental collisions, depending on the gripper's orientation and the position of the parallel link elbow joint. Consequently, it is important to consider potential collisions when planning trajectories along with kinematic and dynamic feasibility for all legs, fingers, and the body to ensure safe climbing.

\subsubsection{\red{Uncertainty and repeatability}\label{sec:episodic_failure}}
\red{In order to conduct safer and more robust climbing, it is essential to consider the uncertainty in the system \cite{risk_aware}. Stochasticity in the grasping contacts, particularly with the spine, introduces a large variance in the maximum force it can sustain \cite{risk_aware}, \cite{GOAT}.
For discrete environments, repeatability is subject to the initial robot state since our trajectory is generated offline, and the estimator tracks the state with respect to the initial state. 
From Section \ref{sec:multi_modal_climb} and \ref{sec:whole_body}, the hook success rate was \SI{40}{\percent}, which is less than pinch, clamp, and sidepull. The hook grasp required more positional accuracy since it relied on the kinematical constraints, as shown in \tab{tb:contact_mode}.}

Localization and state estimation errors are more critical in discrete environments. GOAT gripper's adaptiveness provides margins, but the estimation error can accumulate over time \cite{optistate}. In future work, additional mechanisms, such as visual servo, will be necessary to compensate for the accumulated error and achieve a higher success rate for tasks.

\subsection{GOAT Gripper Limitation}
\subsubsection{Contact State Transitions\label{sec:contact_transition}}
The GOAT gripper features spine-enhanced fingers that aid in grasping by penetrating an object's surface cavities. 
While the spine has shown notable performance in our experiments, the spine-enhanced fingers can also act as a barrier to gripper state transition due to their strong adherence. 
When the spine contact breaks while lifting the gripper, the internally stored energy in leg compliance 
is released abruptly, which can cause a non-negligible disturbance to the robot. 
This potentially accelerates the wear and damages the spine tips, reducing the life cycle of the finger, which is a common issue in spine grippers \cite{lemur3}. 

\subsubsection{Macro and Micro Scale Contacts}
Although the \red{C-}GOAT gripper can realize various grasping types, it can only explicitly consider larger-scale contact modes. However, relatively small-scale surface feature and dynamics significantly affects the success of grasping \cite{shirai2023tactile}. 
In \fig{fig:ceiling_bar}c, SCALER can walk upside-down with spine tips on bare metal bars. However, spine tips require a covered surface for vertical climbing.
The T-slotted aluminum has grooves along the bar, as shown in \fig{fig:ceiling_bar}b. 
In the vertical case, spines only resist forces perpendicular to the grooves, but not gravitational forces.
Hence, SCALER slips down along the grooves because the pure friction force between the spine tips and aluminum is insufficient. 
Hence, small-scale texture variations are significant, and improving micro contact properties, as in \cite{macro_micro}, can potentially benefit climbing robot graspability. 

\subsubsection{C-GOAT C-Shaped Finger Contacts\label{sec:limit_c_shape}}
The \red{C-GOAT} gripper enables SCALER to achieve multiple modes of grasping in climbing. However, the C-shaped finger has limited contact areas due to its curvature. Consequently, this design requires a significantly larger normal force, as the dry adhesive contact depends on the real molecular-level contact area \cite{macro_micro}. The pneumatic actuator used with the C-shaped finger in the GOAT gripper generates twice the normal force compared to the DC linear actuator. Furthermore, the state estimation of all seven modes remains an open problem.

\section{CONCLUSION}
The SCALER mechanisms, including torso and limb designs, coupled with the GOAT gripper, constitute a versatile loco-grasping platform capable of traversing various challenging terrains. SCALER's parallel-serial hybrid limb designs have adequately accomplished body dynamic loco-grasping and high-load capacities both in climbing and on the ground. The torso mechanisms allow SCALER to stretch the end-effector workspace on demand. The spine-enhanced GOAT grippers successfully support SCALER on the ceiling with slippery and rough surfaces. SCALER achieved dynamic climbing with the pneumatically actuated GOAT grippers, $13.8$ times faster than the previous fastest SCALER climbing.
With \red{C-GOAT} grippers, SCALER performs multi-modal grasping, thereby overcoming environments otherwise infeasible. 
SCALER has demonstrated versatility over vertical and inverted walls and has achieved comparable speeds to state-of-the-art robots on the ground. It exhibits one of the highest mechanical efficiencies as a quadruped robot, showcasing its potential for broader applications and research. \redrev{SCALER is one of the first to realize more than three multi-modal grasping with a single end-effector in free-climbing}. Hence, SCALER sets a new precedent for free-climbing robotics and advances the traversability of quadruped-limbed robots. 
\bibliographystyle{IEEEtran}
\bibliography{main}

\vspace{-45pt}
\begin{IEEEbiography}[{\includegraphics[width=1in,height=1.25in,clip,keepaspectratio]{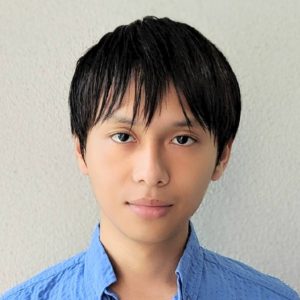}}]{Yusuke Tanaka}
(Member, IEEE) received his B.S. from the University of Massachusetts Amherst and is currently a Ph.D candidate at the University of California, Los Angeles. He has won SICE SIYA-IROS2022 award.
He is a research assistant at RoMeLa (Robotics \& Mechanisms Laboratory) at UCLA with a research interest in contact-rich simultaneous locomotion and grasping, and limbed climbing robotics. 
\end{IEEEbiography}
\vspace{-55pt}
\begin{IEEEbiography}[{\includegraphics[width=1in,height=1.25in,clip,keepaspectratio]{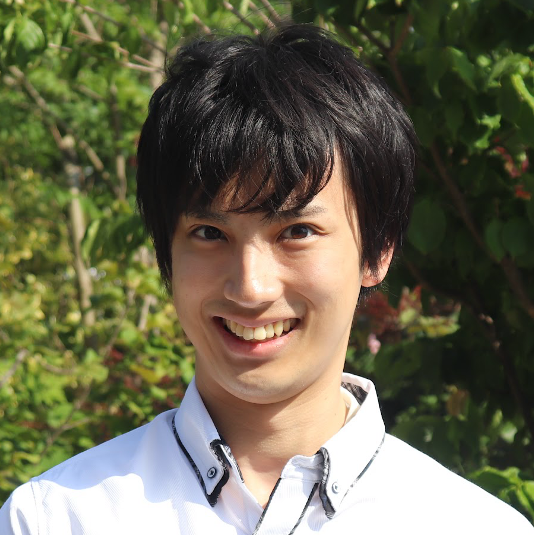}}]{Yuki Shirai}
(Member, IEEE) received the B.E. in Mechanical and Aerospace Engineering from Tohoku University, Sendai, Japan, in 2018, and the M.S. and the Ph.D. in Mechanical Engineering from the University of California, Los Angeles, Los Angeles, CA, USA, in 2019 and 2024, respectively. 
He is an associate editor of IEEE Robotics and Automation Letters (RA-L).
He is currently a Postdoctoral Research Fellow with the Optimization \& Intelligent Robotics team at Mitsubishi Electric Research Laboratories, Cambridge, MA, USA. His research interests lie in the intersection of optimization and machine learning for contact-rich dexterous manipulation and locomotion.
\end{IEEEbiography}
\vspace{-45pt}
\begin{IEEEbiography}[{\includegraphics[width=1in,height=1.25in,clip,keepaspectratio]{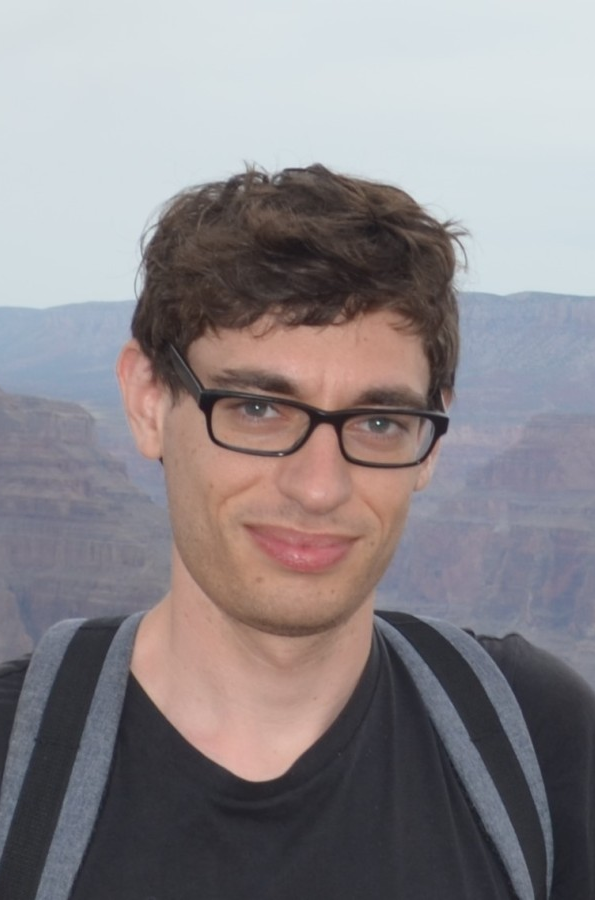}}]{Alexander Schperberg}
(Member, IEEE) received his B.S in Bioengineering from University of California, San Diego, Ph.D. in Robotics in Mechanical and Aerospace Engineering at the University of California, Los Angeles (UCLA), and is an Amazon fellow. He is currently a Postdoctoral Research Fellow with the Control for Autonomy team at Mitsubishi Electric Research Laboratories, Cambridge, MA, USA. His research focuses on the intersection of machine learning, control systems, and computer vision, with applications in legged locomotion, robotic manipulation, and Active SLAM.
\end{IEEEbiography}
\vspace{-45pt}
\begin{IEEEbiography}[{\includegraphics[width=1in,height=1.25in,clip,keepaspectratio]{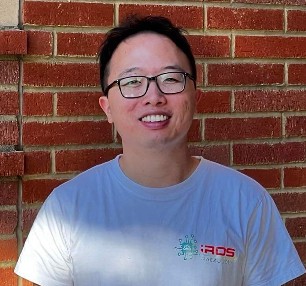}}]{Xuan Lin}
(Member, IEEE) received his Ph.D. in Mechanical Engineering from the University of California, Los Angeles, under the supervision of Prof. Dennis Hong. He is currently a postdoctoral researcher at the Georgia Institute of Technology, working with Prof. Ye Zhao. His research focuses on scalable motion planning algorithms using mixed-integer convex programming, with applications in search and rescue, logistics, and automation. His work has been recognized with the Best Paper Award on Safety, Security, and Rescue Robotics at IROS 2019 and as a finalist for the Best Paper Award at UR 2024.
\end{IEEEbiography}
\vspace{-45pt}
\begin{IEEEbiography}[{\includegraphics[width=1in,height=1.25in,clip,keepaspectratio]{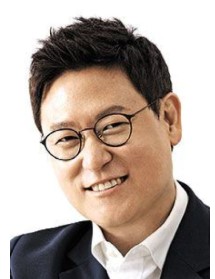}}]{Dennis Hong}
(Member, IEEE) is a Professor of Mechanical and Aerospace Engineering at the University of California, Los Angeles, and the Founding Director of RoMeLa (Robotics \& Mechanisms Laboratory). He received his B.S. degree in Mechanical Engineering from the University of Wisconsin-Madison (1994), his M.S. and Ph.D. in Mechanical Engineering from Purdue University (1999, 2002).
His research focuses on robot locomotion and manipulation, autonomous vehicles, and humanoid robots. The Washington Post magazine called Dr. Hong “the Leonardo da Vinci of robots.” His team won 3rd Place on DARPA Urban Challenge, Finalists in the DARPA Robotics Challenge, and is a 6-time World Champion in RoboCup Humanoid divisions.
\end{IEEEbiography}

\end{document}